\documentclass[11pt]{article}
\usepackage[utf8]{inputenc}
\usepackage[margin=1in]{geometry}
\usepackage[T1]{fontenc}
\usepackage{lmodern}
\usepackage{microtype}
\usepackage{graphicx}
\usepackage{subfigure}
\usepackage{booktabs} 
\usepackage{natbib}
\setcitestyle{open={(},close={)}}
\usepackage{algorithmic}
\usepackage{amsmath}
\usepackage{mathtools}
\usepackage{amsthm}
\usepackage{graphicx}
\usepackage{textcomp}
\usepackage{xcolor}
\usepackage{amsfonts}     
\usepackage{epsfig}
\usepackage{algorithm}
\usepackage{wrapfig}
\usepackage{balance}
\usepackage{url}
\usepackage{mathtools}
\usepackage{natbib}
\usepackage{multirow}

\DeclareMathOperator*{\argmax}{arg\,max}

\usepackage[textsize=tiny]{todonotes}

\usepackage[colorlinks=true,citecolor=blue]{hyperref}

\usepackage{amsmath,amsthm,amsfonts,amssymb,mathdots,array,mathrsfs,bm,bbm,stmaryrd,graphicx,subfigure,xcolor}

\usepackage{breakcites}

\usepackage[T1]{fontenc}
\usepackage{enumerate}
\usepackage{inputenc}

\usepackage{graphicx} 
\usepackage{subfigure}

\usepackage{booktabs,balance}
\usepackage{rotating}
\usepackage{boldline}
\usepackage{makecell}
\usepackage{multirow}
\usepackage{balance}

\usepackage{tikz}

\newtheorem{theorem}{Theorem}[section]

\newtheorem{lemma}[theorem]{Lemma}

\newtheorem{definition}{Definition}[section]
\newtheorem{example}{Example}[section]
\newtheorem{assumption}{Assumption}[section]
\newtheorem{remark}{Remark}[section]

\newcommand{\bsmat}{\begin{bmatrix} }
\newcommand{\esmat}{\end{bmatrix} }

\usepackage{natbib}
\setcitestyle{numbers}
\setcitestyle{square}

\newcommand{\indep}{\perp \!\!\! \perp}

\usepackage{environ}
\NewEnviron{smallequation}{%
    \begin{equation}
    \scalebox{0.97}{$\BODY$}
    \end{equation}
    }
    \NewEnviron{smallalign}{%
    \begin{equation}
    \scalebox{0.97}{$\BODY$}
    \end{equation}
    }

\begin{document}

\title{\bf\Huge Causal Bayesian Optimization via Exogenous Distribution Learning}

\author{\vspace{0.1in}\\\textbf{Shaogang Ren} \vspace{0.05in}  \\ 
shaogang-ren@utc.edu\\
Department of Computer Science and  Engineering\\
University of Tennessee at Chattanooga\\\\
\textbf{Zihao Wang} \vspace{0.05in}  \\ 
Department of Computer Science and  Engineering\\
University of Tennessee at Chattanooga\\\\
\textbf{Yuzhou Chen} \vspace{0.05in}  \\ 
Department of Statistics\\
University of California, Riverside\\\\
 \textbf{Xiaoning Qian} \vspace{0.05in}  \\
Department of Electrical and Computer Engineering\\
 Texas A\&M University\\
}

\date{\vspace{0.5in}}
\maketitle

\begin{abstract}\vspace{0.3in}
\vspace{0.1in}
Maximizing a target variable as an operational objective within a structural causal model is a fundamental problem. Causal Bayesian Optimization (CBO) approaches typically achieve this either by performing interventions that modify the causal structure to increase the reward or by introducing action nodes to endogenous variables, thereby adjusting the data-generating mechanisms to meet the objective. In this paper, we propose a novel method that learns the distribution of exogenous variables-an aspect often ignored or marginalized through expectation in existing CBO frameworks. By modeling the exogenous distribution, we enhance the approximation fidelity of the data-generating structural causal models (SCMs) used in surrogate models, which are commonly trained on limited observational data. Furthermore, the ability to recover exogenous variables enables the application of our approach to more general causal structures beyond the confines of Additive Noise Models (ANMs) and single-mode Gaussian, allowing the use of more expressive priors for context noise. We incorporate the learned exogenous distribution into a new CBO method, demonstrating its advantages across diverse datasets and application scenarios.
\end{abstract}

\vspace{-0.05in}
\section{Introduction}\label{sec:intro}
\vspace{-0.05in}
Bayesian Optimization~(BO) is widely applied in domains such as automated industrial processes, drug discovery, and synthetic biology, where the objective is to optimize black-box functions~\citep{movckus1975bayesian,astudillo2019bayesian,garnett2023,frazier2018tutorial}. In many real-world scenarios, structural knowledge of the unknown objective function is available and can be exploited to enhance the efficiency of BO. Causal Bayesian Optimization~(CBO) has been developed to incorporate such structural information~\citep{aglietti2020causal,aglietti2021dynamic,sussex2022model,gultchin2023functional}. CBO integrates principles from causal inference, uncertainty quantification, and sequential decision-making. Unlike traditional BO, which assumes independence among input variables, CBO accounts for known causal relationships among them~\citep{aglietti2020causal}. This framework has been successfully applied to optimize medical and ecological interventions~\citep{aglietti2020causal,aglietti2021dynamic}, among other applications.

\subsection{Approach and Contributions}
In this paper, we propose a novel method called \emph{EXogenous distribution learning augmented {C}ausal {B}ayesian {O}ptimization} (EXCBO). Given observational data from a structural causal model~(SCM~\cite{pearl2009causality,pearl1995causal}), our method recovers the exogenous variable corresponding to each endogenous node using an encoder-decoder framework, as illustrated in Figure~\ref{fig:node}. The recovered exogenous variable distribution is then modeled using a flexible density estimator, such as a Gaussian Mixture Model. This learned distribution significantly enhances the surrogate model's approximation of the underlying SCM, as shown in Figure~\ref{fig:cbo}.

Unlike existing CBO approaches~\citep{aglietti2020causal,aglietti2021dynamic,sussex2022model}, which are typically confined to Additive Noise Models~(ANMs~\cite{hoyer2008nonlinear}), our method generalizes CBO to broader classes of causal models. By enabling the recovery of exogenous variables and their distributions, our surrogate model provides improved accuracy and flexibility for causal inference in the CBO update process.

The contributions of this work are as follows:
\begin{itemize}
  \item We introduce a method for recovering the exogenous noise variable of each endogenous node in an SCM using observational data, which enables our model to capture \emph{multimodal exogenous distributions}. 
  
  \item This flexible approach to learning exogenous distributions allows our CBO framework to extend naturally to general causal models beyond the limitations of ANMs.
  
  \item We present a theoretical investigation of exogenous variable recovery through the proof of counterfactual identification, and we further analyze the regret bounds of the proposed algorithm.
  
  \item We conduct extensive experiments to evaluate the impact of exogenous distribution learning and demonstrate the practical advantages of EXCBO through applications such as epidemic model calibration, COVID-19 testing, and real-world planktonic predator–prey problem, etc.
\end{itemize}

The remainder of the paper is organized as follows. Section~\ref{sec:background} reviews background and related work. Section~\ref{sec:problem} introduces the problem setup and outlines our proposed CBO framework. Section~\ref{sec:exogenous} presents the method for recovering exogenous variables. The proposed algorithm, EXCBO, is detailed in Section~\ref{sec:excbo}, followed by regret analysis in Section~\ref{sec:theory}. Experimental results are presented in Section~\ref{sec:experiment}, and the paper concludes in Section~\ref{sec:conclude}.

\section{Background}~\label{sec:background}
We provide a brief overview of SCMs, intervention mechanisms, and CBO in this section.

\vspace{-0.05in}
\subsection{Structural Causal Model}
\vspace{-0.05in}
An SCM is denoted by $\mathcal{M} = (\mathcal{G}, \mathbf{F}, \mathbf{V}, \mathbf{U})$, where $\mathcal{G}$ is a directed acyclic graph~(DAG), $\mathbf{F} = \{f_i\}_{i=0}^d$ represents the $d+1$ structural mechanisms, $\mathbf{V}$ denotes the set of endogenous variables, and $\mathbf{U}$ the set of exogenous (background) variables. The generation of the $i$th endogenous variable follows
\begin{align}~\label{eq:node_scm}
&X_i = f_i(\mathbf{Z}_i, U_i);  \ \mathbf{Z}_i = \mathbf{pa}(i), \  U_i\sim p(U_i),  \ \text{for} \  i \in [d] .
\end{align}
Here, $[d] = \{0,1,\ldots,d\}$, and $X_i$ refers to both the variable and its corresponding node in $\mathcal{G}$. The set $\mathbf{pa}(i)$ denotes the parents of node $i$, while $\mathbf{ch}(i)$ refers to its children. We assume $U_i \indep \mathbf{Z}_i$ and $U_i \indep U_j$ for all $i \neq j$. Each $f_i$ is a mapping from $\mathbb{R}^{|\mathbf{pa}(i)|+1}$ to $\mathbb{R}$. The domains of $X_i$, $\mathbf{Z}_i$, and $U_i$ are denoted by $\mathcal{X}_i$, $\mathbf{\mathcal{Z}}_i$, and $\mathcal{U}_i$, respectively. Additionally, we assume that the expectation $\mathbb{E}[X_i]$ exists for all $i \in [d]$.
Most existing CBO approaches~\citep{aglietti2020causal,aglietti2021dynamic,sussex2022model} typically assume an Additive Noise Model~(ANM~\cite{hoyer2008nonlinear}) for exogenous variables, where
$X_i = f_i(\mathbf{Z}_i) + U_i$ with $U_i \sim \mathcal{N}(0,1)$.
\vspace{-0.05in}
\subsection{Intervention}
\vspace{-0.05in}
In an SCM $\mathcal{M}$, let $\mathbf{I} \subset \mathbf{V}$ be a set of endogenous variables targeted for intervention. The post-intervention structural mechanisms are represented by 
$\mathbf{F}_{x} = \{ f_i \mid X_i \notin \mathbf{I} \} \cup  \{   f_j \mid X_j \in \mathbf{I}  \}$.
A hard intervention replaces the mechanism for each $X_j \in \mathbf{I}$ with a constant value, resulting in 
$\mathbf{F}_{x} = \{ f_i \mid X_i \notin \mathbf{I} \} \cup  \{   f_j := \alpha_j \mid X_j \in \mathbf{I}  \}$,
where $\boldsymbol{\alpha}$ is the realized value of the intervened variables. This corresponds to Pearl's do-operation~\citep{pearl2009causality}, denoted as $do(\mathbf{X}_{\mathbf{I}} := \boldsymbol{\alpha})$, which alters $\mathcal{M}$ to a new model $\mathcal{M}_{\boldsymbol{\alpha}}$ by severing the dependencies between each $X_j$ and its parents.

This paper focuses on soft (or imperfect) interventions~\citep{peters2017elements}. Following the Model-based CBO framework~\citep{sussex2022model}, we associate each endogenous variable with an action variable, modifying the mechanisms as
$\mathbf{F}_{x} = \{ f_i \mid X_i \notin \mathbf{I} \} \cup  \{   f_j := f_j(\mathbf{Z}_j, \mathbf{A}_j, U_j) \mid X_j \in \mathbf{I}  \}$,
where $\mathbf{Z}_j = \mathbf{pa}(j)$. Under soft intervention, the data-generating mechanism becomes
\begin{align}\label{eq:Xi}
X_i  =
  \begin{cases} 
   f_i(\mathbf{Z}_i, U_i),     & \quad \text{if } X_i \notin \mathbf{I}\\
  f_i(\mathbf{Z}_i, \mathbf{A}_i, U_i),  & \quad \text{if } X_i \in \mathbf{I}
  \end{cases},
\end{align}
where $\mathbf{A}_i$ is a continuous action variable set associated with $X_i$ and takes values in $\mathcal{A}_i$. The soft intervention is represented using Pearl's notation as $do\big(\mathbf{X}_\mathbf{I} := \mathbf{f}(\mathbf{Z}_\mathbf{I}, \mathbf{A}, U_\mathbf{I}) \big)$.
\vspace{-0.05in}
\subsection{Function Network Bayesian Optimization}
\vspace{-0.05in}
Function Network BO~(FNBO~\cite{astudillo2021bayesian,astudillo2019bayesian}) operates under similar assumptions as CBO, where the functional structure is known but the specific parameterizations are not. FNBO applies soft interventions and employs an expected improvement~(EI) acquisition function to guide the selection of actions. However, FNBO assumes a noiseless environment, which may limit its applicability in practical settings. Both FNBO and CBO contribute to the broader effort of leveraging structured observations to improve the sample efficiency of standard BO techniques~\citep{astudillo2021thinking}.
\vspace{-0.05in}
\subsection{Causal Bayesian Optimization}\label{sec:cbo}
\vspace{-0.05in}
CBO performs sequential actions to interact with an SCM $\mathcal{M}$. The causal graph structure $\mathcal{G}$ is assumed known, while the functional mechanisms $\mathbf{F} = \{f_i\}_{i=0}^d$ are fixed but unknown. CBO uses probabilistic surrogate models - typically Gaussian Processes (GPs~\cite{williams2006gaussian}) - to guide the selection of interventions for maximizing the objective.

In~\citep{aglietti2020causal}, a CBO algorithm was introduced to jointly identify the optimal intervention set and the corresponding input values that maximize the target variable in an SCM. Dynamic CBO (DCBO)~\citep{aglietti2021dynamic} extends this approach to time-varying SCMs where causal effects evolve over time.

The MCBO method~\citep{sussex2022model} optimizes soft interventions to maximize the target variable within an SCM. In this setting, each edge function becomes $f_i: \mathcal{Z}_i \times \mathcal{A}_i \to \mathcal{X}_i$. Let $x_{i,t}$ denote the observation of node $X_i$ at time step $t$, for $i \in [d]$ and $t \in [T]$, where $T$ is the total number of time steps. At each step $t$, actions $\mathbf{a}_{:t} = \{\mathbf{a}_{i,t}\}_{i=0}^d$ are selected, and the resulting observations $\mathbf{x}_{:,t} = \{x_{i,t}\}_{i=0}^d$ are recorded. The relationship between action $\mathbf{a}_{i,t}$ and the observation is modeled using an additive noise structure:
$x_{i,t} = f_i(\mathbf{z}_{i,t}, \mathbf{a}_{i,t}) + u_{i,t}, \quad \forall i \in [d].$
For the target node $d$, the action is fixed at $\mathbf{a}_{d,t} = 0$, and the observed outcome is
$y_t = f_d(\mathbf{z}_{d,t}, \mathbf{a}_{d,t}) + u_{d,t},$
where $y_t$ depends on the entire intervention vector. The optimal action vector $\mathbf{a}^*$ that maximizes the expected reward is obtained by solving
$\mathbf{a}^* = \argmax_{\mathbf{a} \in \mathcal{A}} \mathbb{E}[y \mid \mathbf{a}].$
A GP surrogate model is employed to approximate the reward function and guide the BO process toward optimizing $y$.
\begin{figure}
\centering
\includegraphics[width=0.2\textwidth]{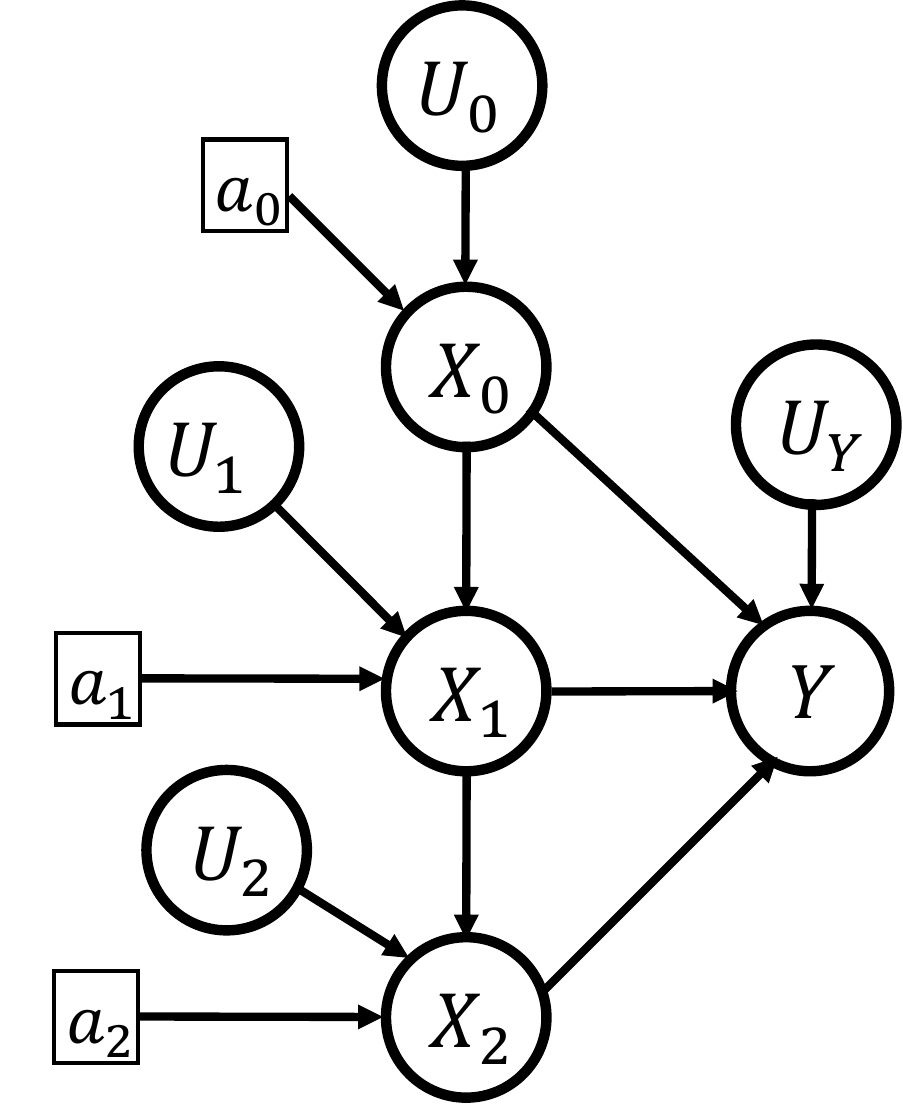}
\caption{\textbf{EXCBO}: Causal Bayesian Optimization via exogenous distribution learning. The distribution of $U_i$ is approximated using the density of the recovered surrogate $\widehat{U}_i$. EXCBO searches for the action vector $\mathbf{a}$ that maximizes the reward $Y$.}~\label{fig:cbo}
\end{figure}

\vspace{-0.05in}
\section{Problem Statement}~\label{sec:problem}
Following prior CBO approaches~\citep{aglietti2020causal,aglietti2021dynamic,sussex2022model,frazier2018tutorial}, we assume that the DAG $\mathcal{G}$ is known. Our framework employs GP surrogate models to guide the optimization of soft interventions, which are controlled via an action vector $\mathbf{a} = \{\mathbf{a}_i\}_{i=0}^d$, with the goal of maximizing the reward. This section details the specific problem setting addressed in this work.
\vspace{-0.05in}
\subsection{Assumptions for EXCBO}
\vspace{-0.05in}
We assume that the causal structure, represented by the DAG $\mathcal{G}$ of the SCM $\mathcal{M} = (\mathcal{G}, \mathbf{F}, \mathbf{V}, \mathbf{U})$, is given. This paper focuses exclusively on this setting. Additionally, we assume that $\mathcal{M}$ is causally sufficient, meaning all endogenous variables in $\mathbf{V}$ are observable. The problems of causal structure learning and handling unobserved confounders are left for future work.
\vspace{-0.05in}
\subsection{CBO via Exogenous Distribution Learning} 
\vspace{-0.05in}
In contrast to prior CBO approaches based on ANMs~\citep{aglietti2021dynamic,sussex2022model}, we propose a more flexible modeling of the mappings $f_i()$ by explicitly incorporating exogenous variables. To this end, we introduce EXCBO - a framework for CBO that leverages exogenous distribution learning, as illustrated in Figure~\ref{fig:cbo}.

Let $\mathcal{R}$ denote the set of root nodes. Since root nodes have no parents, we set $\mathbf{z}_{i,t} = \mathbf{0}$ for all $i \in \mathcal{R}$. Similarly, we define $\mathbf{a}_{d,t} = 0$ at the target node $d$, and denote the reward at time $t$ as
$y_{t} = f_d(\mathbf{z}_{d,t}, \mathbf{a}_{d,t}, u_{d,t}).$ Given an action vector $\mathbf{a} = \{\mathbf{a}_i\}_{i=0}^d$ and exogenous variables $\mathbf{u} = \{u_i\}_{i=0}^d$, the reward is denoted as $y = \mathbf{F}(\mathbf{a}, \mathbf{u})$. The optimization objective becomes
\begin{align}~\label{eq:cbo_exo}
&\mathbf{a}^* = \argmax_{\mathbf{a} \in \mathcal{A}} \mathbb{E}[y \mid \mathbf{a}],
\end{align}
where the expectation is taken over the exogenous variables $\mathbf{u}$. The goal is to identify a sequence of interventions $\{\mathbf{a}_t\}_{t=0}^T$ that achieves high average expected reward. To evaluate convergence, we study the cumulative regret over a time horizon $T$:
$R_T = \sum_{t=1}^T \left[ \mathbb{E}[y \mid \mathbf{a}^*] - \mathbb{E}[y \mid \mathbf{a}_{:,t}] \right].$ In our experiments, we use the observed objective or reward value $y$ as the primary performance metric for comparing EXCBO against baseline methods. The best choice of evaluation metric may vary depending on the application and the effectiveness of the optimized action sequence.

\vspace{-0.05in}
\subsection{Motivations for Exogenous Distribution Learning}

In existing CBO frameworks, the distributions of exogenous variables are either ignored or marginalized to simplify the intervention process~\citep{aglietti2020causal,aglietti2021dynamic,sussex2022model}. Learning the exogenous distribution, however, yields a more accurate surrogate model when observational data is available. As outlined in later sections, we propose an encoder-decoder architecture (illustrated in Figure~\ref{fig:node}) to recover the exogenous variable associated with each endogenous node in an SCM. The distribution of an exogenous variable $U_i$ is approximated by the density of its recovered surrogate $\widehat{U}_i$, modeled using a flexible distribution such as a Gaussian Mixture. This learned exogenous distribution improves the surrogate model's approximation of the underlying SCM.  As discussed in Sections~\ref{sec:onenode}, \ref{sec:proof_thm}, and~\ref{sec:edl_app}, under moderate assumptions, the independence between the recovered exogenous variable $\widehat{U}$ and both the parents $\mathbf{Z}$ and  actions $\mathbf{A}$ empowers the structured surrogate model in EXCBO to be \emph{counterfactually identifiable} and to perform effective interventions.

As a result, EXCBO extends beyond the ANM framework assumed by prior work~\citep{aglietti2020causal,aglietti2021dynamic,sussex2022model}, enabling optimization under a broader class of causal models. Moreover, by enhancing the surrogate model's fidelity, our approach can potentially achieve superior reward outcomes. Additional justification and motivation are provided in the Appendix.

\subsection{Decomposable Generation Mechanism}~\label{sec:dgm}

In our setting, the edges in the SCM $\mathcal{M}$ correspond to a fixed but unknown set of functions $\mathbf{F} = \{f_i\}_{i=0}^d$. We assume the structure of the SCM is known and that the system is causally sufficient—that is, it contains no hidden variables or confounders. We now define the \emph{Decomposable Generation Mechanism (DGM)} used in our analysis.

\begin{definition}
(DGM) A data-generating function $f$ follows a decomposable generation mechanism if $X = f(\mathbf{Z}, U) = f_a(\mathbf{Z}) + f_b(\mathbf{Z}) f_c(U)$, where $f_a: \mathcal{Z} \to \mathbb{R}$, $f_b: \mathcal{Z} \to \mathbb{R}$, and $f_c: \mathcal{U} \to \mathbb{R}$. All mappings are continuous, and $f_b(\mathbf{z}) \neq 0$ for all $\mathbf{z} \in \mathcal{Z}$.
\end{definition}

In a DGM, the function $f_c(U)$ may be a one-dimensional, nonlinear, and nonmonotonic transformation of the exogenous variable $U$. The term $f_b(\mathbf{Z}) f_c(U)$ implies that the variance of the generated variable $X$, conditioned on its parents $\mathbf{Z}$, depends on both $U$ and $\mathbf{Z}$. Consequently, DGMs represent a broad class of mechanisms in which both parents and exogenous variables contribute to variance modulation.

This modeling framework is notably more general than \emph{Location-Scale or Heteroscedastic Noise Models (LSNMs)}~\citep{immer2023identifiability}, which typically assume linear $f_c()$ and strictly positive $f_b()$. Therefore, DGMs constitute a superset of LSNMs. In Section~\ref{sec:onenode}, we demonstrate that the distribution of exogenous variables can be recovered when the data-generating mechanism $f$ in each node~\eqref{eq:node_scm} adheres to the DGM formulation.

\section{Exogenous Distribution Learning}~\label{sec:exogenous} 
Given observations of an endogenous node and its parents within an SCM, our goal is to recover the distribution of that node’s exogenous variable. This exogenous distribution learning is carried out using GPs. We begin by focusing on the recovery of the exogenous distribution for a single node.

\vspace{-0.05in}
\subsection{Exogenous Variable Recovery for One Node}\label{sec:onenode}
According to~\eqref{eq:Xi}, an endogenous variable $X_i$ may or may not have  associated action variables $\mathbf{A}_i$. To simplify notation, we use $\mathbf{Z}_i$ in this section to denote both the parents of $X_i$ and its action variable, i.e., $\mathbf{Z}_i = (\mathbf{Z}_i, \mathbf{A}_i)$ if $X_i \in \mathbf{I}$. The task of learning the exogenous distribution for $X$ then becomes the problem of recovering the distribution of $U$ given observations of $X$ and $\mathbf{Z}$ from the generative model $X = f(\mathbf{Z}, U)$. For clarity, we define the causal mechanism for the triplet $(\mathbf{Z}, U, X)$ corresponding to a single node in an SCM.
\begin{figure}
\centering
\includegraphics[width=0.08\textwidth]{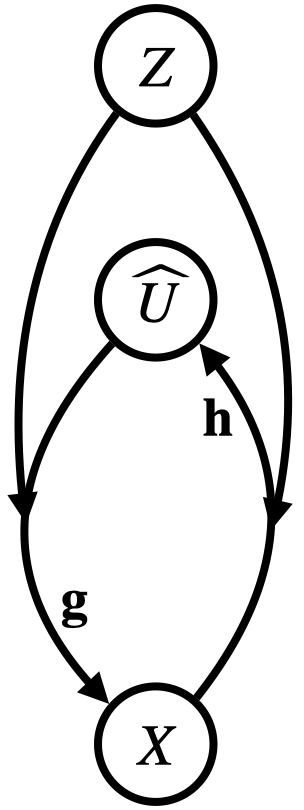}
  \vspace{-0.02in}
\caption{Structure in one node. $\mathbf{Z}$ denotes the parent set of $X$. Our algorithm learns an encoder $h$ and a decoder $g$ such that the surrogate $\widehat{U} = h(\mathbf{Z}, X)$ and $X = g(\mathbf{Z}, \widehat{U})$.} \label{fig:node}
\end{figure}

\begin{assumption}~\label{assum:node_scm}
 Let $X$ be a node of an SCM $\mathcal{M}$, and let $f()$ be the causal mechanism generating $X$ with parent $\mathbf{Z}$ and  an exogenous variable $U$,  i.e. $X = f(\mathbf{Z}, U)$. We use $(\mathbf{Z}, U, X, f)$ to denote the \emph{node SCM} of $X$, and we  assume $\mathbf{Z} \indep U$.
\end{assumption} 

We have Assumption~\ref{assum:node_scm} for any SCM discussed in this paper. 
In a node SCM, $\mathbf{Z}$ may be multi-dimensional, representing the parents of $X$, while $U$ is the exogenous variable. This differs from the \emph{Bijective Generation Mechanism} (BGM~\cite{nasr2023counterfactual}), where $f(\mathbf{Z}, U)$ is assumed to be monotonic and invertible with respect to $U$ given fixed $\mathbf{Z}$.

We adopt an encoder-decoder framework (Figure~\ref{fig:node}) to construct a surrogate for the exogenous variable. For an observation $(\mathbf{z}, x)$, we use $x(\mathbf{z}, u)$ to denote $f(\mathbf{z}, u)$, and here $u$
is an exogenous value  generating $x$. Here $u \sim p(U)$, and $p(U)$ is the exogenous distribution regarding node $X$. The encoder and decoder are learned via BO~\citep{balandat2020botorch} and a training set that involves $N$ observations or base samples. By following the analysis of~\cite{balandat2020botorch}, we have the following definition. 

\begin{definition}\label{def:eds}
(Encoder-Decoder Surrogate; EDS) Let $(\mathbf{Z}, U, X, f)$ be a node SCM. Let $\phi(): \mathbf{\mathcal{Z}} \to \mathcal{X}$ be a  probabilistic regression model. Each $\mathbf{z} \in \mathcal{Z}$ has $N$ base samples in the close neighborhood of $\mathbf{z}$, $\big\{\mathbf{z}, x_i(\mathbf{z}, u_i)\big\}_{i=1}^N$, and here $\{u_i\}_{i=1}^N \ i.i.d \sim p(U)$. In addition, $\phi()$ has a  prediction mean $\mu_{\phi}(\mathbf{z}) = \frac{1}{N} \sum^N_i x_i(\mathbf{z}, u_i)$ and a variance $\sigma^2_{\phi}(\mathbf{z}) = \frac{1}{N-1} \sum^N_i \big(x_i(\mathbf{z}, u_i) - \mu_{\phi}(\mathbf{z}) \big)^2 $. We define $(\widehat{U}, \phi, h, g)$ as an encoder-decoder surrogate (EDS) for the exogenous variable $U$, where the encoder is $h():\mathcal{Z} \times \mathcal{X} \to \widehat{\mathcal{U}}$, defined as $\widehat{U} := h(\mathbf{Z}, X) := \frac{X - \mu_{\phi}(\mathbf{Z})}{\sigma_{\phi}(\mathbf{Z})}$, and the decoder is $g(): \mathcal{Z} \times \widehat{\mathcal{U}} \to \mathcal{X}$. 
\end{definition}

Given observations of $X$ and its parents $\mathbf{Z}$, our method learns the encoder $h()$ to approximate the true value of $U$ via $\widehat{u} = h(\mathbf{z}, x)$. Concurrently, the decoder $g()$ serves as a surrogate for the causal mechanism $f()$, reconstructing $x = g(\mathbf{z}, \widehat{u})$. GPs are effective to learn  $\sigma^2_{\phi}()$ and $\mu_{\phi}()$ in EDS, and we $\phi()$ and $g()$ are implemented via GPs in our experiments. 
Theorem~\ref{thm:exo1} establishes that surrogate values of the exogenous variable $U$ can be recovered from observations under the DGM assumption on $f$. 

\begin{theorem}~\label{thm:exo1}
Let $(\mathbf{Z}, U, X, f)$ be a node SCM, and $(\widehat{U}, \phi, h, g)$ an EDS surrogate of $U$. Suppose $f$ has the DGM structure, i.e. $X = f(\mathbf{Z}, U) = f_a(\mathbf{Z}) + f_b(\mathbf{Z}) f_c(U)$ with $f_b(\mathbf{z}) \neq 0$ for all $\mathbf{z} \in \mathcal{Z}$. In addition, each $\mathbf{z} \in \mathcal{Z}$ has $N$ base samples in the close neighborhood of $\mathbf{z}$,  i.e., $\big\{\mathbf{z}, x_i(\mathbf{z}, u_i)\big\}_{i=1}^N$ with $\{u_i\}_{i=1}^N \ i.i.d \sim p(U)$. 
Then with $N \rightarrow \infty$,  the surrogate $\widehat{U} \rightarrow \frac{s}{\sigma_{f_c}} \big(f_c(U)  -   \mathbb{E}[f_c(U)] \big)$, $\mathbb{E}[\widehat{U}] \rightarrow 0$, $\textrm{Var}[\widehat{U}] \rightarrow 1$, and $\widehat{U} \indep \mathbf{Z}$, where $\sigma_{f_c} = \sqrt{\mathbb{E}\big[\big(f_c(U)  -   \mathbb{E}[f_c(U)] \big)^2\big]}, s\in\{-1, 1\}$.
\end{theorem}

\begin{figure}
\centering 
\includegraphics[width=0.36\textwidth]{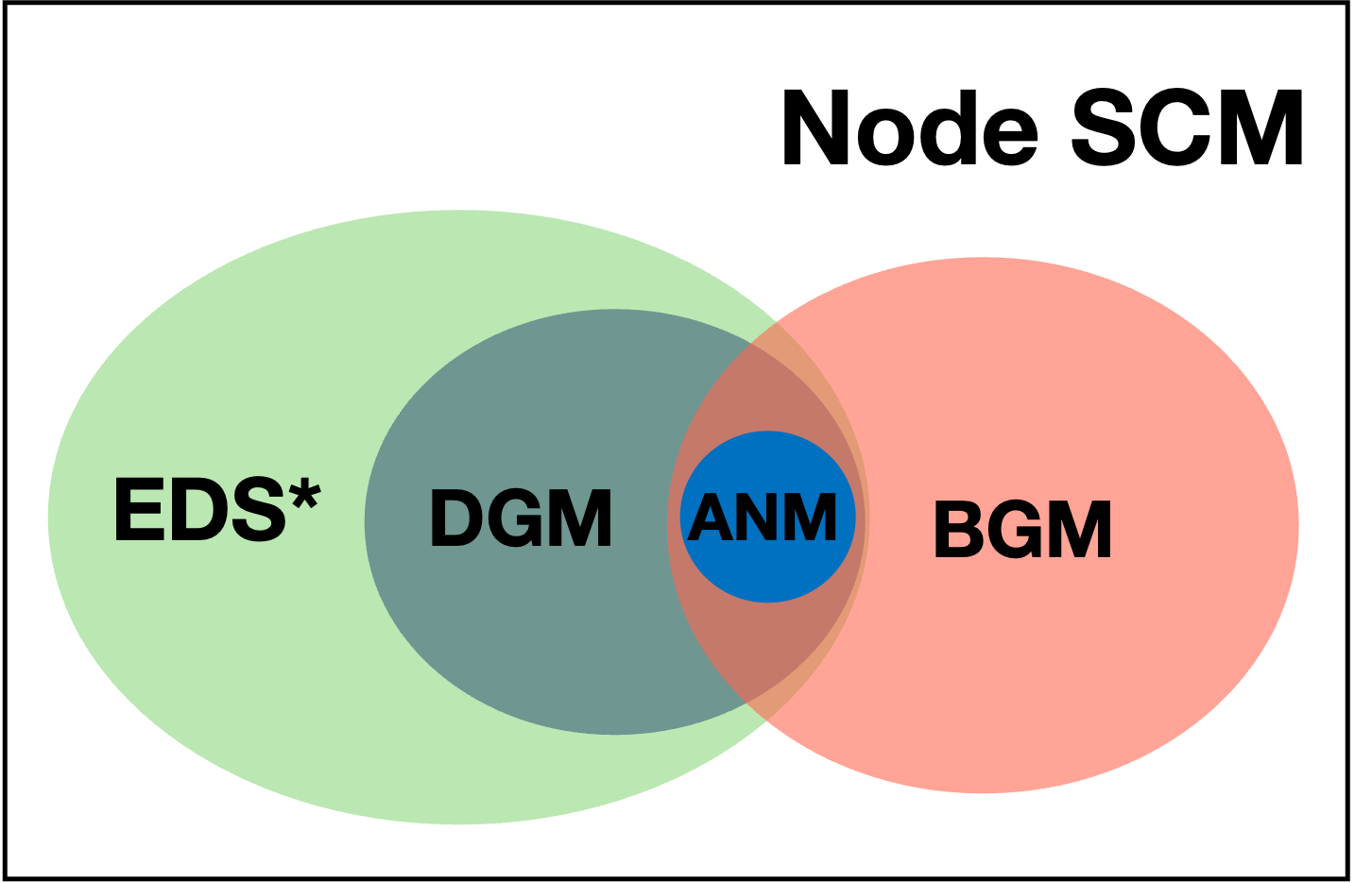}
\caption{Scopes of different mechanisms.} \label{fig:scope}
\end{figure}

According to Theorem~\ref{thm:exo1}, for triplets $(x, \mathbf{z}, u)$ with $\mathbf{z} \in  \mathcal{Z}$,  $ u \in  \mathcal{U}$,  $x$ generated with DGM, and with the number of base samples $N \rightarrow \infty$, the surrogate $\widehat{u} = \frac{s}{\sigma_{f_c}} \big(f_c(u)  -   \mathbb{E}[f_c(U)] \big)$, and the exogenous surrogate $\widehat{U} \indep \mathbf{Z}$. Notably, our framework generalizes beyond ANM (linear)~\cite{hoyer2008nonlinear} and BGM (monotonic)~\citep{nasr2023counterfactual} to a new class of nonlinear, nonmonotonic, and \emph{counterfactually identifiable} models through DGM.  Definition and analysis on counterfactual identifiability can be found in Appendix-\ref{sec:edl_app}. This extends the identifiability of $U$ significantly beyond the standard assumption $X = f(\mathbf{pa}(X)) + U$ used in many BO and CBO methods. We use EDS$^*$ to represent the node SCMs that are counterfactually identifiable via EDS either with or without the condition of $\widehat{U} \indep \mathbf{Z}$. Figure~\ref{fig:scope} illustrates the relationship among different data generation mechanisms regarding ~\emph{counterfactual identifiability}.
We use the distribution of the recovered surrogate $\widehat{U} =  h(\mathbf{Z}, X)$ - denoted as $p(\widehat{U})$ - as a proxy for the true $p(U)$ in the surrogate model. Consequently, the function $f$ is approximated via the learned decoder $g$ and the surrogate $\widehat{u}$: 
\[x = f(\mathbf{z}, u) = g(\mathbf{z}, \widehat{u}) = g(\mathbf{z}, h(\mathbf{z}, x)).\]

The proof of Theorem~\ref{thm:exo1} is provided in Appendix~\ref{sec:proof_thm}. Our surrogate variable $\widehat{U}$ and encoder $h()$ are valid under both DGM and BGM assumptions. In the BGM case, recovery of $U$ requires enforcing $\widehat{U} \indep \mathbf{Z}$, as detailed in Appendix~\ref{sec:edl_app}, which can be achieved through independence regularization - albeit at additional computational cost. If $f$ does not satisfy the DGM or BGM assumptions, then the recovered $\widehat{U}$ may be dependent on $\mathbf{Z}$, potentially degrading the accuracy of the surrogate model and limiting the effectiveness of CBO in finding optimal $y$ using limited data.
\subsection{Implementation of Exogenous Distribution Learning}~\label{sec:implement_edl}
The encoder-decoder architecture in Figure~\ref{fig:node} can be implemented in various ways, such as using Variational Autoencoders (VAEs)~\citep{kingma2013auto} or sample efficient deep-generative models~\citep{liang2024diffusion,wang2023flowreg}. To keep the implementation straightforward, we adopt GP regression for both the encoder and decoder, consistent with the EDS definition in Definition~\ref{def:eds}.  

For nodes with  action variables $\mathbf{A}$, the decoder becomes $g(): \mathcal{Z} \times \mathcal{A} \times \widehat{\mathcal{U}} \to \mathcal{X}$, and the encoder becomes $h(): \mathcal{Z} \times \mathcal{A} \times \mathcal{X} \to \widehat{\mathcal{U}}$, while the regression model is $\phi(): \mathcal{Z} \times \mathcal{A} \to \mathcal{X}$. Both $g()$ and $\phi()$ are implemented using GP regression models~\citep{williams2006gaussian}. To approximate the distribution of the recovered exogenous surrogate $\widehat{U}$, we use a Gaussian Mixture model to estimate $p(\widehat{U})$, which serves as a replacement for $p(U)$ in the probabilistic surrogate objective. For all nodes in the SCM $\mathcal{M}$, we denote the collection of decoders as $\mathbf{G} = \{g_i\}_{i=0}^d$ and the collection of encoders as $\mathbf{H} = \{h_i\}_{i=0}^d$.

\section{CBO with Exogenous Distribution Learning}~\label{sec:excbo}
In this section, we present the EXCBO algorithm, describing the probabilistic model and acquisition function used.
\subsection{Statistical Model}
In our model, the function $f_i$ that generates variable $X_i$ is learned through $g_i$, and $X_i = g_i(\mathbf{Z}_i, \mathbf{A}_i, \widehat{U}_i)$. We use  GPs~\citep{williams2006gaussian} to  learn  the surrogate  of $g_i$, i.e.,  $\tilde{g}_i$. For $i\in [d]$, let $\mu_{g,i,0}$ and $\sigma_{g,i,0}$ denote the prior mean and variance function for each $f_i$, respectively. 
At step $t$, the observation set is $\mathcal{D}_t=\{\mathbf{z}_{:, 1:t},\mathbf{a}_{:, 1:t}, x_{:, 1:t}\}$.  The posterior of $g_i$ with the input of node $i$, $({\mathbf{z}}_{i}, {\mathbf{a}}_{i}, {\widehat{u}}_{i})$, is given by 
\begin{align*}
&{g}_{i,t}({\mathbf{z}}_{i}, {\mathbf{a}}_{i}, {\widehat{u}}_{i}) \sim \mathcal{GP}(\mu_{g,i,t-1}, \sigma^2_{g,i,t-1}) ; \\ 
&\mu_{g, i,t-1} = \mu_{g,i,t-1}({\mathbf{z}}_{i}, {\mathbf{a}}_{i}, {\widehat{u}}_{i}) ; \\
&\sigma_{g,i,t-1}= \sigma_{g,i,t-1}({\mathbf{z}}_{i}, {\mathbf{a}}_{i}, {\widehat{u}}_{i}) .
\end{align*}
Then ${x}_{i, t}= {g}_{i,t}({\mathbf{z}}_{i}, {\mathbf{a}}_{i}, {\widehat{u}}_{i})$ denotes observations from one of the plausible models. Here $ {\widehat{u}}_{i} \sim p(\widehat{U}_i)$ in the sampling of the learned distribution of $\widehat{U}_i$. 

Given an observation $(\mathbf{z}_i, \mathbf{a}_i, x_i)$  at node $i$, the exogenous recovery $\widehat{u}_i =h_i(\mathbf{z}_i, \mathbf{a}_i, x_i) = \frac{x_i - \mu_{\phi,i}(\mathbf{z}_i, \mathbf{a}_i)}{\sigma_{\phi,i}(\mathbf{z}_i, \mathbf{a}_i)}$.  At time step $t$, the posterior of $\phi_i$ with the input of node $i$, $(\mathbf{z}_{i}, \mathbf{a}_{i})$, is given by 
\begin{align}~\label{eq:phi}
&{\phi}_{i,t}(\mathbf{z}_{i}, \mathbf{a}_{i}) \sim \mathcal{GP}\big(\mu_{\phi,i,t-1}(\mathbf{z}_{i}, \mathbf{a}_{i}) , \sigma^2_{\phi,i,t-1}(\mathbf{z}_{i},\mathbf{a}_{i})\big)
\end{align}
Therefore, ${\widehat{u}}_i ={h}_{i, t}(\mathbf{z}_i, \mathbf{a}_i, x_i) = \frac{x_i - \mu_{\phi,i, t-1}(\mathbf{z}_i, \mathbf{a}_i)}{\sigma_{\phi,i, t-1}(\mathbf{z}_i, \mathbf{a}_i)}$. According to the definition of $h()$ in Theorem~\ref{thm:exo1}, $h()$ also follows a GP, i.e.
$h_{i,t}(\mathbf{z}_{i}, \mathbf{a}_{i}, x_{i}) \sim \mathcal{GP}(\mu_{h,i,t-1}, \sigma^2_{h,i,t-1}).$
This GP is defined by ${\phi}_{i,t}()$ which is sampled with~\eqref{eq:phi}. 
Different from $g_i()$, the observations of the input $(\mathbf{Z}_i, \mathbf{A}_i, X_i)$ for $h_i()$ are only required at the training time, and we only need to sample the learned $p(\widehat{U}_i)$ to get value $\widehat{u}_{i}$ for model prediction or model sampling.  

\subsection{Acquisition Function}
Algorithm~\ref{alg:ucb} describes the proposed  method solving~\eqref{eq:cbo_exo}. In iteration $t$, it uses GP posterior belief of $y$ to construct an upper confidence bound~(UCB~\cite{brochu2010tutorial,frazier2018tutorial}) of $y$:
\begin{align}~\label{eq:ucb}
&\text{UCB}_{t-1}(\mathbf{a}) =\mu_{t-1}(\mathbf{a}) + \beta_t \sigma_{t-1}(\mathbf{a}). 
\end{align}
Here 
$\mu_{t-1}(\mathbf{a}) = \mathbb{E}[ \mu_{g,d,t-1}({\mathbf{z}}_{d}, {\mathbf{a}}_{d}, {\widehat{u}}_{d}) ] 
 \ ; \  \  \sigma_{t-1}(\mathbf{a}) = \mathbb{E}[ \sigma_{g,d, t-1}({\mathbf{z}}_{d}, {\mathbf{a}}_{d}, {\widehat{u}}_{d}) ],$
where the expectation is taken over $p(\widehat{U})$.
In~\eqref{eq:ucb}, $\beta_t$ controls the tradeoff between exploration and exploitation of Algorithm~\ref{alg:ucb}. The UCB-based algorithm is a classic strategy that is widely used in BO and stochastic bandits~\citep{lattimore2020bandit,srinivas2009gaussian}. 
The proposed algorithm adapts
 the ``optimism in the face of uncertainty'' (OFU) strategy by taking the expectation of the UCB as part of the acquisition process.
\subsection{Algorithm}
Let $k_{g,i}, k_{\phi,i}$, $\forall i \in [d]$  represent the kernel functions of $g_i$ and $\phi_i$. 
The proposed EXCBO algorithm is summarized by Algorithm~\ref{alg:ucb}. In each iteration, a new sample is observed according to the UCB values. Then the posteriors of $\mathbf{G}$ and $\mathbf{H}$ are updated with the new dataset. The next section gives a theoretical analysis of the algorithm. 

\begin{algorithm}[ht]
\caption{EXCBO}\label{alg:ucb}
\begin{algorithmic}
\STATE {\bfseries Input:} $k_{g,i}, k_{\phi,i}$, $\forall i \in [d]$
\STATE {\bfseries Result:} Intervention actions $\mathbf{a}_i$, $\forall i \in [d]$ 

\FOR{$t=1$ {\bfseries to} $T$}
   \STATE Find $\mathbf{a}_t$ by optimizing the acquisition function, $\mathbf{a}_t \in \argmax \text{UCB}_{t-1}( \mathbf{a})$;

  \STATE  Observe samples $\{\mathbf{z}_{i,t}, x_{i,t}\}_{i=0}^d$ with the action sequence $\mathbf{a}_t$ and update $\mathcal{D}_t$;

   \STATE Use $\mathcal{D}_t$ to update posteriors  $\{\mu_{\phi,i,t}, \sigma^2_{\phi,i,t}\}_{i=0}^d$ and  exogenous surrogate $\{\widehat{u}_{i,t}\}_{i=0}^d$;
 
  \STATE Use $\mathcal{D}_t \cup \{\widehat{u}_{i,t}\}_{i=0}^d$ to update the decoder posteriors $\{\mu_{g,i,t}, \sigma^2_{g,i,t}\}_{i=0}^d$ ;
\ENDFOR
\end{algorithmic}
\end{algorithm}

\section{Regret Analysis}~\label{sec:theory}
This section describes the convergence guarantees for EXCBO using soft interventions. Our analysis shows that EXCBO has a sublinear cumulative regret bound~\citep{sussex2022model}. In DAG $\mathcal{G}$ over $\{X_i\}_{i=0}^d$, let $N$ be the maximum distance from a root to $X_d$, i.e., $N = \max_i \text{dist}(X_i, X_d)$. Here $ \text{dist}(\cdot, \cdot)$ is a measure of the edges in the longest path from $X_i$ to the reward node $Y := X_d$. Let $M$ denote the maximum number of parents of any variables in $\mathcal{G}, M = \max_i |\mathbf{pa}(i)|$.
 Let $L_t$  be a function of  $L_{g}$,  $L_{\sigma_g}$, and $N$. With Assumptions~\ref{assum:Lipschitz}-~\ref{assum:calibrated} in the Appendix, 
 the following theorem bounds the performance of EXCBO  in terms of cumulative regret.
 We present the assumptions used in the regret analysis in  Appendix~\ref{sec:regret_app}.  Assumption~\ref{assum:Lipschitz} gives the Lipschitz conditions of $g_i$, $\sigma_{g,i}$, and $\mu_{g,i}$. It holds if the  RKHS  of each $g_i$ has a Lipschitz continuous kernel~\citep{curi2020efficient,sussex2022model}. Assumption~\ref{assum:calibrated} holds when we assume that the $i$th GP prior uses the same kernel as the RKHS of $g_i$ and that $\beta_{i,t}$ is sufficiently large to ensure the confidence bounds in 
\begin{align*}
&\bigg| g_i(\mathbf{z}_{i}, \mathbf{a}_i, \widehat{u}_i) - \mu_{g,i,t-1}(\mathbf{z}_{i}, \mathbf{a}_i, \widehat{u}_i)  \bigg|   \leq  \beta_{i,t}\sigma_{g,i, t-1} (\mathbf{z}_{i}, \mathbf{a}_i, \widehat{u}_i), \\
&\forall  \mathbf{z}_i \in  \mathcal{Z}_i, \mathbf{a}_i \in  \mathcal{A}_i, \widehat{u}_i \in  \widehat{\mathcal{U}}_i.
\end{align*}
 
\begin{theorem}\label{thm:regret}
Consider the optimization problem in~\eqref{eq:cbo_exo}, with the SCM satisfying Assumptions~\ref{assum:Lipschitz}-~\ref{assum:calibrated}, where $\mathcal{G}$ is known but $\mathbf{F}$ is unknown. Then with probability at least $1-\alpha$, the cumulative regret of Algorithm~\ref{alg:ucb} is bounded by 
$R_T \leq  \mathcal{O}(L_{T} M^{N} d  \sqrt{ T \gamma_T}).$
\end{theorem}
Here $\gamma_T = \max_t \gamma_{i,T}$ denote the maximum  information gain at  time $T$.  The proof of Theorem~\ref{thm:regret} and further analysis can be found in Appendix~\ref{sec:regret_app}.

\section{Experimental Study}~\label{sec:experiment} 
This section presents experimental comparisons of the proposed EXCBO and existing algorithms. Different from the single-mode Gaussian noise in MCBO~\citep{sussex2022model},  We use two-mode exogenous distributions in the synthetic  datasets, i.e.
\begin{align}\label{eq:gm_noise}
&p(U) = w_1 \mathcal{N}(\mu_1, c_1\sigma^2) + w_2 \mathcal{N}(\mu_2, c_2\sigma^2), \\ \notag
& w_1, w_2, c_1, c_2 > 0, w_1 + w_2 =1.0.
\end{align}
Additional experimental results and analysis are presented in Appendix~\ref{sec:expriment_add}.

\begin{figure*}[h!]
\centering 
(a)  $\lambda = 1.0, \sigma = 0.1$  ~~~~~ (b) $\lambda = 2.0, \sigma = 0.1$~~~~~ (c) $\lambda = 3.0, \sigma = 0.1$ \\ 
 \vspace{-0.03in}
\includegraphics[width=0.25\textwidth]{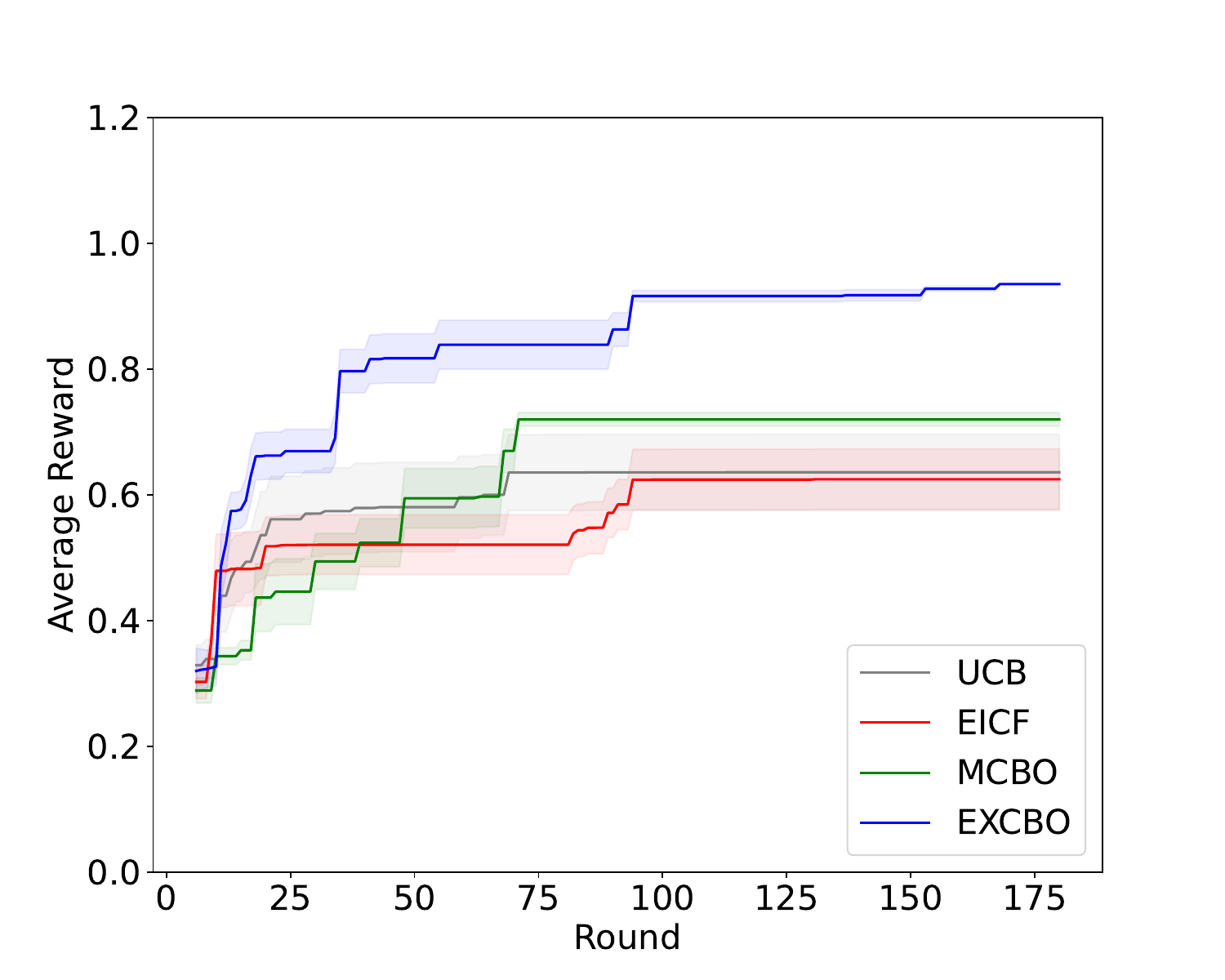} 
\includegraphics[width=0.25\textwidth]{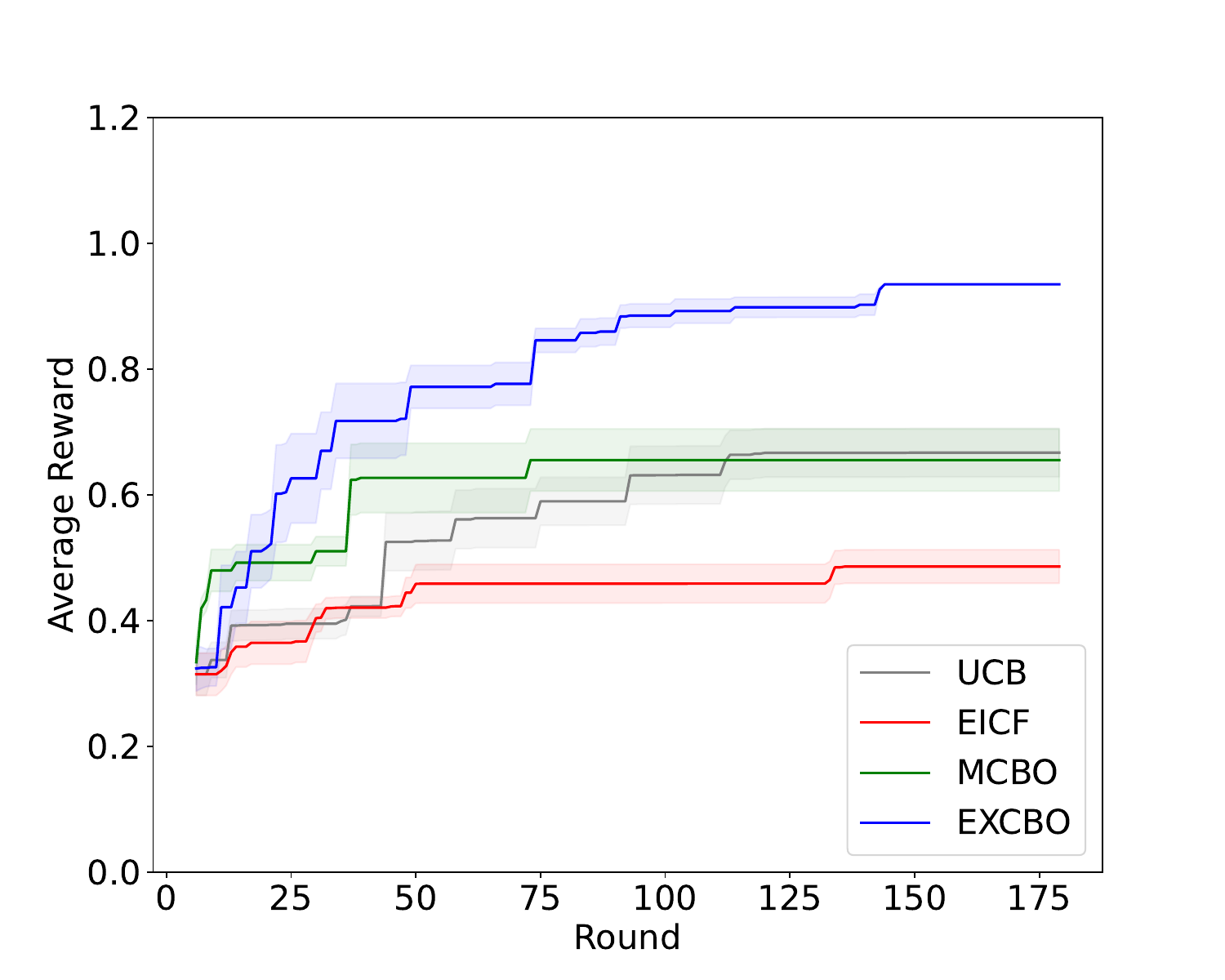}
\includegraphics[width=0.25 \textwidth]{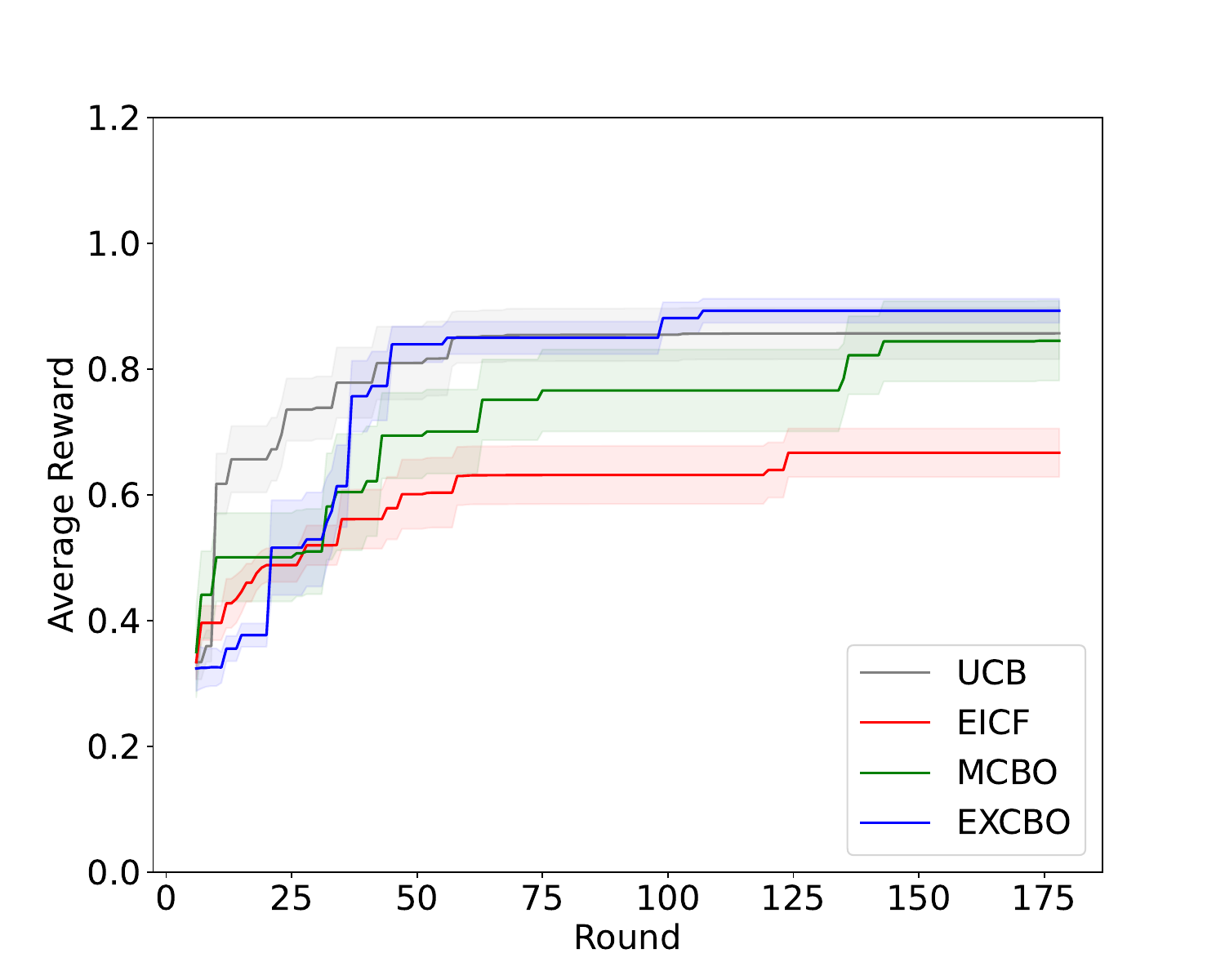}\\
(d) $\lambda = 1.0, \sigma = 0.3$ ~~~~~~~ (e) $\lambda = 2.0, \sigma = 0.3$~~~~ (f) $\lambda = 3.0, \sigma = 0.3$ \\ 
 \vspace{-0.03in}
\includegraphics[width=0.25\textwidth]{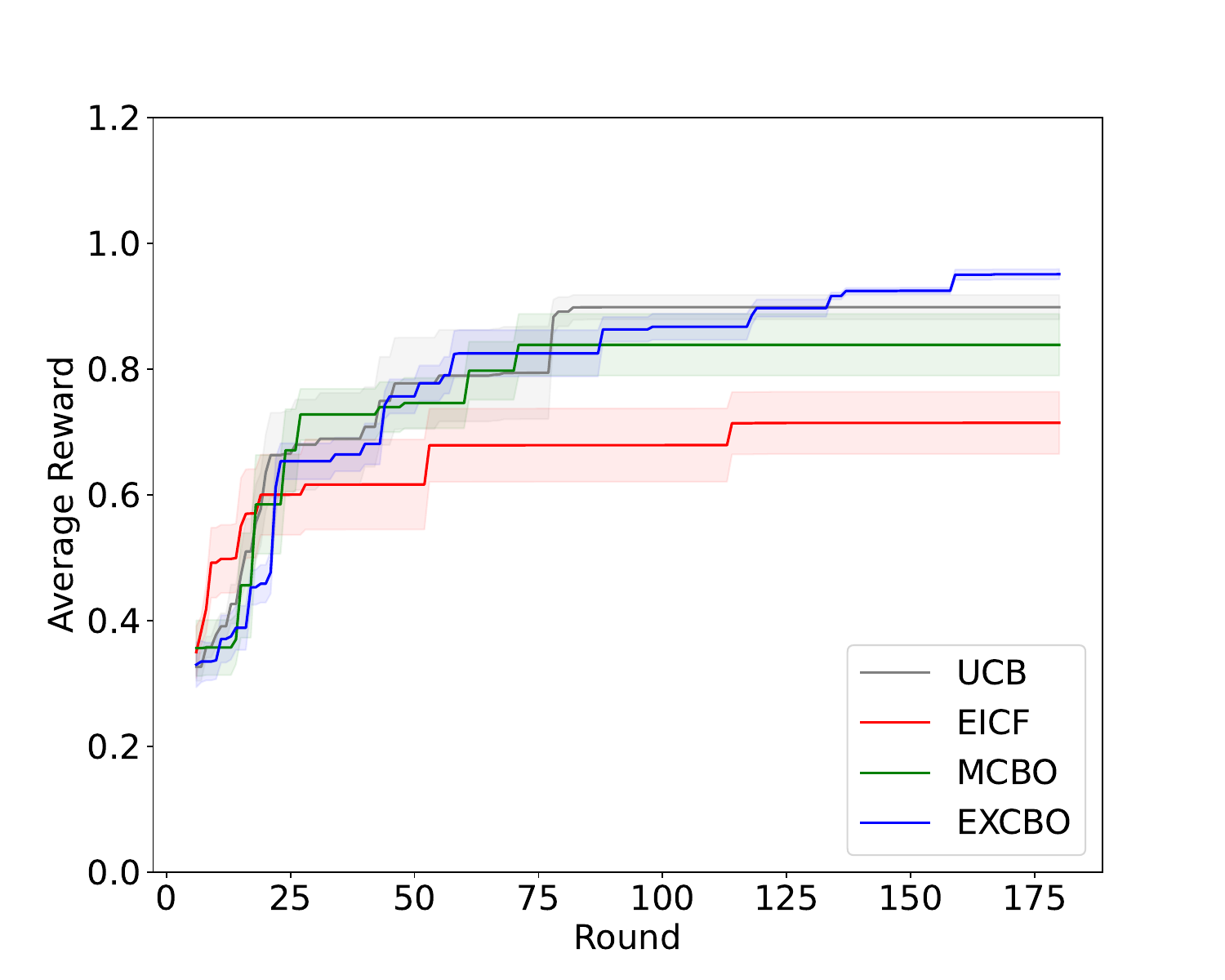} 
\includegraphics[width=0.25\textwidth]{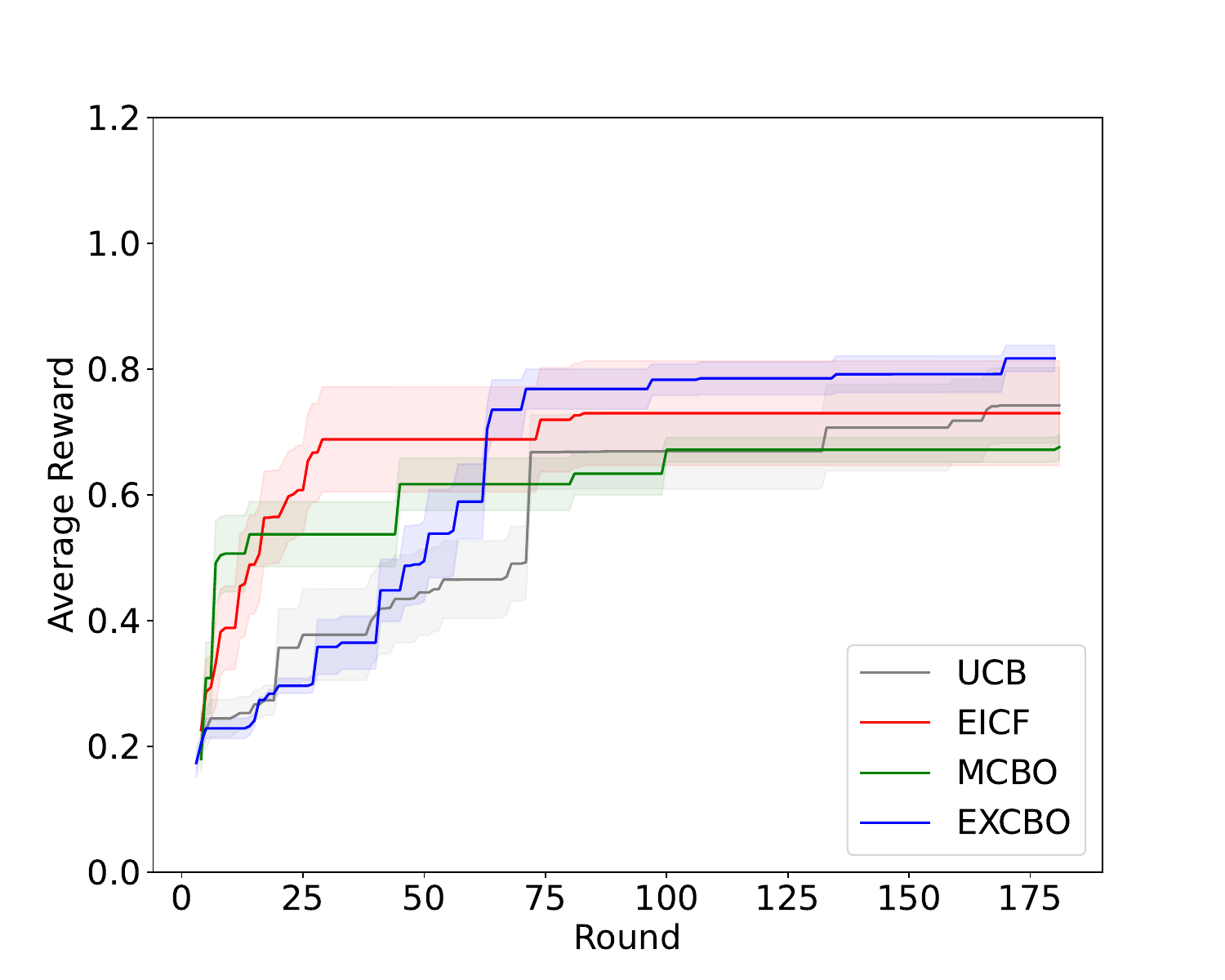}
\includegraphics[width=0.25\textwidth]{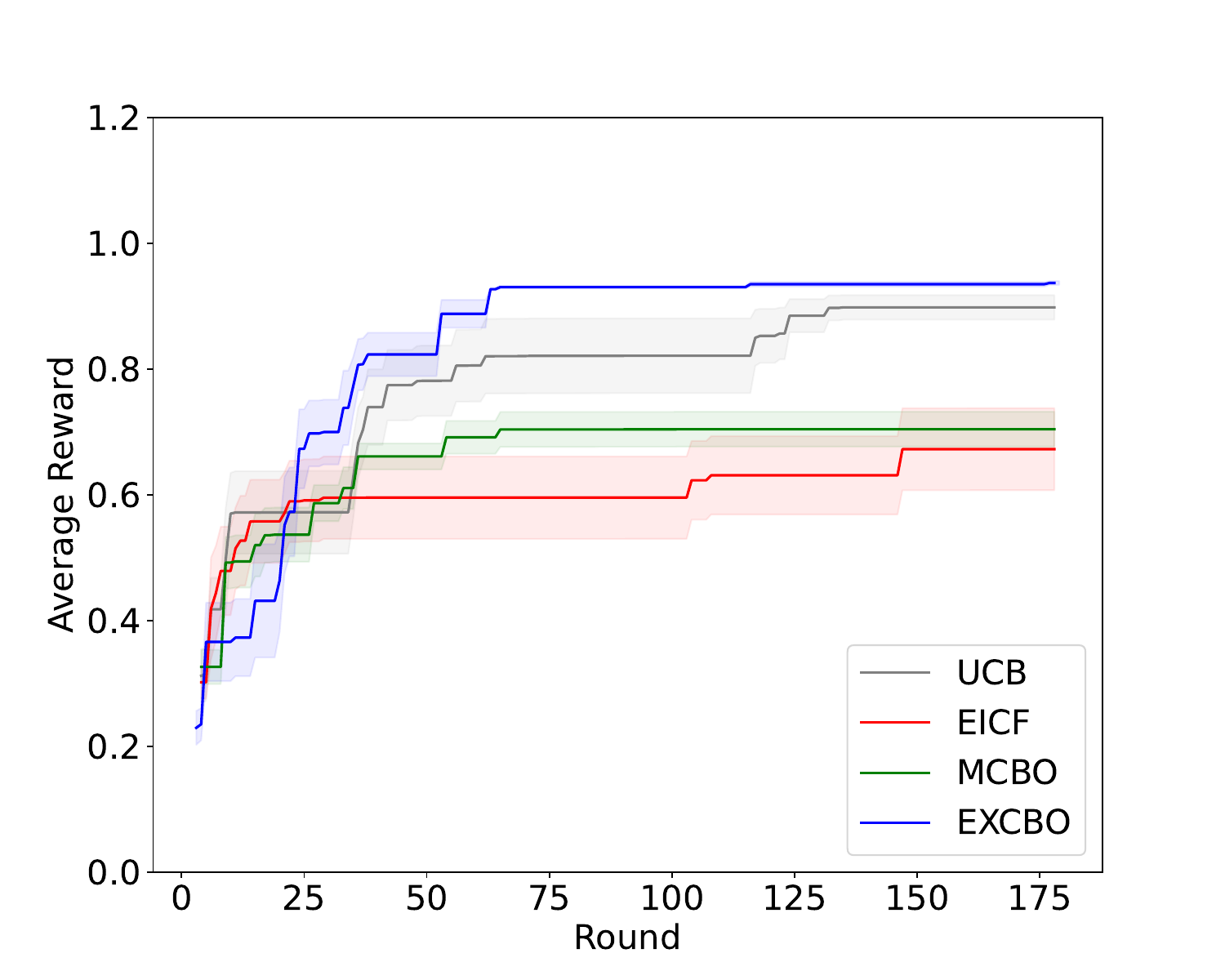}
\caption{ Results of Dropwave with $\sigma \in \{0.1, 0.3\}$ and $\lambda \in \{1.0, 2.0, 3.0\}$. }~\label{fig:dropwave_result}
\end{figure*}

\subsection{Baselines}~\label{sec:baseline}
We compare EXCBO against three representative soft-intervention-based BO algorithms: UCB~\citep{brochu2010tutorial,frazier2018tutorial}, EICF~\citep{astudillo2019bayesian}, and MCBO~\citep{sussex2022model}. UCB is a standard Bayesian Optimization (BO) method~\citep{brochu2010tutorial,frazier2018tutorial}, EICF applies a composite function approach to BO, and MCBO is a Causal Bayesian Optimization method discussed in previous sections. Unlike the other baselines, MCBO incorporates neural networks alongside GPs to capture model uncertainty. All algorithms are implemented in Python using the BoTorch library~\citep{balandat2020botorch}. In the following experiments, each algorithm is executed four times with different random seeds to compute the mean and standard deviation of the resulting reward values. 


\subsection{Dropwave}

There are two endogenous nodes in Dropwave, i.e., $X$ and the target node $Y$ (Figure~\ref{fig:Dropwave} in~\ref{sec:app:Dropwave}). There are two action nodes associated with $X$, i.e. $a_0, a_1 \in [0, 1]$. Here $X = \sqrt{(10.24 a_0- 5.12)^2+ (10.24 a_1 -5.12)^2} + \lambda U_{X}$,  and $Y = (1.0 + \cos(12.0 X))/(2.0 + 0.5X^2)+ \lambda U_{Y}$,  $U_X  \sim p(U_X)$, and $U_Y  \sim p(U_Y)$. 
We vary $\sigma$ and $\lambda$ to simulate different levels of noise. While $\sigma$ controls the variance of the exogenous variables ($U_X$ and $U_Y$), $\lambda$ scales their effect on the target variable $Y$. 
Figure~\ref{fig:dropwave_result} presents  the results under various $\sigma$ and $\lambda$ settings.  EXCBO outperforms the baselines in this set of experiments. 

\subsection{Alpine2}\label{sec:Alpine2}
\begin{figure*}[ht!]
\centering 
(a) $\lambda = 0.3, \sigma = 0.05$ ~~~~~~~~ (b) $\lambda = 0.3, \sigma = 0.2$~~~~~~~ (c) $\lambda = 0.3, \sigma = 0.4$ \\ 
 \vspace{-0.03in}
\includegraphics[width=0.25\textwidth]{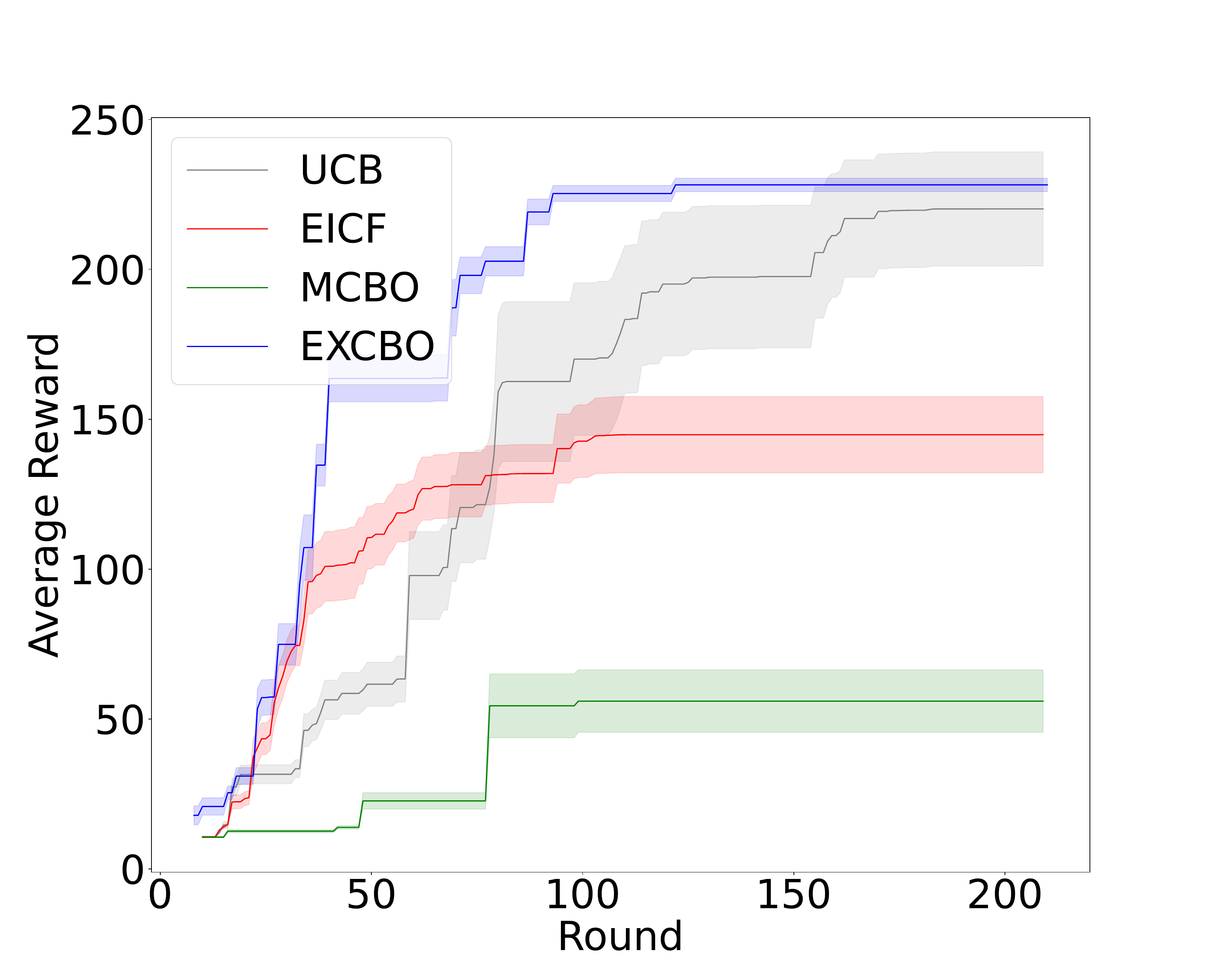} 
\includegraphics[width=0.25\textwidth]{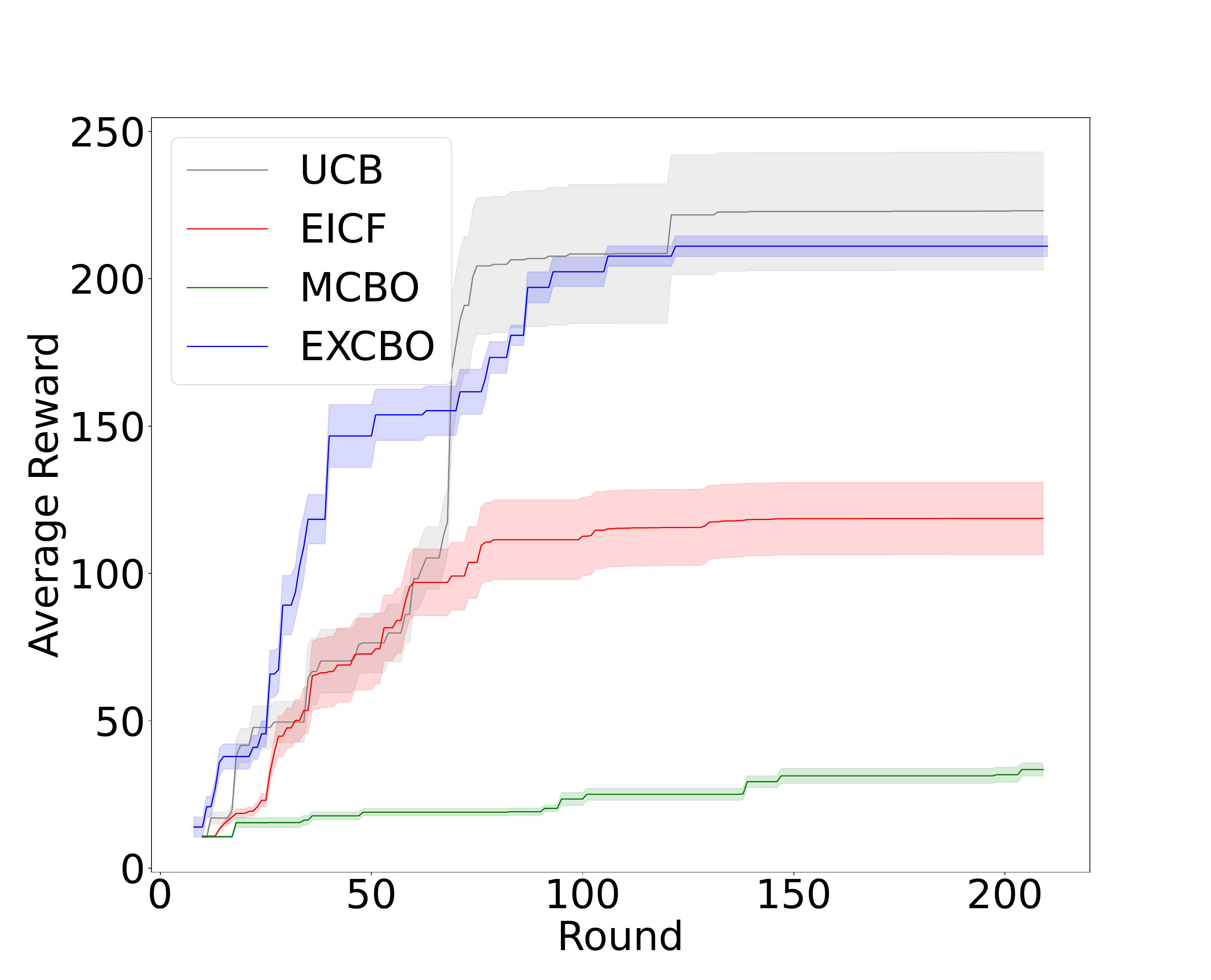}
\includegraphics[width=0.25\textwidth]{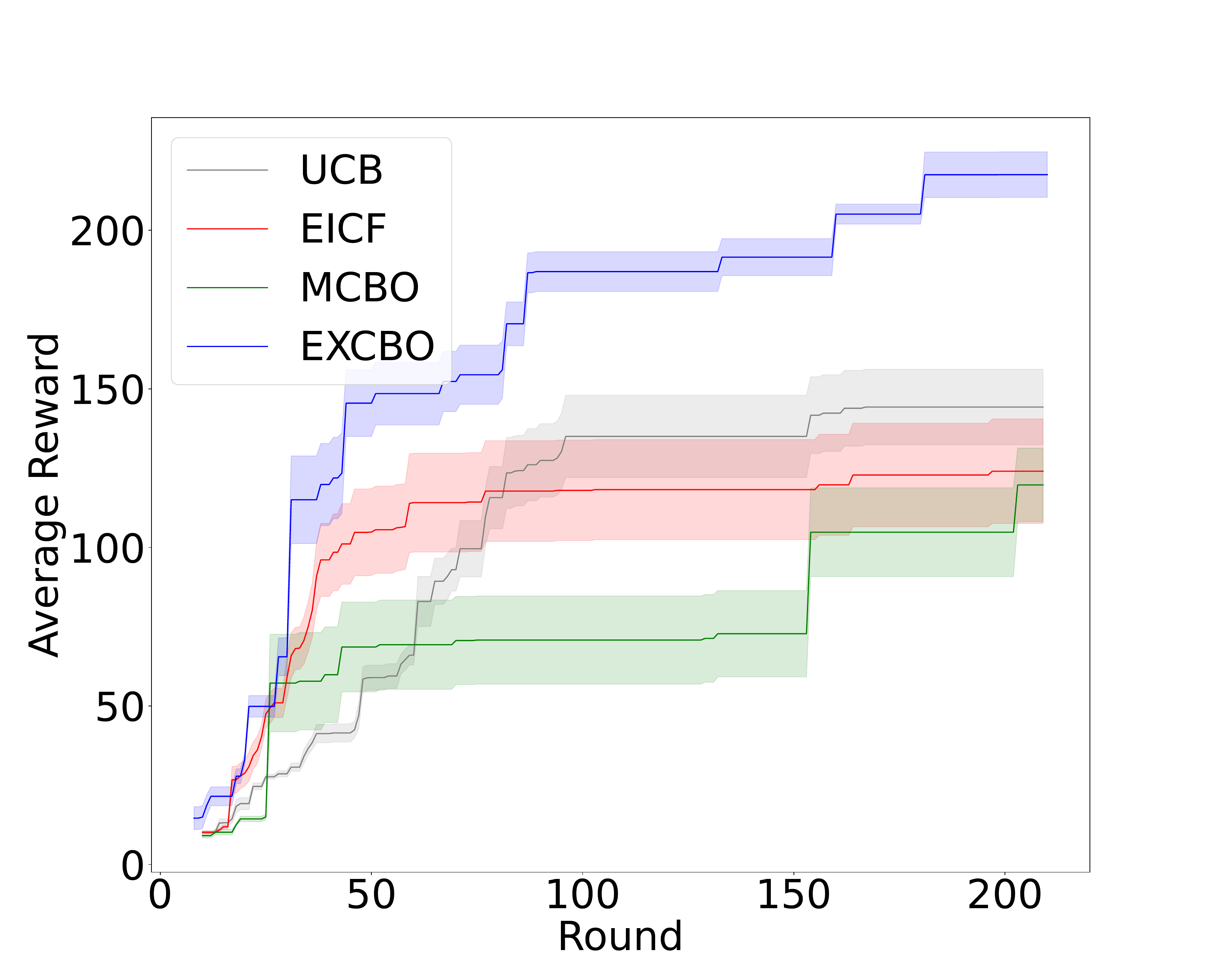}\\
(d) $\lambda = 1.0, \sigma = 0.05$ ~~~~~ (e) $\lambda = 1.0, \sigma = 0.2$~~~~~~~ (f) $\lambda = 1.0, \sigma = 0.4$ \\ 
 \vspace{-0.03in}
\includegraphics[width=0.25\textwidth]{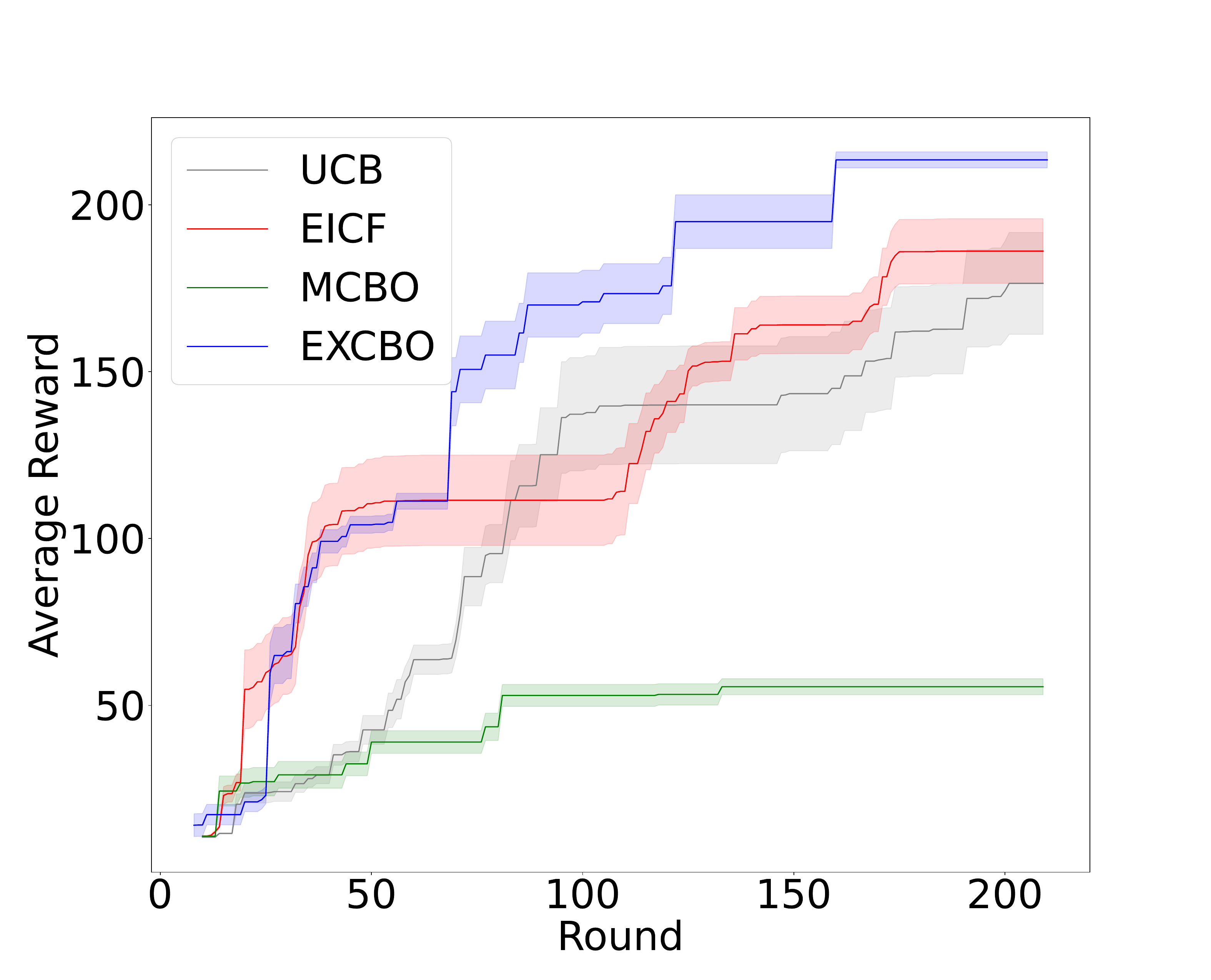} 
\includegraphics[width=0.25\textwidth]{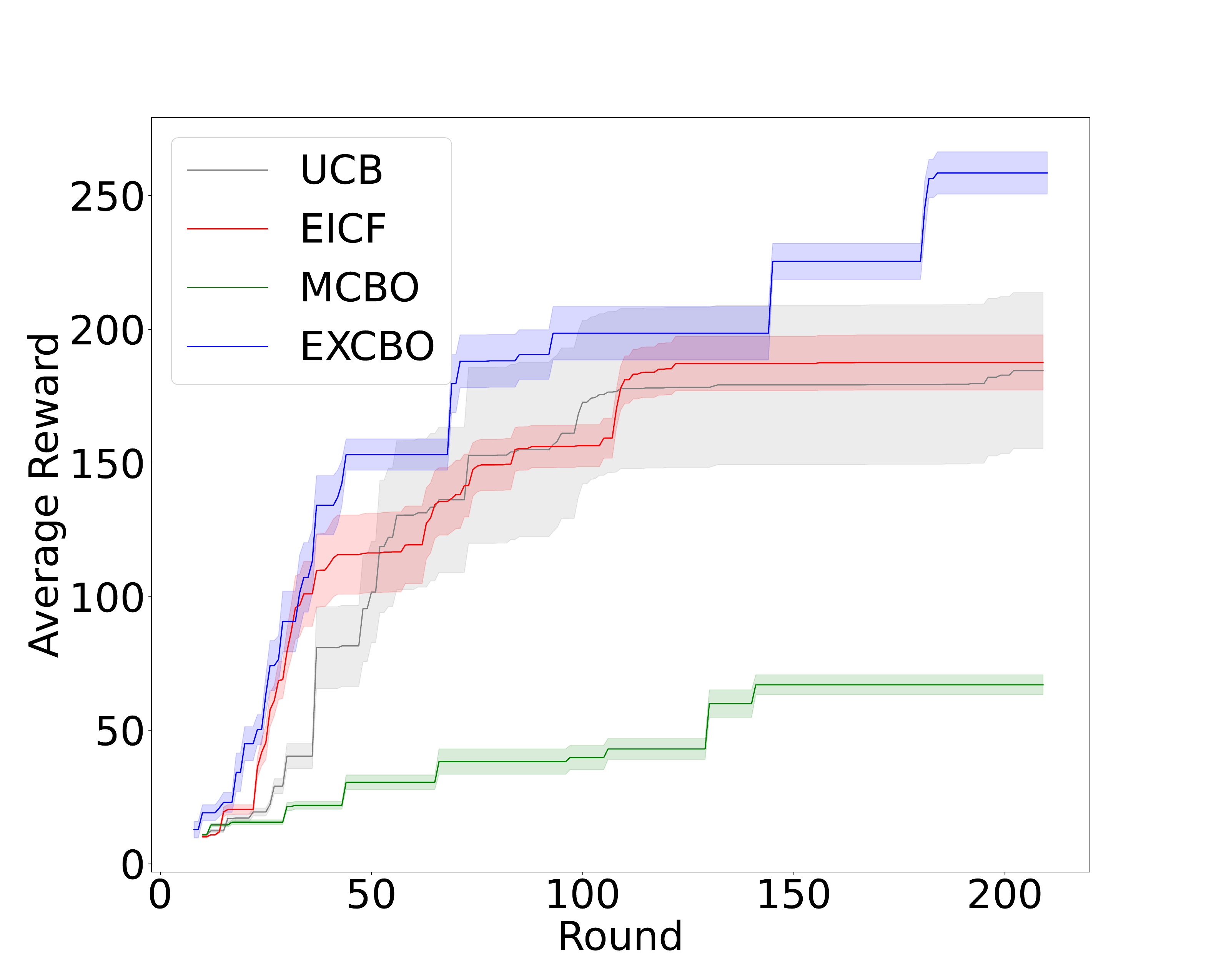}
\includegraphics[width=0.25\textwidth]{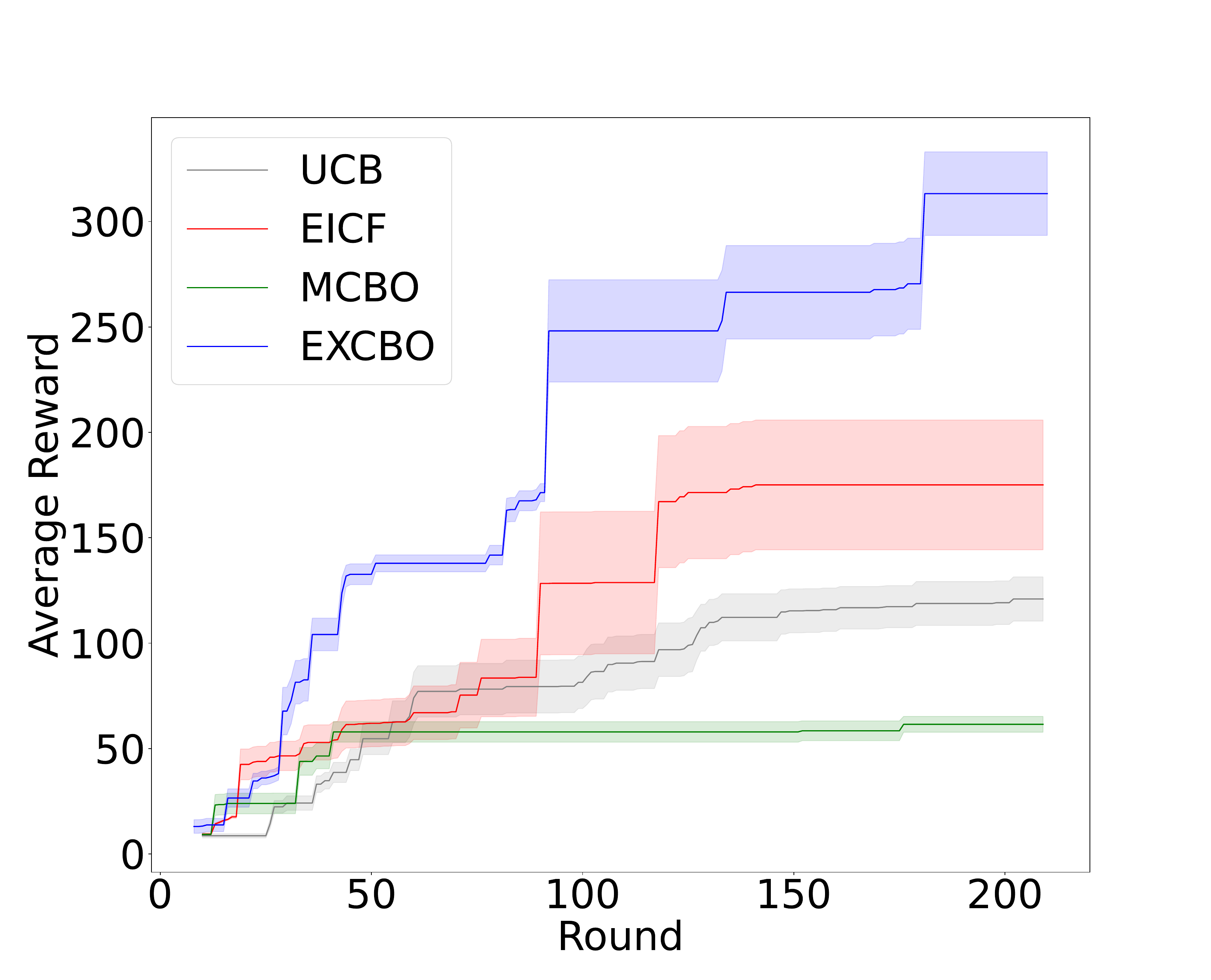}
 \vspace{-0.1in}
\caption{ Results of Alpine2 with $\sigma \in \{0.05, 0.2, 0.4\}$ and $\lambda \in \{0.3, 1.0\}$. }~\label{fig:alpine2B}
 \vspace{-0.1in}
\end{figure*}
We study the algorithms using the Alpine2 dataset~\citep{sussex2022model}. There are six endogenous nodes in the Alpine2 dataset as shown in Figure~\ref{fig:Alpine2_graph}. 
In first set of experiments, Alpine2 is generated via DGM with multimodal exogenous distributions as given in~\eqref{eq:Alpine2_DGM} in Section~\ref{sec:app:alphine2}. The results of Alpine2 are shown in Figures~\ref{fig:alpine2B}. We also compared the algorithms on Non-DGM generated Alpine2 dataset in Section~\ref{sec:app:Alpine2_NonDGM}.
As shown in the plots, our EXCBO gives better results than the other methods at different noise levels. It demonstrates effectiveness and benefits of the proposed EXCBO method in multimodal exogenous distribution and mechanism learning.

\subsection{Epidemic Model Calibration}~\label{sec:epidemic}
\begin{figure*}[h!]
\vspace{-0.05in}
\centering 
 ~~~~(a) $\sigma = 0.1$ ~~~~~~~~~~~~~~~~ (b) $\sigma = 0.3$~~~~~~~~~~~~~~~~~~~~~~(c)~~~~~~~~~~~~~~~~~~ \\ 
\includegraphics[width=0.25\textwidth]{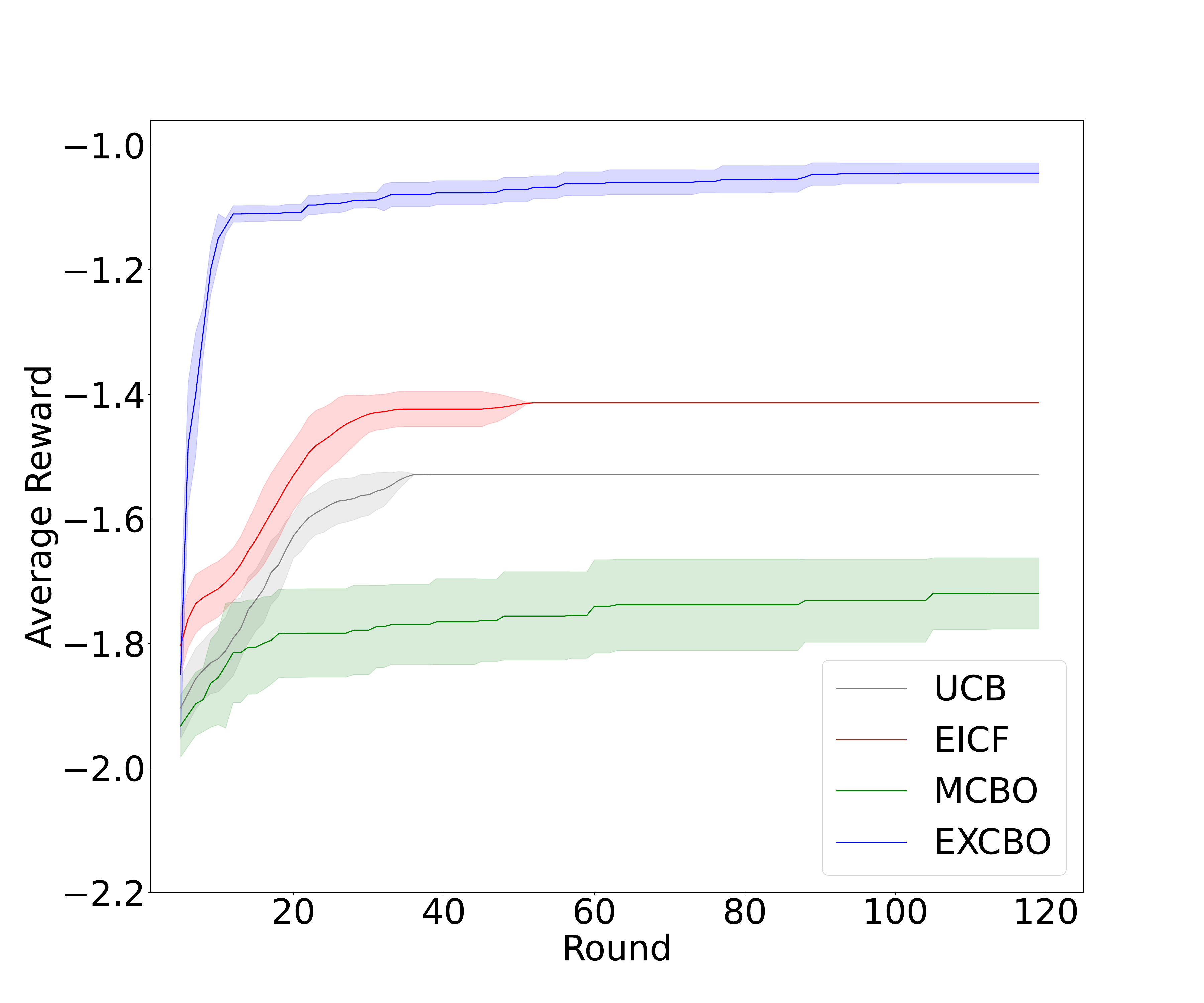} 
\includegraphics[width=0.25\textwidth]{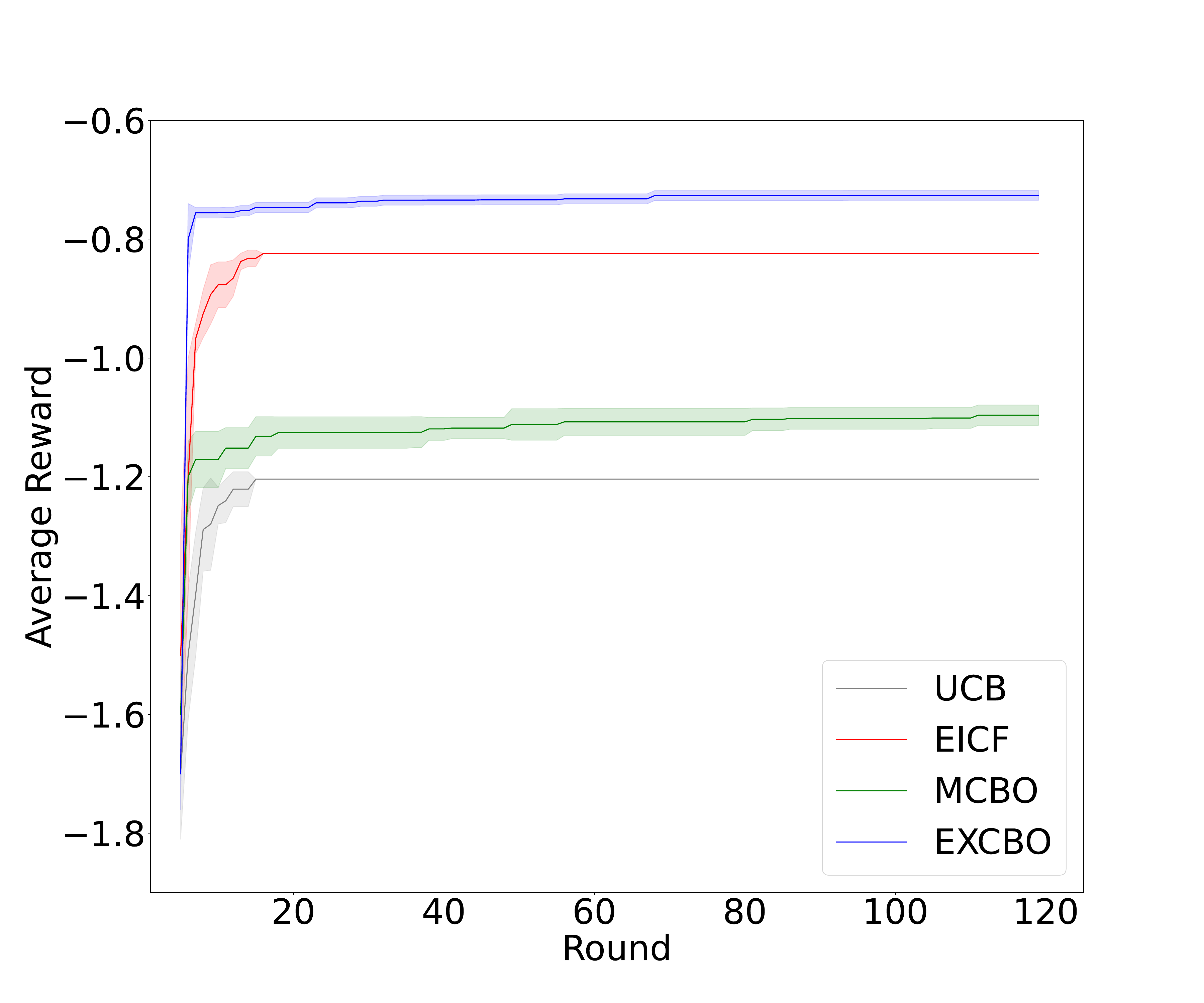}
\includegraphics[width=0.25\textwidth]{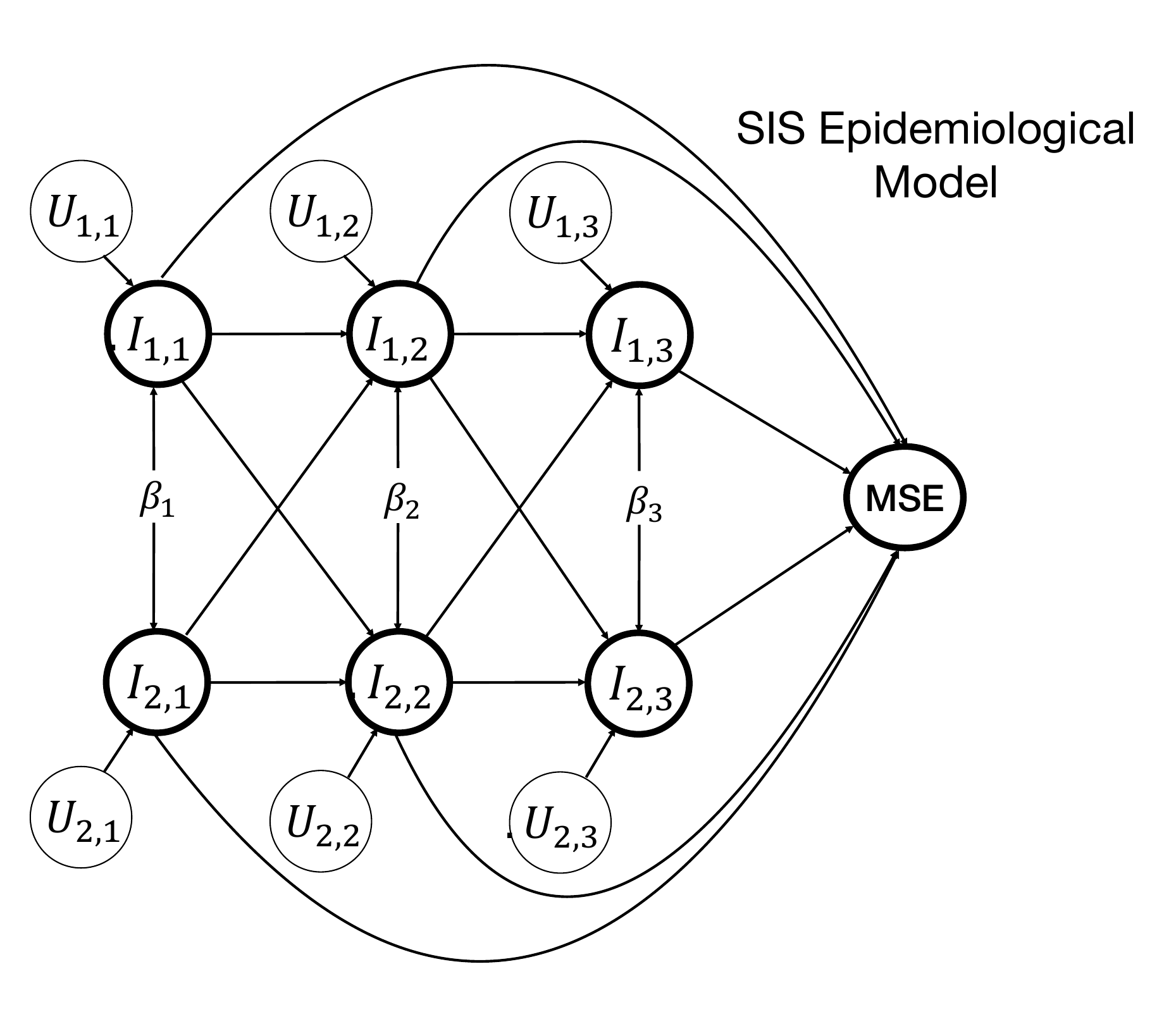}
\caption{ (a-b): Results of epidemic model calibration; (c): Graph structure for epidemic model calibration. } \label{fig:epidemic}
\end{figure*}

We test EXCBO on an epidemic model calibration by following the setup in~\cite{astudillo2021bayesian}. In this model, as shown in Figure~\ref{fig:epidemic}-(c), $I_{i,t}$ represents the fraction of the population in group $i$ that are ``infectious'' at time $t$; $\beta_{i,j,t}$ is the rate of the people from group $i$ who are ``susceptible'' have close physical contact with people in group $j$ who are ``infectious'' at time $t$. We assume there are two groups, and infections resolve at a rate of $\gamma$ per period.  The number of infectious individuals in group $i$ at the start of the next time period is $I_{i,t+1} = I_{i,t}(1-\gamma) + (1-I_{i,t})\sum_{j}\beta_{i,j,t}I_{j,t}$. We assume each $I_{i,t}$ has an observation noise $U_{i,t}$. 
The model calibration problem is that given limited noisy observations of $I_{i,t}$s, how to efficiently find the $\beta_{i,j,t}$ values in the model. The reward is defined as the negative mean square error~(MSE) of all the $I_{i,t}$ observations as the objective function to optimize. In this model, $\beta_{i,j,t}$s are the action variables.  The noise is added with two-mode as in~\eqref{eq:gm_noise} under ANM~\citep{hoyer2008nonlinear}. Figure~\ref{fig:epidemic}-(a-b) visualize the results at the noise levels with $\sigma = 0.1$ and $\sigma = 0.3$. 

\subsection{Planktonic Predator–prey Community in a Chemostat}

\begin{figure}[h!]
\vspace{-0.1in}
\centering 
\includegraphics[width=0.35\textwidth]{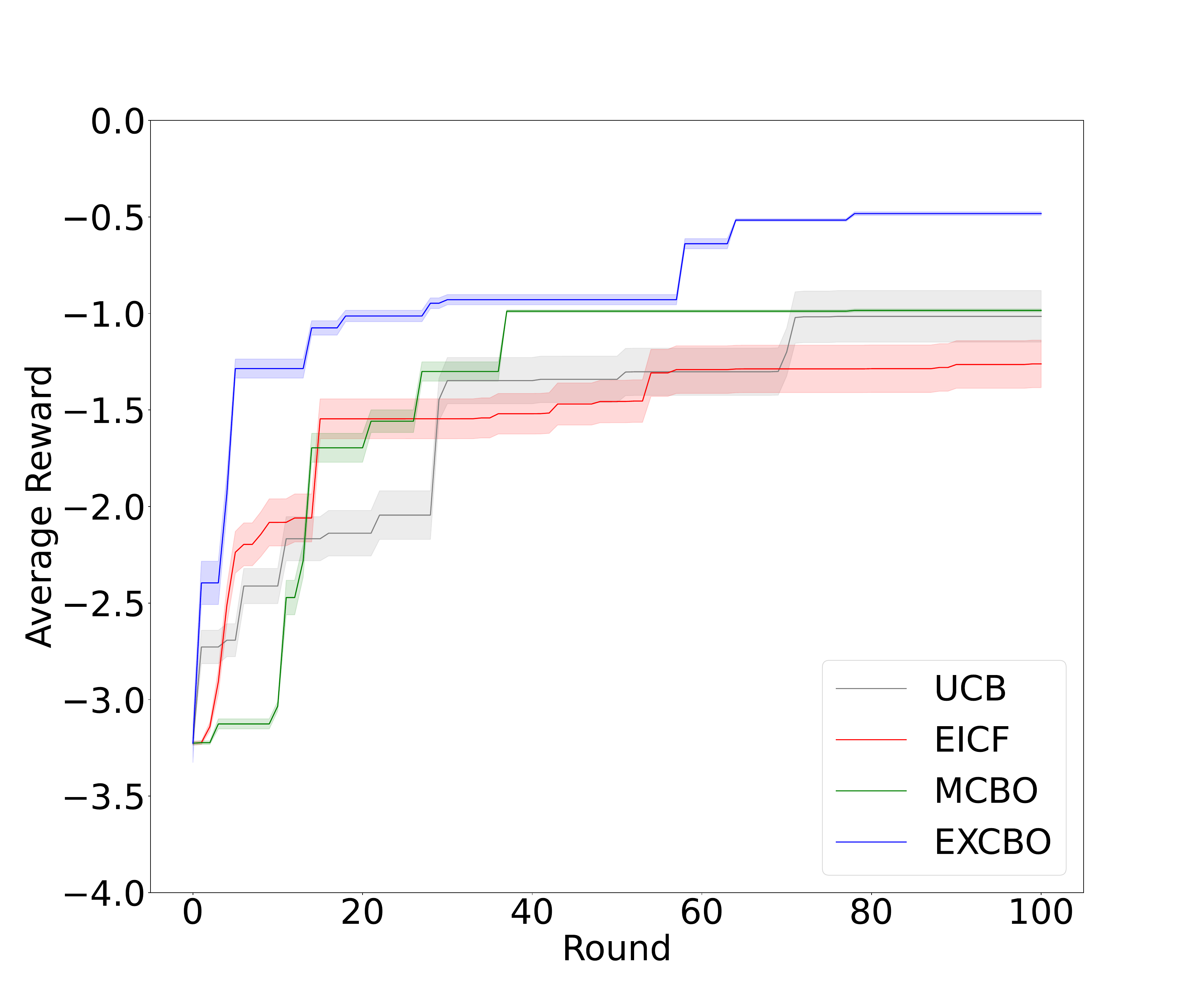}
\vspace{-0.1in}
\caption{Results of P3C$^2$ dataset; the reward $y = -D_t$.} \label{fig:p3c2}
\end{figure}

\begin{figure}[h]
\centering 
\includegraphics[width=0.3\textwidth]{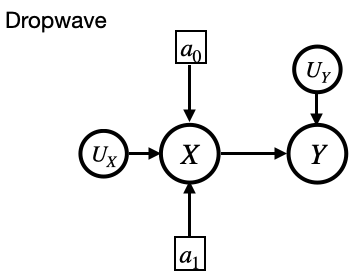}
\caption{Graph structure of Dropwave dataset.} \label{fig:Dropwave}
\end{figure}

We evaluate the algorithms on a \emph{real-world} dataset from the \textbf{p}lanktonic \textbf{p}redator–\textbf{p}rey \textbf{c}ommunity in a \textbf{c}hemostat (P3C$^2$). This biological system involves two interacting species, one predator and one prey, and our objective is to identify interventions that reduce the concentration of dead animals in the chemostat, $D_t$. We adopt the system of ordinary differential equations (ODE) from~\cite{blasius2020long,aglietti2021dynamic} as the SCM, and construct the DAG by unrolling the temporal dependencies of two adjacent time steps. Observational data from~\cite{blasius2020long} are used to compute the dynamic causal prior. Unlike dynamic sequential CBO~\citep{aglietti2021dynamic}, we employ the causal structure at $t$ and $t+1$ as the DAG for the algorithms. Figure~\ref{fig:p3c2} compares the performance of EXCBO with baselines. Additional experimental details are provided in the Appendix.

\section{Conclusions}~\label{sec:conclude}
We propose a novel CBO algorithm, EXCBO, that approximately recovers the exogenous variables in a structured causal model.  
With the recovered exogenous distribution, our method naturally improves the surrogate model's accuracy in the approximation of the SCM. Furthermore, the recovered exogenous variables may enhance the surrogate model's capability in causal inference and hence improve the reward values attained by EXCBO. We additionally provide theoretical analysis on both exogenous variable recovery and the algorithm's cumulative regret bound. Experiments on multiple datasets show the algorithm's soundness and benefits. 

\clearpage
\section*{Impact Statement}\label{sec:impact}

As a new causal Bayesian optimization framework, EXCBO may help reduce the required training samples for more efficient and cost-effective decision-making, which may have broader impacts in many science and engineering applications, such as future pandemic preparedness with better-calibrated epidemic dynamic models as illustrated in the paper. There are many potential societal consequences of our work, none of which we feel must be specifically highlighted here.


\bibliographystyle{plainnat}
\bibliography{main}

\section{Additional Remarks}

\subsection{Remarks on Motivations}

Learning the exogenous distribution enhances the surrogate model’s ability to approximate the ground truth SCMs. As discussed in Sections~\ref{sec:onenode}, \ref{sec:proof_thm}, and~\ref{sec:edl_app}, under moderate assumptions, the independence between the recovered exogenous variable $\widehat{U}$ and both the parents $\mathbf{Z}$ and  actions $\mathbf{A}$ empowers the structured surrogate model in EXCBO to be \emph{counterfactually identifiable} for effective interventions. This independence reduces the influence of environmental noise or exogenous variables on the actions or interventions derived from the acquisition function.

This work considers the setting where the causal structure is known, and the model $\mathcal{M}$ is causally sufficient. The challenges of learning causal structures and dealing with unobserved confounders are left for future research.

We believe multi-modal and non-Gaussian exogenous distributions are prevalent in real-world systems. When each exogenous variable is viewed as an unobserved latent factor, it is highly plausible that such factors follow non-Gaussian distributions with multiple modes. 

\subsection{Performance Gaps}

 UCB, EICF, and MCBO use $X = \widehat{f}(\mathbf{Z}, \mathbf{A})$ or $X = \widehat{f}(\mathbf{Z}, \mathbf{A}, \epsilon), \epsilon \in \mathcal{N}(0,1)$  to approximate  $X = f(\mathbf{Z}, \mathbf{A}, U)$ for each node or the overall reward function. The absence of information about $U$ introduces irreducible bias into the surrogate model of the reward function.   In contrast, EXCBO explicitly recovers the exogenous variable $U$ and learns its multi-modal distribution, producing a more accurate surrogate, i.e., $X = \widehat{f}(\mathbf{Z}, \mathbf{A}, \widehat{U})$,  for the objective reward function, even when the variance $\sigma^2$ in the data is small. Experimental results further show that EXCBO enhances the robustness of CBO, particularly in scenarios with limited data samples.


\section{Nomenclature}

\begin{table}[H]
\begin{center}
\begin{tabular}{  c|l } 
 \hline
Symbol  & Description \\ 
\hline
$U$ & a single exogenous variable \\ 
$\mathbf{U}$ & the exogenous variable set of a SCM, i.e., $\mathbf{U}=\{U_1, ..., U_d\}$ \\ 
$\widehat{U}$ & the exogenous variable recovered via EDS, i.e., the  EDS surrogate of $U$\\ 
$u$ & a value or realization of variable $U$\\ 
$\widehat{u}$ & a value or realization of variable $\widehat{U}$\\
$\mathcal{U}$ &  the domain, or value space of variable $U$ \\
$\widehat{\mathcal{U}}$ & the domain, or value space of variable $\widehat{U}$\\
$X_i$ &  endogenous variable $i$; node $i$ \\
$\mathbf{A}_i$ & action variable of node $i$\\
$\mathbf{a}_i$ & a value of action variable $\mathbf{A}_i$\\
$h_i()$ & the EDS encoder function for node $X_i$\\
$\phi()$ & a regression model from $\mathbf{Z}$ to $X$, being used to construct encoder $h()$\\
$\widehat{u}_i$ &  the output of  $h_i()$ given an input \\
$\tilde{h}_i()$ & a plausible  function of $h_i$ via  posterior  GP  trained  with data in some step of EXCBO \\
$\tilde{\widehat{u}}_i$ &  the output of  $\tilde{h}_i()$ given an input \\
$\mathbf{Z}_i$ &  the parent of $X_i$, i.e. $\mathbf{pa}(i)$ \\
$g_i()$ & the EDS decoder function of node $X_i$\\
$\mathcal{G}$ & the causal graph (DAG) of an SCM\\
$\mathbf{V}$ & the endogenous set of an SCM, $\mathbf{V} = \{X_i\}_{i=0}^d$\\
$\mathbf{F}$ & the mechanism set of an SCM, $\mathbf{F} = \{f_i\}_{i=0}^d$\\
$\mathcal{M}$ & an SCM \\
$\mathbf{I}$ & an intervention target set\\
$\mathbf{G}$ & the collection of decoders, $\mathbf{G} = \{g_i\}_{i=0}^d$ \\
$\mathbf{H}$ & the collection of encoders, $\mathbf{H} = \{h_i\}_{i=0}^d$ \\
 \hline
\end{tabular}
\end{center}
\end{table}


\section{Additional Experimental Results and Analysis}~\label{sec:expriment_add}

In our experiments, the synthetic data are generated via  DGM,  Non-DGM, and ANM mechanisms. EXCBO and all the baselines are soft intervention methods; therefore, the hard intervention CBO method~\citep{aglietti2020causal} is not included in the experiments.

\subsection{Experimental Setup}

We report the expected reward, $\mathbb{E}_U[y \mid \mathbf{a}_t]$, as a function of the number of system interventions performed. Each figure presents the mean performance over four random seeds, with error bars representing the interval $[-0.2 \sigma, 0.2 \sigma]$. The GPs used in our models are implemented via the \texttt{SingleTaskGP()} function from BoTorch~\citep{balandat2020botorch}, and are trained using the default hyperparameters described in~\cite{hvarfner2024vanilla}. Most of the synthetic datasets are generated using Gaussian Mixture Models (GMMs) with two components.  Action node domains are normalized to lie within $[0,1]$. To reduce computational overhead, we restrict the number of $\sigma$ values considered for the exogenous variables in each dataset.

\subsection{Dropwave}\label{sec:app:Dropwave}
In Dropwave Dataset, the values of action nodes $a_0, a_1 \in [0, 1]$, $X = \sqrt{(10.24 a_0- 5.12)^2+ (10.24 a_1 -5.12)^2} + \lambda U_{X}$,  and $Y = (1.0 + \cos(12.0 X))/(2.0 + 0.5X^2)+ \lambda U_{Y}$,  $U_X  \sim p(U_X)$, and $U_Y  \sim p(U_Y)$. Here $p(U_X)= 0.5 \mathcal{N}(-0.2,, 1.4\sigma^2) + 0.5\mathcal{N}(0.4, \sigma^2)$, and $p(U_Y)= 0.5 \mathcal{N}(-0.1, 0.32\sigma^2) + 0.5\mathcal{N}(0.05, 0.32\sigma^2)$. Clearly, the data generation here belongs to the  ANMs~\citep{hoyer2008nonlinear}. 

As shown in the plots, UCB’s performance improves with increasing $\sigma$ or $\lambda$, suggesting that strong exogenous noise may diminish the benefits of structural knowledge utilized by EICF and EXCBO. Nevertheless, EXCBO  achieves superior  performance under different settings as shown in Figure~\ref{fig:dropwave_result}.

\subsection{Alpine2}~\label{sec:app:alphine2} 

\begin{figure}[H]
\centering
\includegraphics[width=0.3\textwidth]{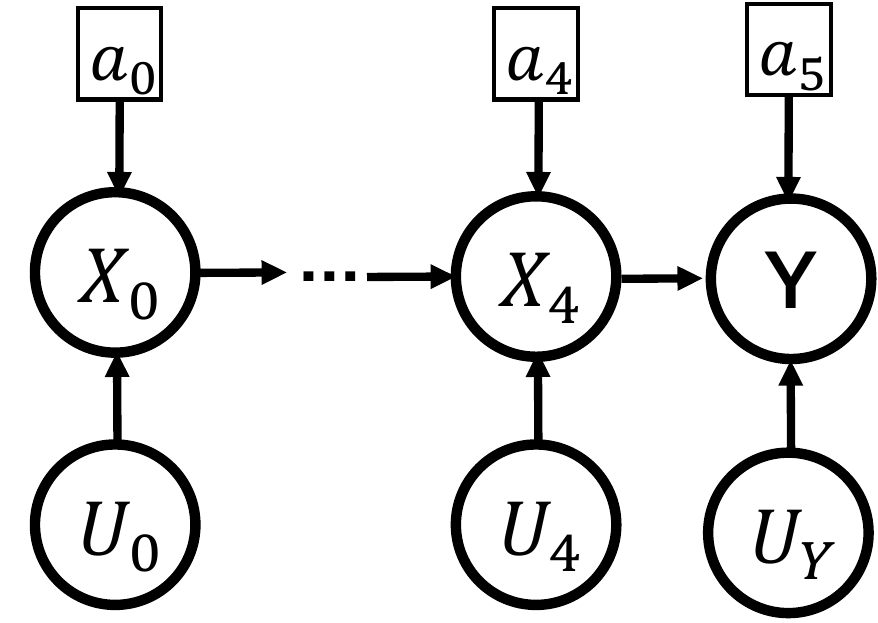}
\caption{Graph structure of the Alpine2 dataset.} \label{fig:Alpine2_graph}
\end{figure}

The Alpine2 dataset contains six endogenous nodes, as illustrated in Figure~\ref{fig:Alpine2_graph}. The exogenous distributions for $X$ and $Y$ follow Gaussian Mixture models with two components, as defined in equation~\eqref{eq:gm_noise}. 
Due to the high computational cost of evaluating MCBO~\citep{sussex2022model}, we restrict our comparisons in this experiment to UCB~\citep{brochu2010tutorial,frazier2018tutorial} and EICF~\citep{astudillo2019bayesian}.

\subsubsection{Alpine2 with DGM Mechanism}\label{sec:Alpine2_DGM}

We evaluate the algorithms on the synthetic Alpine2 dataset~\citep{sussex2022model}, generated using a DGM mechanism with multimodal exogenous distributions. Each node is defined as 
\begin{align}\label{eq:Alpine2_DGM}
&X_0 = -\sqrt{10.0a_{0} } \sin{(10.0a_{0} )} + \big(\cos(10.0a_{0} ) + 1.2\big)\cdot\lambda U_0^4; \\\notag
&X_i = \sqrt{10.0a_{i} } \sin{(10.0a_{i} )} X_{i-1} + 0.1\big(\cos(10.0a_{i} ) + X_{i-1}^2 + 1.2\big)\cdot\lambda U_i^4, \  1 \leq i \leq 5; \\\notag
&Y = X_5.
\end{align}
Here, $U_i \sim p(U_i)$ as specified in~\eqref{eq:gm_noise}, with $w_1 = w_2 = 0.5$, $\mu \in [-1.0, 1.0]$, and $c_1, c_2 \in [0.05, 1.5]$. 
The results for $\sigma \in \{0.05, 0.2, 0.4\}$ and $\lambda \in \{0.3, 1.0\}$ are shown in Figure~\ref{fig:alpine2B}.  

\subsection{Alpine2 with Non-DGM Mechanism}\label{sec:app:Alpine2_NonDGM}

For the non-DGM setting of the Alpine2 dataset~\citep{sussex2022model}, each node is defined as 
\begin{align}\label{eq:Alpine2_complex}
&X_0 = -\sqrt{10.0a_{0} + U_{0}} \sin{(10.0a_{0} + U_{0})}; \\\notag
&X_i = \sqrt{10.0a_{i} + U_{i}} \sin{(10.0a_{i} + U_{i})} X_{i-1}, \  1 \leq i \leq 5; \\\notag
&Y = X_5.
\end{align}
Here, $U_i \sim p(U_i)$ as defined in~\eqref{eq:gm_noise}, with $w_1 = w_2 = 0.5$, $\mu \in [-1.0, 1.0]$, and $c_1, c_2 \in [0.05, 1.5]$.  

\begin{figure}[H]
\centering
(a) $\sigma = 0.05$ ~~~~~~~~~~~~~~~~~~~~~~ (b) $\sigma = 0.1$ ~~~~~~~~~~~~~~~~~~~~ (c) $\sigma = 0.2$ \\
\includegraphics[width=0.32\textwidth]{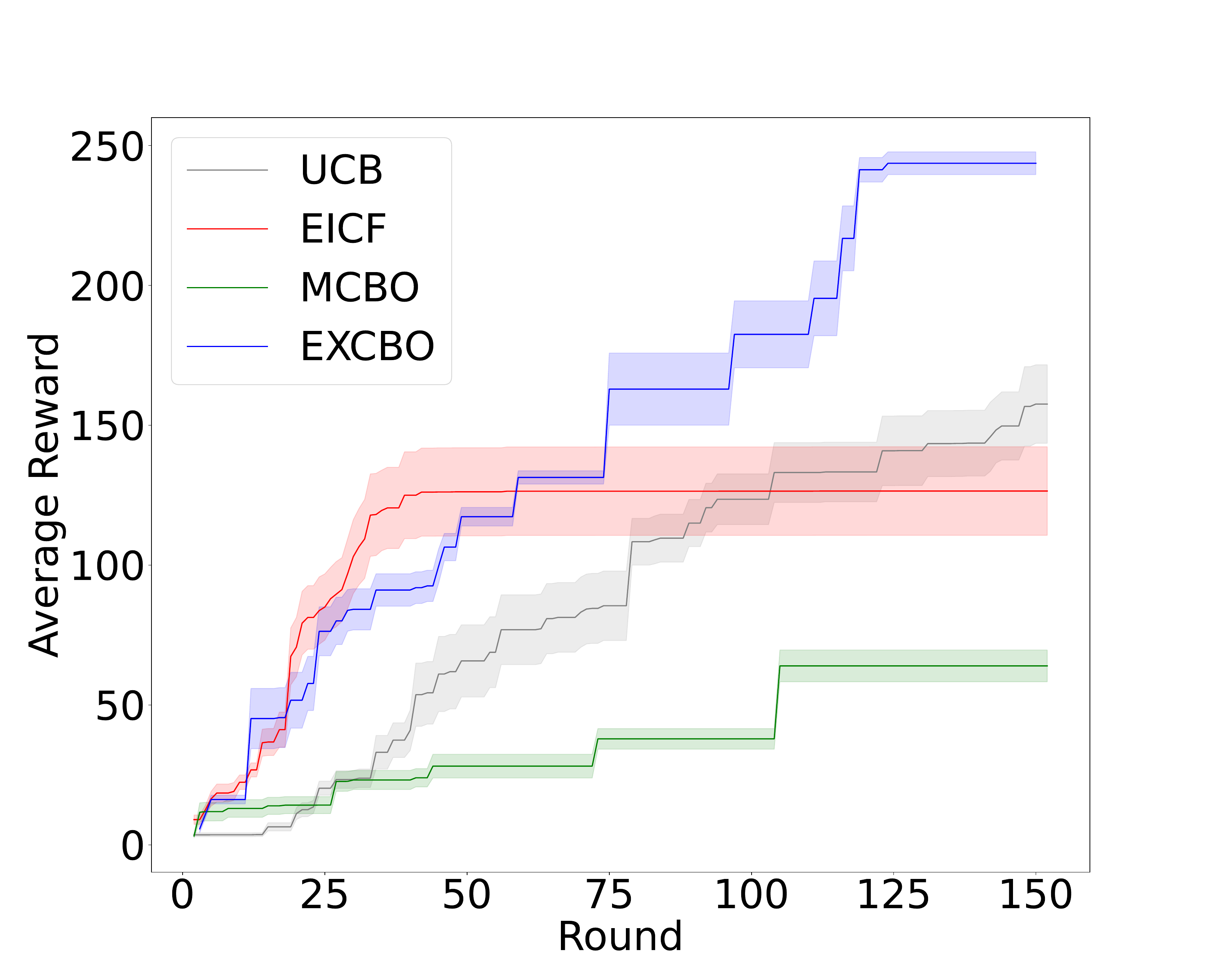} 
\includegraphics[width=0.32\textwidth]{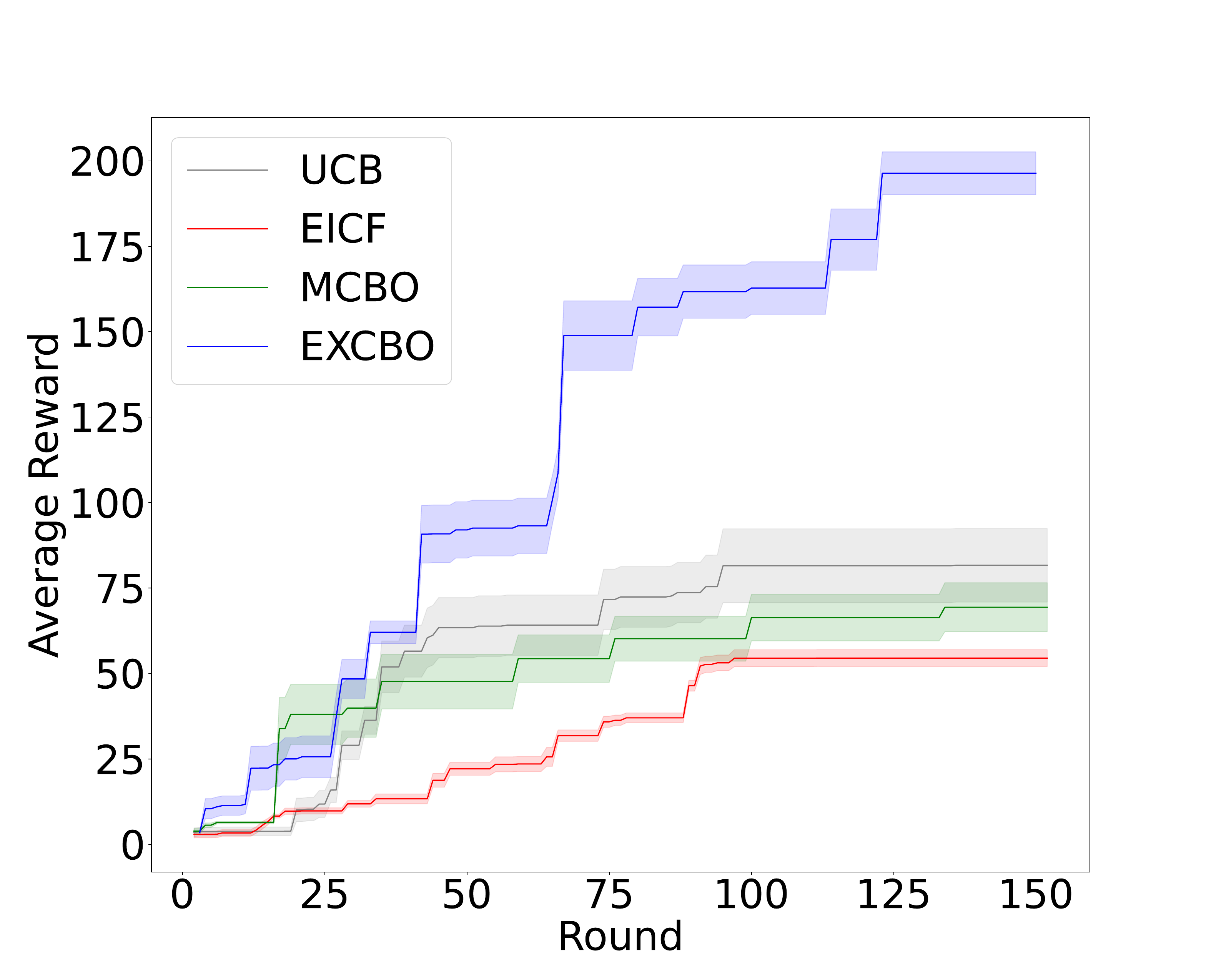} 
\includegraphics[width=0.32\textwidth]{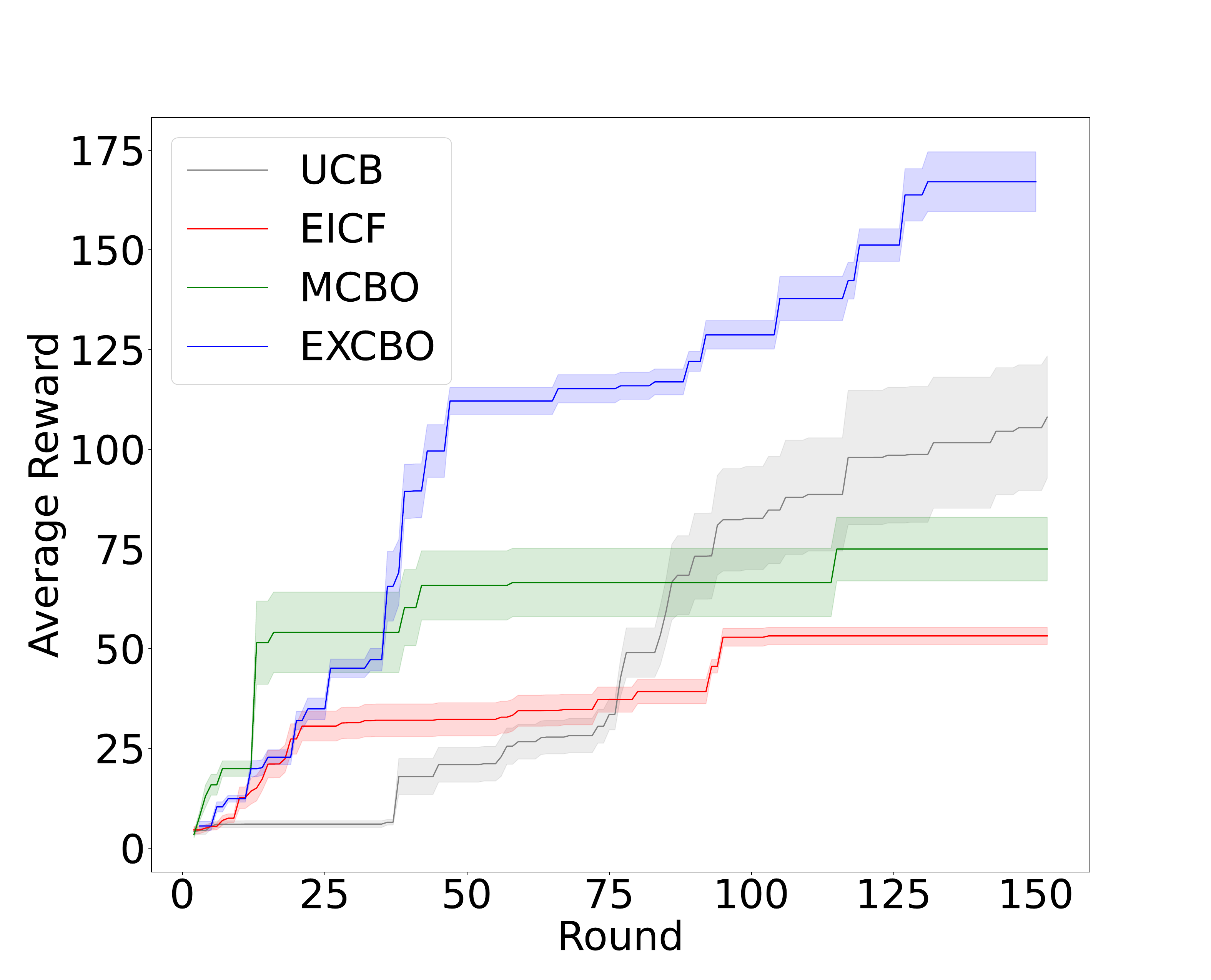}
\caption{(a-c): Results of Alpine2 (generated via Non-DGM mechanism in equation~\eqref{eq:Alpine2_complex}). } \label{fig:Alpine2}
\end{figure}

Due to computational constraints, we use $\sigma \in \{0.05, 0.1, 0.2\}$. The corresponding results are reported in Figures~\ref{fig:Alpine2}-(a–c). As shown, EXCBO consistently achieves the best performance across all noise levels, demonstrating the effectiveness and advantages of the proposed method. 
Although the Alpine2 generation mechanism does not strictly follow DGM or BGM, the strong results of EXCBO, as illustrated in Figures~\ref{fig:Alpine2}-(a–c), highlight its generalization capability, providing further empirical support for the theoretical claims in Sections~\ref{sec:onenode} and~\ref{sec:edl_app}.

\subsection{Epidemic Model Calibration} 

We adopt the additive noise model (ANM~\cite{hoyer2008nonlinear}), i.e., $X_i = f(\mathbf{Z}_i) + U_i$, where $U_i \sim p(U) = 0.5\mathcal{N}(\mu_1, c_1\sigma^2) + 0.5\mathcal{N}(\mu_2, c_2\sigma^2), \ c_1, c_2 > 0$. Since ANM is a subset of DGM, this setup also satisfies the DGM assumption. To ensure consistency, we normalize and standardize all action nodes to the range $[0,1]$. Specifically, $\beta$ is rescaled to $[0,1]$, with $\gamma = 0.5$, $I_{i,0} = 0.01$ for $i \in \{0,1\}$, and $T = 3$. For $U_{i,j}$ with $i \in \{1,2\}$ and $j \in \{1,2,3\}$, we set $w_1 = w_2 = 0.5$, $\mu_1, \mu_2 \in [-1.0, 1.0]$, and $c_1, c_2 \in \{0.5, 1.0, 1.5\}$. With the capability to recover and learn the exogenous distributions, our method is more robust and stable in this application scenario. Similarly constrained by computational overhead, we use $\sigma \in \{0.1, 0.3\}$, with the other $p(U)$ hyperparameters set as in the Alpine2 experiments. Figure~\ref{fig:epidemic} shows that increased exogenous noise enhances the performance of all methods. Our EXCBO  performs better than state-of-the-art model calibration methods in both cases, and our method has a faster convergence rate compared to the baselines.

\subsection{Planktonic Predator–prey Community in a Chemostat}

\begin{figure}[h]
\centering
\includegraphics[width=0.25\textwidth]{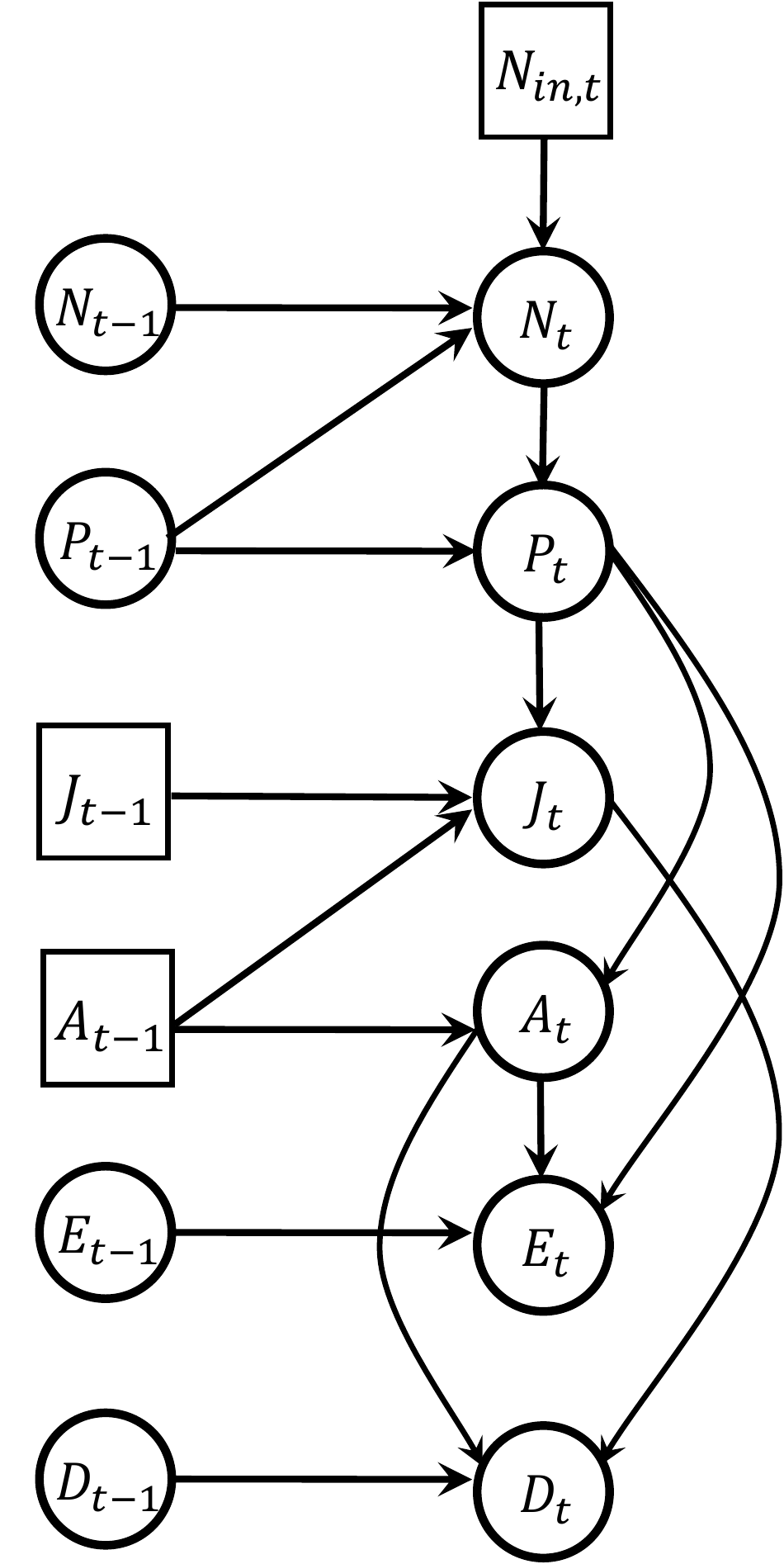}
\caption{P3C$^2$ graph structure; exogenous nodes are not included. } \label{fig:p3c2_graph}
\end{figure}

 We use the system of ordinary differential equations (ODE) given by~\cite{blasius2020long,aglietti2021dynamic} as our SCM and construct the DAG by rolling out the temporal variable dependencies in the ODE of two adjacent time steps while removing graph cycles. Observational data are provided in ~\cite{blasius2020long}, and are use to compute the dynamic causal prior. So different from  dynamic sequential CBO~\citep{aglietti2021dynamic},  we use the causal structure at $t$ and $t+1$ as the DAG for the  algorithms. The causal graph is given in Figure~\ref{fig:p3c2_graph}.

At each time step, the system includes the following variables:

- $N_{in}$: Nitrogen concentration in the external medium  

- $N$: Nitrogen (prey) concentration  

- $P$: Phytoplankton (predator) concentration  

- $E$: Predator egg concentration  

- $J$: Predator juvenile concentration  

- $\mathbf{A}$: Predator adult concentration  

- $D$: Dead animal concentration

Equations (21–26) in~\cite{aglietti2021dynamic} define the ODE, and equations~(\ref{eq:p3c2_scm}-\ref{eq:Dt}) specify the corresponding SCM. The action variables are $N_{in,t}$, $J_t$, and $A_t$, which we manipulate to minimize $D_{t+1}$.  We use GPs to fit the following SCM 
\begin{align}\label{eq:p3c2_scm}
N_t &= f_N(N_{in,t}, N_{t-1}, P_{t-1}, \epsilon_N) \\ 
P_t &= f_P(N_t, P_{t-1}, \epsilon_P) \\ 
J_t &= f_J(P_t, J_{t-1}, A_{t-1}, \epsilon_J) \\ 
A_t &= f_A(P_t, A_{t-1}, \epsilon_A) \\ 
E_t &= f_E(P_t, A_{t}, E_{t-1}, \epsilon_E) \\~\label{eq:Dt}
D_t &= f_D(J_t, A_{t}, D_{t-1}, \epsilon_D) .\\ \notag
\end{align}
The data~\footnote{\url{https://figshare.com/articles/dataset/Time_series_of_long-term_experimental_predator-prey_cycles/10045976/1}} processing is following~\cite{aglietti2021dynamic}, and $\big\{\epsilon_j | j \in \{ N, P, J, A, E, D\} \big\}$ are standard normal distributions. As shown in Figure, the three action nodes are $N_{in,t}, J_{t-1}$, and $A_{t-1}$. The intervention domains are $N_{in,t} \in [60.0, 100.0]$, $J_{t-1} \in [0.0, 36.0]$, and $A_{t-1} \in [0.0, 180.0]$. Here, the domains are from the value range of the data. According to the result in Figure~\ref{fig:p3c2}, EXCBO outperforms all the baselines on this real-world dataset. In addition, MCBO achieves better results compared to the other two baselines.  It indicates that the proposed EDS and DGM framework can approximately recover the exogenous distribution from the real-world data and improve the intervention results.

\subsection{Pooled Testing for COVID-19}\label{sec:covid19}

\begin{figure}[H]
\centering 
(a) $\sigma = 0.2$ ~~~~~~~~~~~~~~~~~~~~~~~~~~~~~~~~~~~~ (b) $\sigma = 0.4$~~~~~ ~~~~~~~~~~~~~~~~~~~~~~~~~~~~~~~ (c) ~~~~~~~~~~  \\ 
 \includegraphics[width=0.32\textwidth]{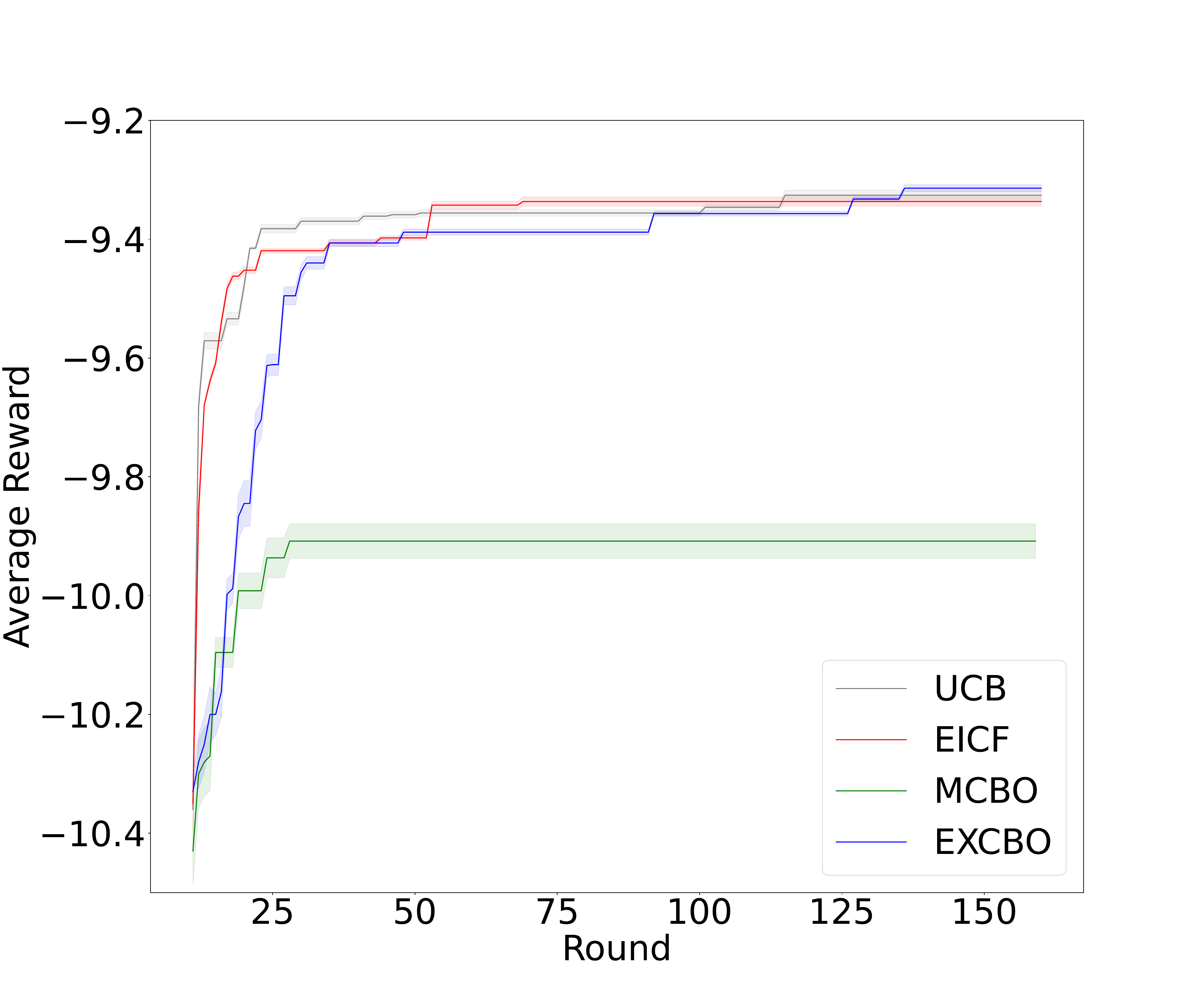} 
\includegraphics[width=0.32\textwidth]{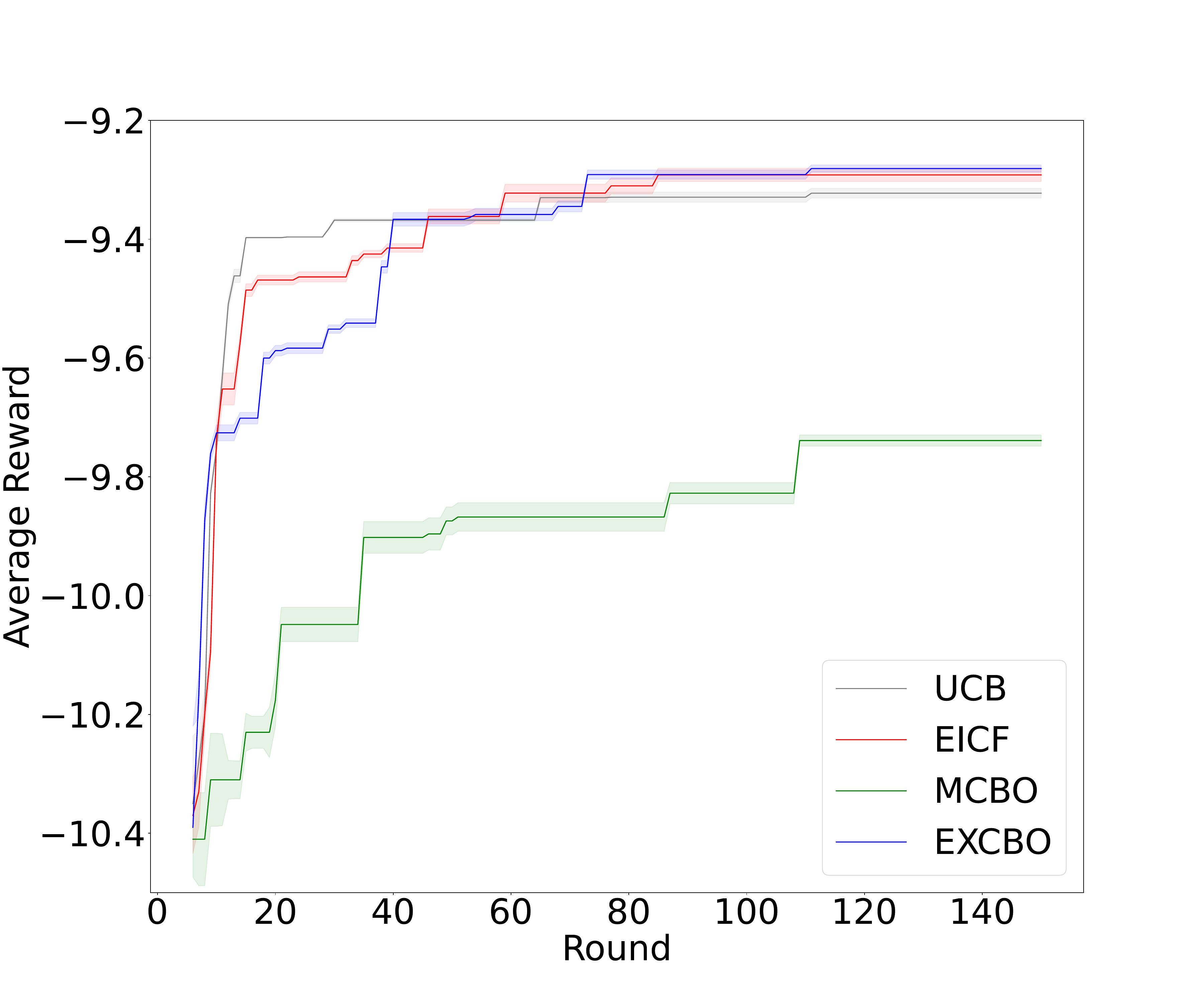}
\includegraphics[width=0.32\textwidth]{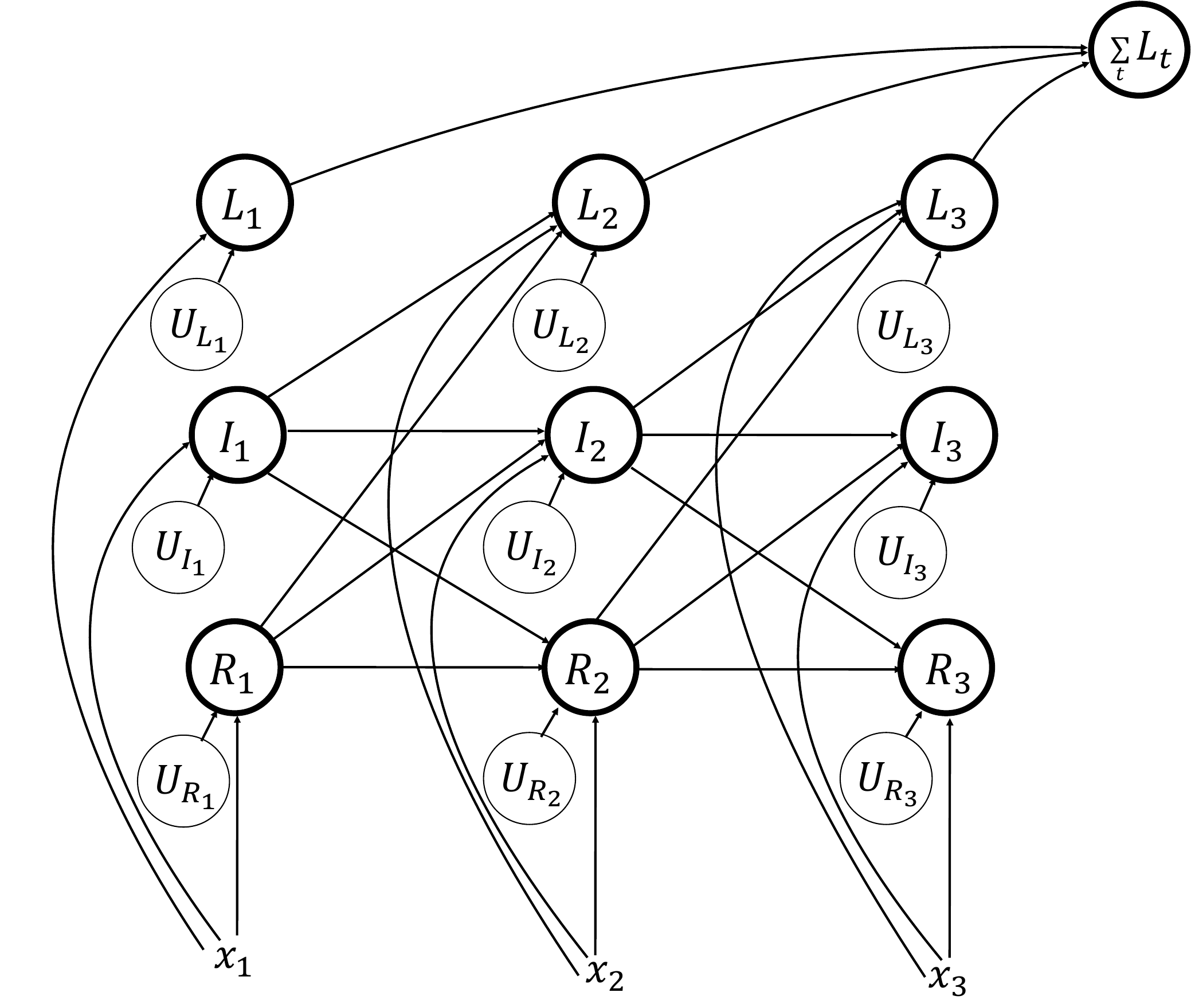}
\caption{ (a-b): Results of COVID-19 pooled testing optimization; (c): Graph structure for COVID-19 pooled testing problem. } \label{fig:covid}
\end{figure}

We further compare EXCBO and existing methods using the COVID-19 pooled testing problem~\citep{astudillo2021bayesian}. The graphical structure is given by Figure~\ref{fig:covid}-(c). In Figure~\ref{fig:covid}-(c), $I_t$ is the fraction of the population that is infectious at time $t$; $R_t$ is the fraction of the population that is recovered and cannot be infected again, and time point $t \in \{1, 2, 3\}$. The additional fraction $S_t = 1- I_t - R_t$  of the population is susceptible and can be infected. During each period $t$, the entire population is tested using a pool size of $x_t$. The loss $L_t$, incorporates the costs resulting from infections, testing resources used, and individuals isolated at period $t$. The objective is to choose pool size $x_t$ to minimize the total loss $\sum_t L_t$. Therefore, $x_t$s are the action variables/nodes that the algorithms try to optimize to achieve lower costs.

We employ the  ANM~\citep{hoyer2008nonlinear} setup: $X_i = f(\mathbf{Z}_i) + U_i$, where $U_i \sim p(U) = 0.5\mathcal{N}(\mu_1, c_1\sigma^2) + 0.5\mathcal{N}(\mu_2, c_2\sigma^2), \ c_1, c_2 > 0$. Data are generated using the dynamic SIR model from~\cite{astudillo2021bayesian} with $\beta = 3.23$. For varying exogenous distributions $p(U)$, we use $\mu_1, \mu_2 \in [-0.5, 0.5]$ and $c_1, c_2 \in \{0.05, 0.5, 1.0\}$.

Figure~\ref{fig:covid}-(a-b) presents the optimization results obtained from different methods, where the reward is defined as $y = -\sum_t L_t$. As shown in Figure~\ref{fig:covid}, UCB, EICF, and EXCBO exhibit similar performance across both $\sigma$ values. However, after 140 rounds, EXCBO achieves the best overall performance. The relatively poor performance of MCBO can be attributed partly to the bias introduced by the use of single-mode Gaussian distribution, and partly to the overfitting issues of the neural networks.

\subsection{Results on Multimodal Exogenous Distribution}

\begin{figure}[h]
\centering
{\small  (a) Dropwave, $\lambda = 1.0, \sigma = 0.1$; ~~~~~~ (b) Alpine2, $\lambda = 1.0, \sigma = 0.4$; ~  (c)  Alpine2, $\lambda = 1.0, \sigma = 0.1$, NonDGM;     } \\ 
 \vspace{-0.03in}
\includegraphics[width=0.32\textwidth]{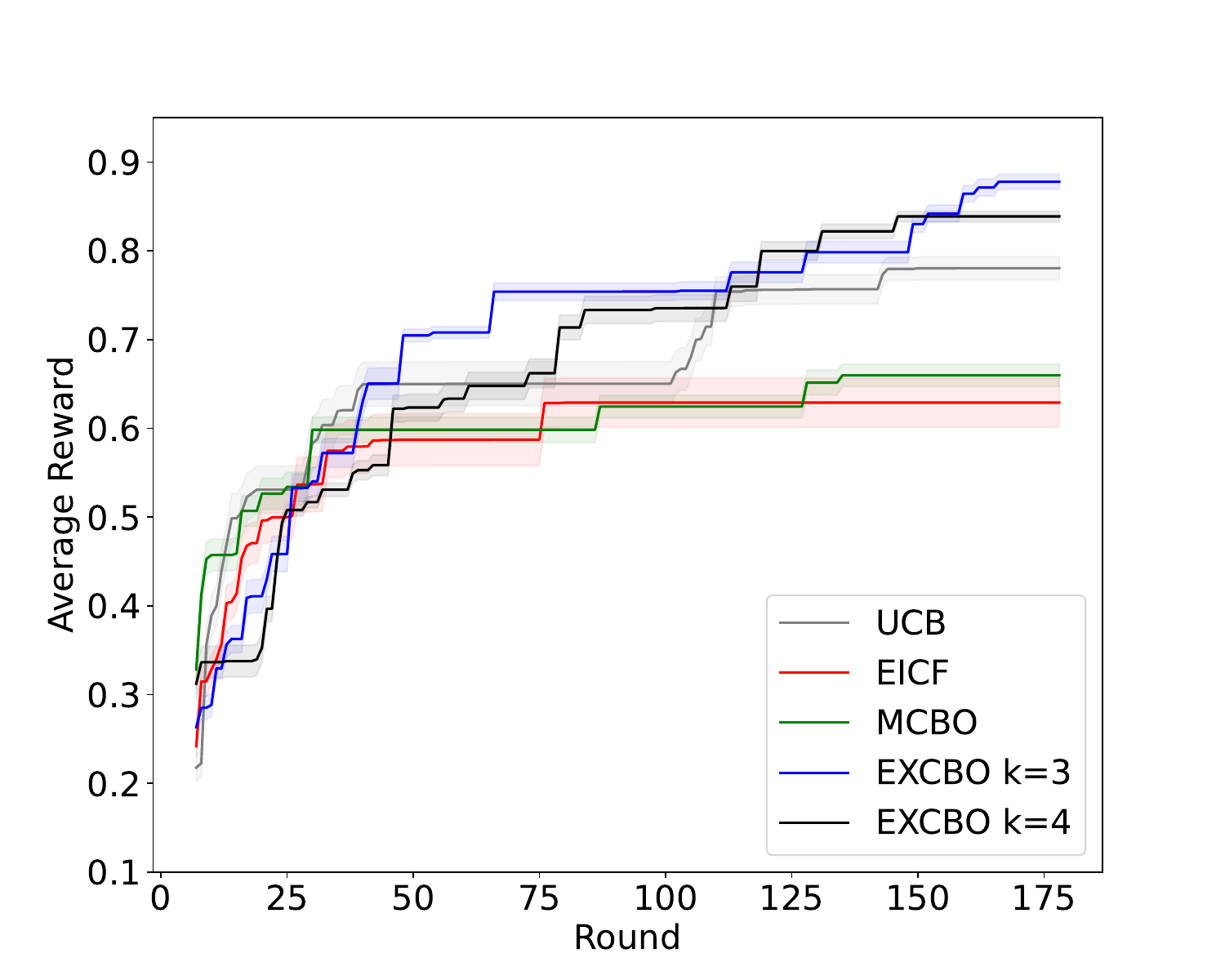} 
\includegraphics[width=0.32\textwidth]{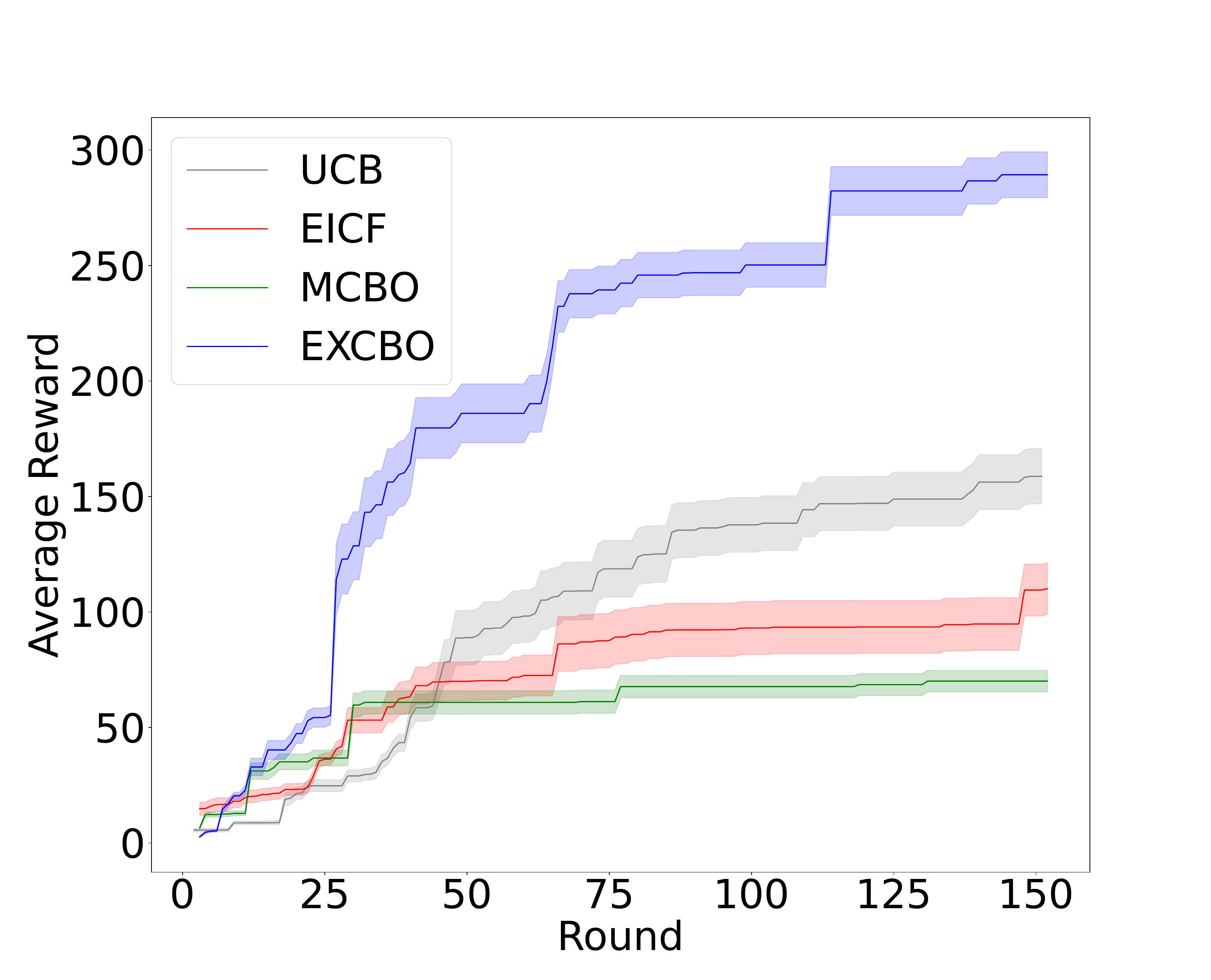}
\includegraphics[width=0.32\textwidth]{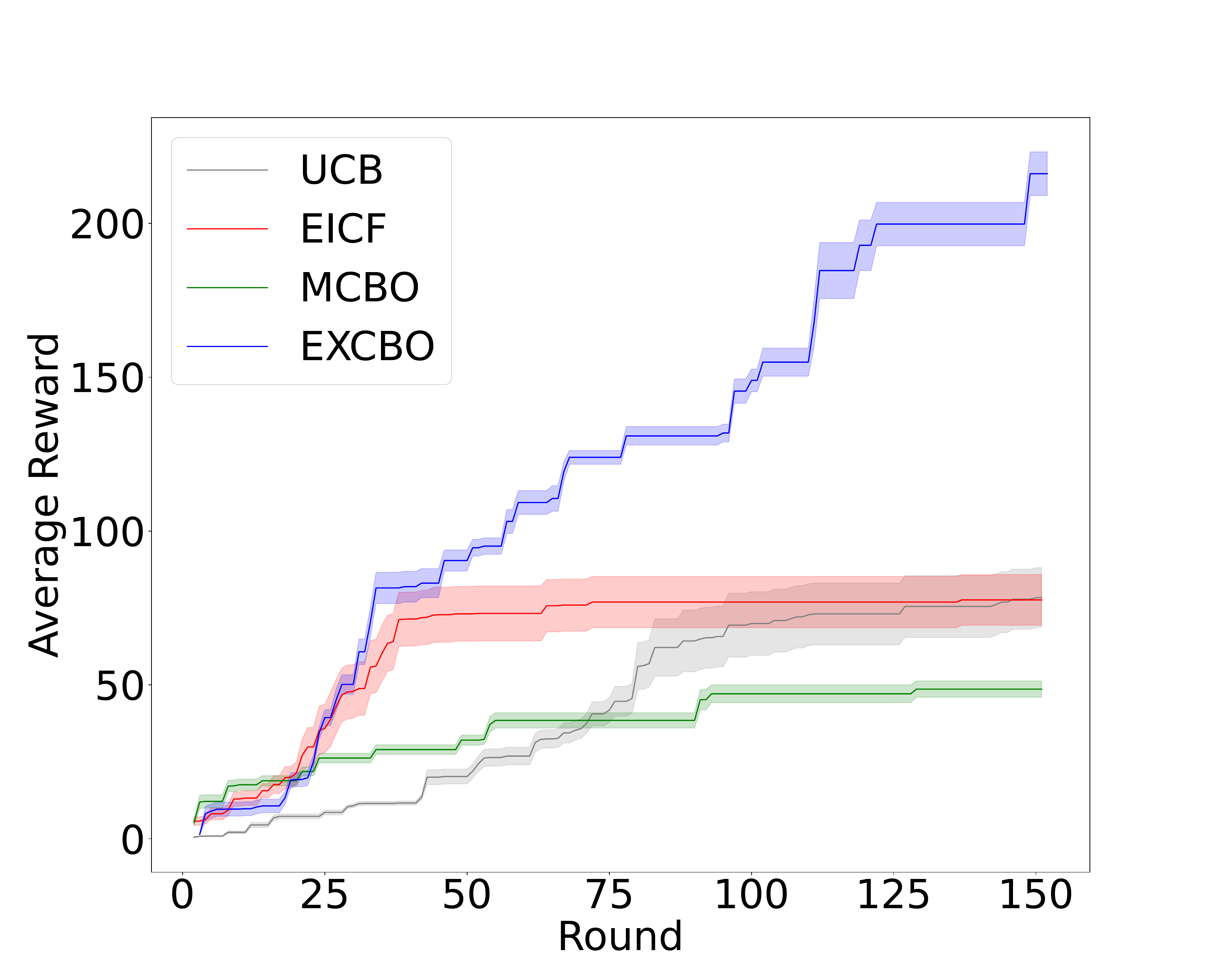} 
\caption{Results on datasets with three-mode exogenous distributions.} \label{fig:multicomp}
\end{figure}

We compare the algorithms by using datasets with three-mode exogenous distributions. 
The plots in Figure~\ref{fig:multicomp} show the results of different algorithms on datasets generated with three-mode exogenous distributions, i.e. \begin{align}\label{eq:3com}
&p(U) = w_1 \mathcal{N}(\mu_1, c_1\sigma^2) + w_2 \mathcal{N}(\mu_2, c_2\sigma^2) + w_3 \mathcal{N}(\mu_3, c_3\sigma^2) , \\ \notag
& w_1, w_2, w_3, c_1, c_2, c_3 > 0, w_1 + w_2 + w_3 =1.0.
\end{align}
The data generation setups follow Sections~\ref{sec:app:Dropwave}, \ref{sec:app:alphine2}, and \ref{sec:app:Alpine2_NonDGM}. EXCBO uses three-component GMM prior ($k=3$) for plots ~\ref{fig:multicomp}-(b) and  \ref{fig:multicomp}-(c). EXCBO runs with three-component and four-component GMM prior ($k=3, 4$) on Dropwave dataset. These results prove that EXCBO can robustly outperform the baselines when the data is generated with multimodal noise.

\begin{figure}[h]
\centering
{\small (a) Dropwave, $\lambda = 1.0, \sigma = 0.1$; ~~~~~ (b) Alpine2, $\lambda = 2.0, \sigma = 0.1$; ~~~~~~~~~~~~  (c) ~~~ P3C$^2$  ~~~~~~~~~~~     } \\ 
 \vspace{-0.03in}
\includegraphics[width=0.32\textwidth]{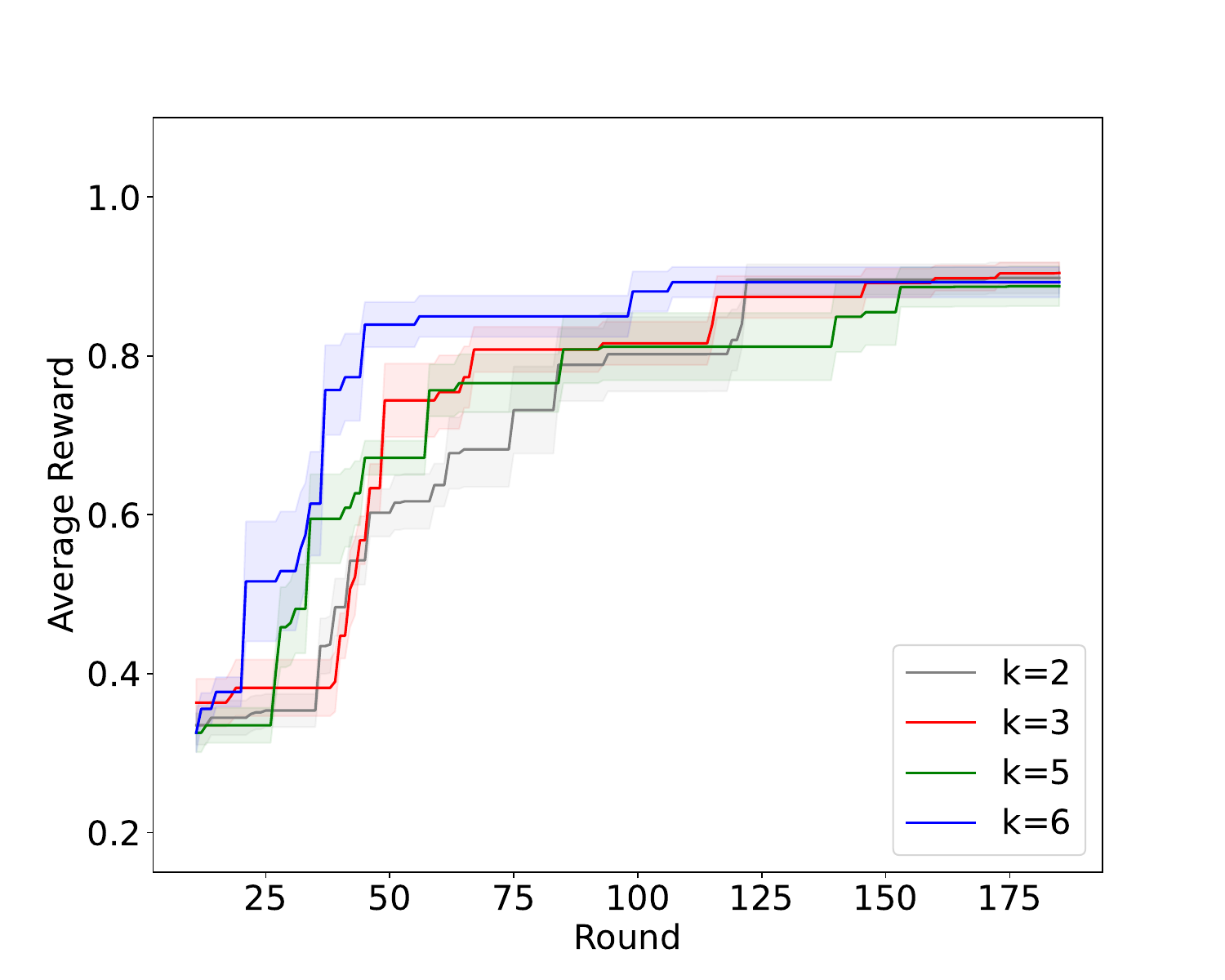} 
\includegraphics[width=0.32\textwidth]{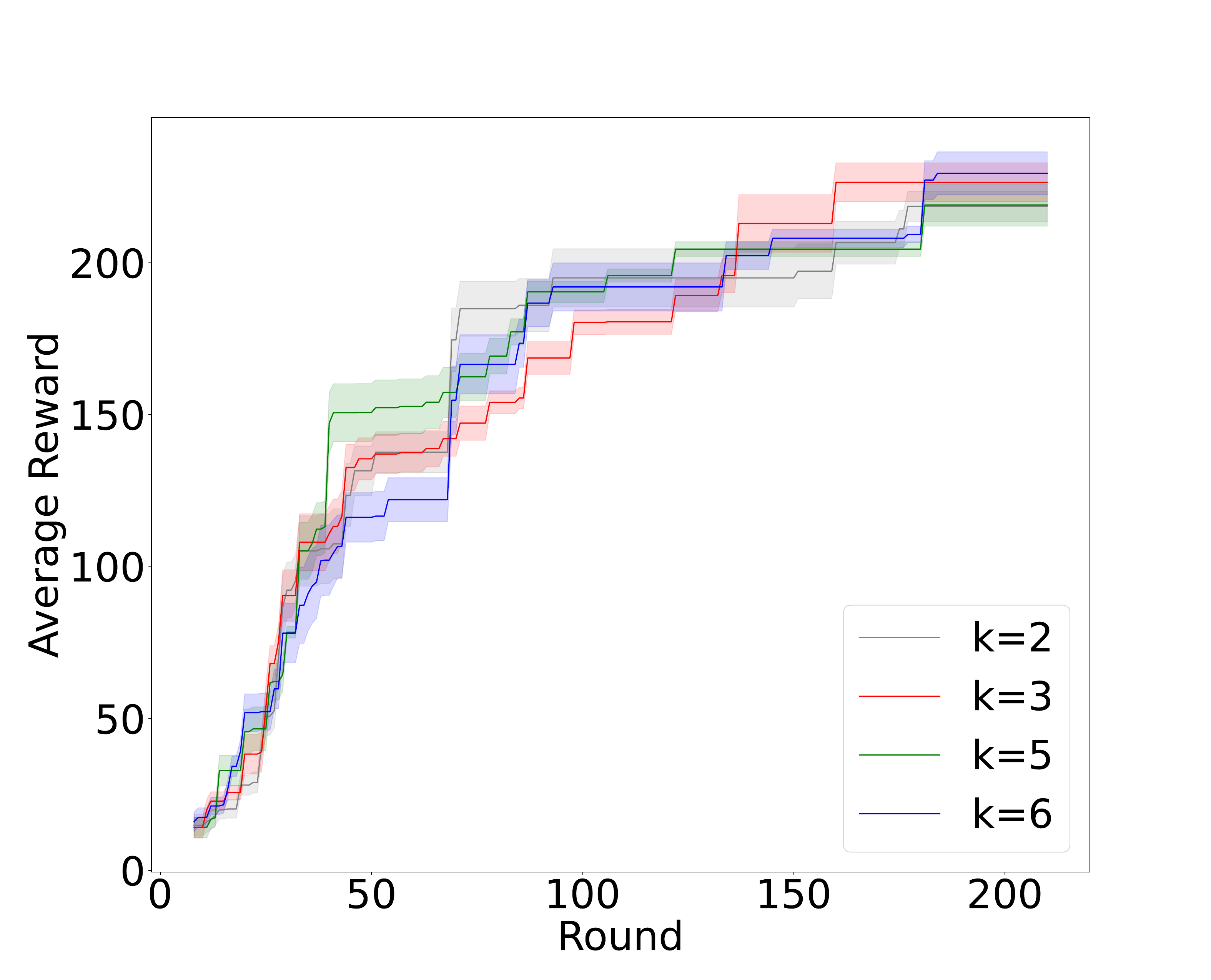}
\includegraphics[width=0.32\textwidth]{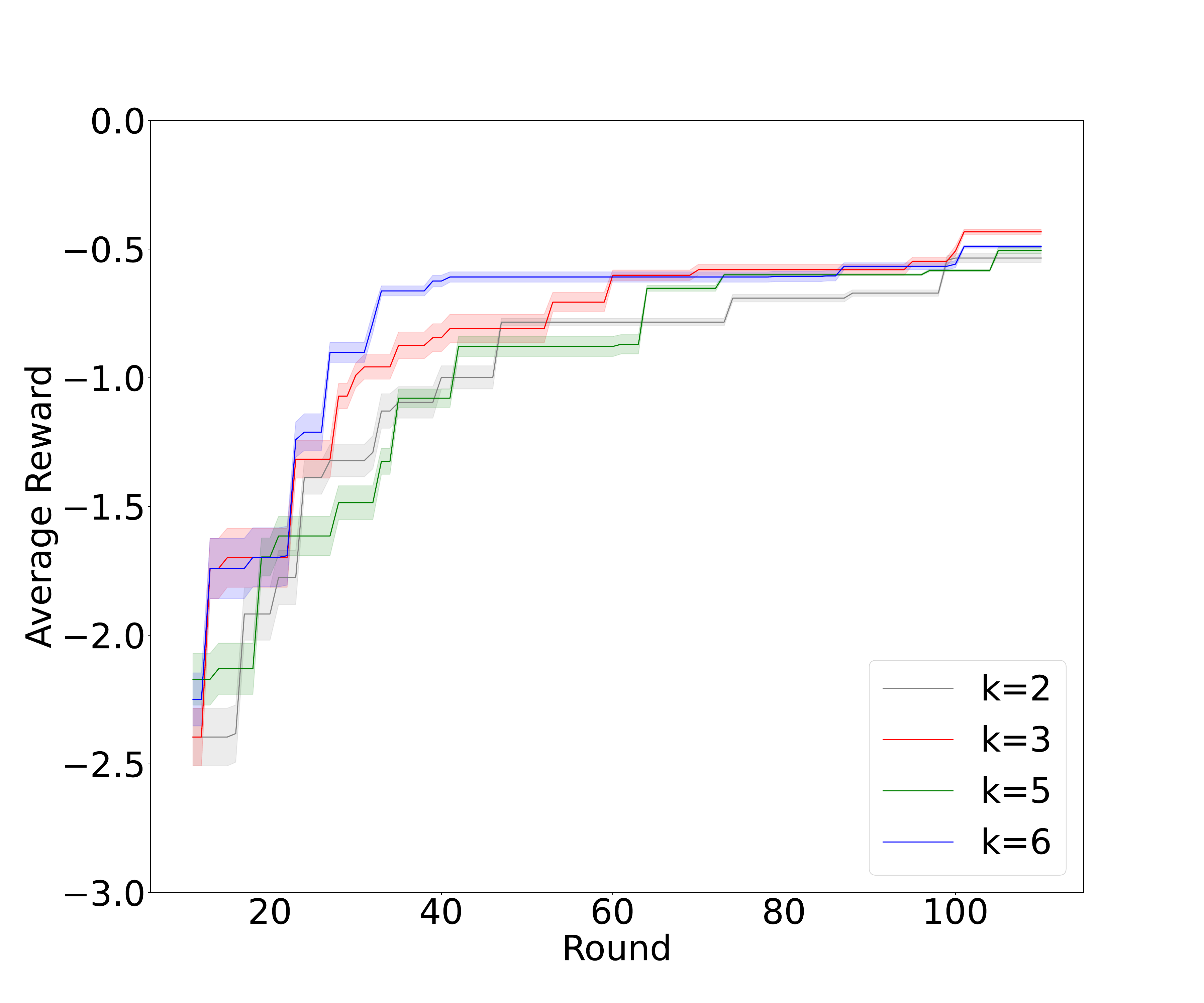} 
\caption{Results of EXCBO using  different  GMM component numbers} \label{fig:GMM_k}
\end{figure}

Figure~\ref{fig:GMM_k} gives the results of EXCBO  on two synthetic datasets and one real-world dataset using different GMM component numbers. The synthetic datasets are generated using two-component GMM noise.   We can see that  EXCBO models with different GMM  component numbers give similar results.

\subsection{Running Time}

\begin{figure}[h]
\centering 
\includegraphics[width=0.4\textwidth]{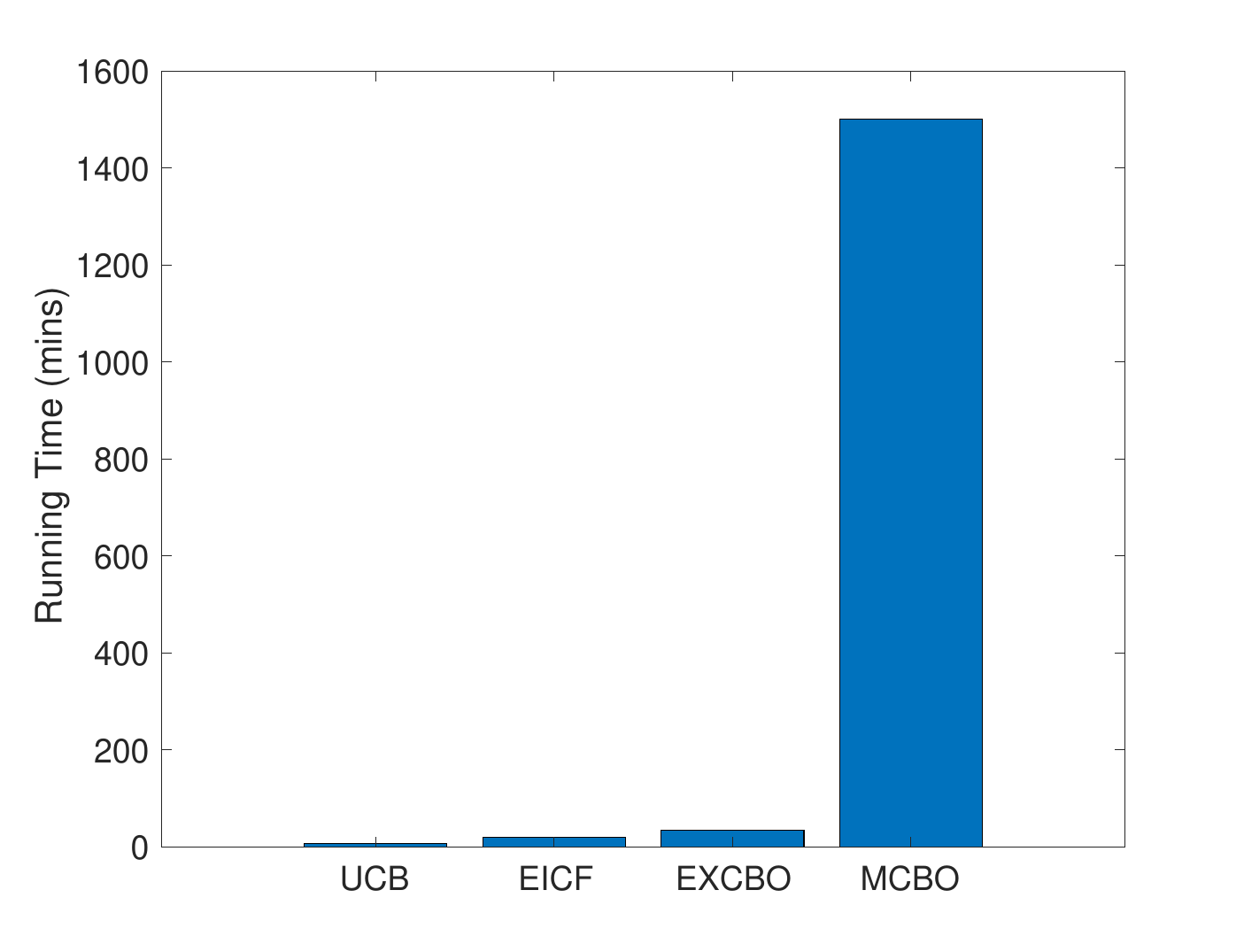}
\caption{Running time of the algorithms on Dropwave data with  $\sigma = 0.1$ and $\lambda = 1.0$ for four random seeds.} \label{fig:run_time}
\end{figure}

\begin{table}[H]
\begin{center}
\caption{Running time of the algorithms on Dropwave data with  $\sigma = 0.1$ and $\lambda = 1.0$ for four random seeds.}~\label{tb:run_time}
\begin{tabular}{ | c|c|c|c|c | c  } 
 \hline
Methods  & UCB & EICF & EXCBO & MCBO  \\ 
\hline
Running Time (mins) & $6.5$  &  $19.6$ & $35.1$  &$1501.3$   \\
 \hline
\end{tabular}
\end{center}
\end{table}

Figure~\ref{fig:run_time} and Table~\ref{tb:run_time} report the actual running time of  the four algorithms on the Dropwave dataset with $\sigma = 0.1$ and $\lambda = 1.0$. Relative running times across datasets are consistent with the ratios shown in the figure. Empirically, EXCBO requires a similar amount of CPU time per iteration as UCB and EICF. In contrast, MCBO consumes significantly more computational resources - around 40 times as much - due to its reliance on neural networks. This highlights EXCBO’s scalability advantage over existing state-of-the-art methods.

\subsection{EXCBO and MCBO on Single-Mode Exogenous Distribution}

We  follow exactly the same setting in MCBO paper~\citep{sussex2022model} to compare EXCBO and MCBO using Dropwave data, i.e.,
$a_0, a_1 \in [0, 1]$, $X = \sqrt{(10.24 a_0- 5.12)^2+ (10.24 a_1 -5.12)^2} $,  and $Y = (1.0 + \cos(12.0 X))/(2.0 + 0.5X^2)+  0.1U $, $U \sim \mathcal{N}(0,1)$, and  the data generation code is from the MCBO package.
The exogenous environment noise is unit-Gaussian scaled  by 0.1.
We report the best expected reward for both EXCBO and MCBO  in Table~\ref{tb:Dropwave_single}. We can see EXCBO achieves improved  performance in most steps, but  MCBO gives a  better result in the final round step $t=100$.

\begin{table}[H]
\begin{center}
\caption{Results of Dropwave with unit-Gaussian noise.}\label{tb:Dropwave_single}
\begin{tabular}{ | c|c|c|c|c | c | c } 
 \hline
Round  & 20 & 40 & 60 & 80 & 100 \\ 
\hline
MCBO  & $0.78 \pm 0.05$ & $0.83 \pm 0.04$  & $0.87 \pm 0.03$  & $0.88  \pm 0.03$   & $0.91 \pm 0.02$  \\ 
EXCBO & $0.76 \pm 0.04$ & $0.84 \pm 0.04$  & $0.89  \pm 0.03$ & $0.89  \pm 0.02$ & $0.89 \pm 0.02$  \\ 
 \hline
\end{tabular}
\end{center}
\end{table}

Similarly, we follow the exact setting of Alphine2 in MCBO paper, 
i.e., $X_0 =- \sqrt{10.0a_{0}} \sin{(10.0a_0 )}  + U_0 $,  
$X_i = \sqrt{10.0a_{i}}   \sin{(10.0a_{i} )}   X_{i-1} + U_i$ for $1 \leq i \leq 5$; and here  $\mathbf{a}_i \in [0, 1],  U_i \sim \mathcal{N}(0,1), 0\leq i \leq 5$.
The exogenous environment noise is unit-Gaussian as reported in the MCBO paper.
We report the best expected reward for both EXCBO and MCBO  in Table~\ref{tb:Alphine2_single}.

\begin{table}[H]
\begin{center}
\caption{Results of Alphine2 with unit-Gaussian noise.}\label{tb:Alphine2_single}
\begin{tabular}{ | c|c|c|c|c | c | c } 
 \hline
Round  & 20 & 40 & 60 & 80 & 100 \\ 
\hline
MCBO  & $38.46 \pm 14.13$  &  $76.47 \pm 16.56$ & $189.40\pm 15.43$  &$327.07  \pm 12.38$   & $363.86 \pm 3.26$ \\ 
EXCBO & $28.98 \pm 13.32$ & $106.42 \pm 33.44$   & $166.48  \pm 42.43$ & $196.22  \pm 32.33$ &  $241.57 \pm 14.00$   \\ 
 \hline
\end{tabular}
\end{center}
\end{table}

From these results, we conclude that for  single-mode Gaussian exogenous distributions, MCBO performs better than EXCBO when the exogenous noise is strong (i.e., large $\sigma$, or large scale coefficient).
In contrast, EXCBO achieves comparable or superior performance when the exogenous signal is weak  or when $\sigma$ is small.

For multimodal exogenous distributions, as reported in Sections~\ref{sec:epidemic} and \ref{sec:covid19}, MCBO tends to be more vulnerable to complex exogenous distributions, particularly when they involve multimodal exogenous distributions with small variances. By comparison, the proposed exogenous learning framework effectively mitigates these challenges.

\subsection{Analysis on Experimental Results}

The experimental results across different datasets demonstrate that learning the exogenous distributions enhances EXCBO's ability to achieve optimal reward values. In particular, incorporating the distribution of exogenous variables yields a more accurate surrogate model when given an SCM and observational data.  

Our method shows clear advantages over existing approaches when the exogenous noise is relatively weak. In such cases, the Gaussian Processes employed by UCB, EICF, and MCBO fail to capture the multimodality of the exogenous distribution, leading to a biased surrogate model with respect to the optimal intervention values. In contrast, EXCBO leverages a Gaussian mixture model, which effectively captures the multimodal exogenous distribution recovered by the proposed EDS under the DGM conditions. When the multimodal distribution of $U_i$ in $X_i = f(\mathbf{Z}_i, U_i)$ has small variances, the uncertainty is highly concentrated, making it harder to distinguish different modes in the plausible function map and resulting in larger bias in the objective approximation. By contrast, larger variances in the exogenous distribution allow the GPs in UCB, EICF, and MCBO to better discriminate between modes, thereby providing more accurate estimates of the expected objective function, i.e., $X_i = \mathbb{E}_{p(U_i)} f(\mathbf{Z}_i, U_i)$.  

Gaps among different methods have been reported in previous studies, e.g., in MCBO~\citep{sussex2022model}, Figures~2-f, 2-c, and 2-d. We speculate that this discrepancy arises because GPs with plain kernels are not universal approximators. Consequently, their limited expressiveness leads to irreducible bias, even with infinite data samples. This underscores the importance of incorporating structural knowledge to improve performance, as evidenced in MCBO, EICF, and EXCBO.  

Finally, the regret bound in Theorem~\ref{thm:regret} depends on Assumptions~\ref{assum:Lipschitz}--\ref{assum:calibrated} and holds with probability $1-\alpha$, where $\alpha$ is specified in Assumption~\ref{assum:calibrated}. This implies that different GP-based CBO methods are not guaranteed to converge to the same optimal reward value.

\section{Proof of Theorem~\ref{thm:exo1}}~\label{sec:proof_thm}

Before we prove Theorem~\ref{thm:exo1}, we present a similar result for ANMs~\citep{hoyer2008nonlinear}. 

\begin{theorem}~\label{thm:exo_anm}
Let  $(X, \mathbf{Z}, U, f)$ be a node SCM.  Let $\rho(): \mathcal{X}\times \mathcal{Z}   \to  \mathbb{R}^1 $ be a predefined function regarding $X$ and $\mathbf{Z}$, and $\phi()$ be a regression model with  $\phi(): \mathcal{Z} \to  \rho(\mathcal{X}, \mathcal{Z})$. We define an encoder function $h(): \mathcal{Z} \times  \mathcal{X} \to  \widehat{\mathcal{U}}$ with $\widehat{U}:= h(\mathbf{Z}, X):= \rho(X, \mathbf{Z}) -  \phi(\mathbf{Z})$.  The decoder is $g(): \mathcal{Z} \times  \widehat{\mathcal{U}} \to  \mathcal{X}$, i.e., $X= g(\mathbf{Z}, \widehat{U})$.
Let $\rho()$ maps the values of  $X$ and  $\mathbf{Z}$ to an additive function of $\mathbf{Z}$ and $U$, i.e., $\rho(X, \mathbf{Z}) = \rho_1(\mathbf{Z}) + \rho_2(U)$. 
Then $ \widehat{U}= h(\mathbf{Z}, X) = \rho_2(U) -  \mathbb{E}[\rho_2(U)]$, and $\widehat{U} \indep \mathbf{Z}$.
\end{theorem}

\begin{proof}
As $\phi(\mathbf{z})$ is an optimal approximation of $\rho(X, \mathbf{z})$, with $\mathbf{Z} \indep U$, we have 
\begin{align*}
 \phi(\mathbf{z}) &=  \mathbb{E}[\rho(X, \mathbf{z})] = \mathbb{E}[\rho_1(\mathbf{z}) + \rho_2(U)] =\int \big(  \rho_1(\mathbf{z}) + \rho_2( u)\big) p(u) d u \\
&=  \rho_1(\mathbf{z}) +  \mathbb{E}[\rho_2(U)].
\end{align*}
Thus, the decoder becomes
\begin{align*}
 h(\mathbf{z}, x) &= \rho(x, \mathbf{z})-  \phi(\mathbf{Z}=\mathbf{z}) \\
 &=\rho_1(\mathbf{z}) + \rho_2(u) - \rho_1(\mathbf{z}) -  \mathbb{E}[\rho_2(U)]\\
 &=  \rho_2(u) -  \mathbb{E}[\rho_2(U)]. 
\end{align*}

Therefore, $\widehat{U} = h(\mathbf{Z}, X)= \rho_2(U) -  \mathbb{E}[\rho_2(U)]$ is a function of $U$, and $h(\mathbf{Z}, X) \indep  \mathbf{Z} $, i.e., $\widehat{U} \indep  \mathbf{Z} $.
\end{proof}

\begin{example}\label{exp:anm}
For an ANM~\citep{hoyer2008nonlinear} model $X = f(\mathbf{Z}) + U$, we have  $\rho(X, \mathbf{Z}) = X$, $\rho_1(\mathbf{Z}) = f(\mathbf{Z})$, and $\rho_2(U) =U$, then  $ \widehat{U}= h(\mathbf{Z}, X) = U -  \bar{U}$ . 
\end{example}

\begin{example}
For a model $X =2Z e^{-U} - e^{-Z}$, we have  $\rho(X, Z) = \log (X +  e^{-Z})$, $\rho_1(Z) =\log(2Z)$, and $\rho_2(U) =-U$, then  $ \widehat{U}= h(Z, X) = -U +  \bar{U}$ . 
\end{example}

Example~\ref{exp:anm} shows that the exogenous variable in any ANM model is identifiable. 
In practice, variable $X$'s generation mechanism $f()$ is generally unknown, and it is hard to propose a  general form function $ \rho()$ that can perform on any $f()$s and transform them to ANMs. 

\noindent\textbf{Theorem~\ref{thm:exo1}}{\it \ \ 
Let $(\mathbf{Z}, U, X, f)$ be a node SCM, and $(\widehat{U}, \phi, h, g)$ an EDS surrogate of $U$. Suppose $f$ has the DGM structure, i.e. $X = f(\mathbf{Z}, U) = f_a(\mathbf{Z}) + f_b(\mathbf{Z}) f_c(U)$ with $f_b(\mathbf{z}) \neq 0$ for all $\mathbf{z} \in \mathcal{Z}$. In addition, each $\mathbf{z} \in \mathcal{Z}$ has $N$ base samples in the close neighborhood of $\mathbf{z}$,  i.e., $\big\{\mathbf{z}, x_i(\mathbf{z}, u_i)\big\}_{i=1}^N$ with $\{u_i\}_{i=1}^N \ i.i.d \sim p(U)$. 
Then with $N \rightarrow \infty$,  the surrogate $\widehat{U} \rightarrow \frac{s}{\sigma_{f_c}} \big(f_c(U)  -   \mathbb{E}[f_c(U)] \big)$, $\mathbb{E}[\widehat{U}] \rightarrow 0$, $\textrm{Var}[\widehat{U}] \rightarrow 1$, and $\widehat{U} \indep \mathbf{Z}$, where $\sigma_{f_c} = \sqrt{\mathbb{E}\big[\big(f_c(U)  -   \mathbb{E}[f_c(U)] \big)^2\big]}, s\in\{-1, 1\}$.
}

\begin{proof}
$\forall \mathbf{z} \in \mathcal{Z}$, with $\{u_i\}_{i=1}^N \ i.i.d \sim p(U)$,  $N \rightarrow \infty$, and $\mathbf{Z} \indep U$, 
the mean function is
\begin{align*}
  \mu_{\phi}(\mathbf{z}) &= \lim_{N \rightarrow \infty}  \frac{1}{N} \sum^N_i x_i(\mathbf{z}, u_i) = \int \big(  f_a(\mathbf{z}) +   f_b(\mathbf{z})f_c(u) \big) p(u) d u \\
&=   f_a(\mathbf{z}) +  \int  f_b(\mathbf{z})f_c(u) p(u) d u \\
&=  f_a(\mathbf{z}) +  f_b(\mathbf{z})\mathbb{E}\big[f_c(U)\big].
\end{align*}

With mean $\mu_{\phi}(\mathbf{z})$, and an observation  $x_i(\mathbf{z}, u_i)$, and $u_i \sim p(U)$, 
\begin{align}\notag
 &x_i(\mathbf{z}, u_i) -  \mu_{\phi}(\mathbf{z}) \\\notag
 =&f_a(\mathbf{z}) +   f_b(\mathbf{z})f_c(u_i)  -  f_a(\mathbf{z}) - f_b(\mathbf{z})\mathbb{E}\big[f_c(U)\big]\\ \label{eq:x_mn}
 =& f_b(\mathbf{z}) \big(f_c(u_i)  -   \mathbb{E}[f_c(U)] \big). 
\end{align}

With~\eqref{eq:x_mn}, $\forall \mathbf{z} \in \mathcal{Z}$, the variance of the regression model $\phi()$ is 

\begin{align} \notag
\sigma^2_{\phi}(\mathbf{z}) &= \lim_{N \rightarrow \infty} \frac{1}{N-1} \sum_{i=1}^N \big(x_i(\mathbf{z}, u_i) - \mu_{\phi}(\mathbf{z}) \big)^2 \\ \notag
& =  \lim_{N \rightarrow \infty} \frac{1}{N-1} \sum_{i=1}^N \big( f_b(\mathbf{z}) \big(f_c(u_i)  -   \mathbb{E}[f_c(U)] \big) \big)^2\\ \notag
& = f^2_b(\mathbf{z}) \lim_{N \rightarrow \infty} \frac{1}{N-1} \sum_{i=1}^N  \big(f_c(u_i)  -   \mathbb{E}[f_c(U)] \big)^2\\ \notag
& = f^2_b(\mathbf{z}) \int \big(f_c(u)  -   \mathbb{E}[f_c(U)] \big)^2 p(u) d u\\ \notag
&=f^2_b(\mathbf{z})\mathbb{E}\bigg[\big(f_c(U)  -   \mathbb{E}[f_c(U)] \big)^2\bigg] \\ \label{eq:fc_var}
 &= f^2_b(\mathbf{z}) \sigma_{f_c}^2.
\end{align}

$\sigma^2_{\phi}(\mathbf{z})$ is the variance function with respect to variable $\mathbf{Z}$, i.e., $\sigma_{\phi}(\mathbf{z}) = \sigma_{f_c}|f_b(\mathbf{z})|$. Then, by~\eqref{eq:x_mn} and ~\eqref{eq:fc_var}, with $N \rightarrow \infty$,

\begin{align}\notag
 \frac{x -  \mu_{\phi}(\mathbf{z})}{\sigma_{\phi}(\mathbf{z})} 
 =&\frac{ f_b(\mathbf{z}) \big(f_c(u)  -   \mathbb{E}[f_c(U)] \big)}{\sigma_{f_c} |f_b(\mathbf{z})|} \\ \label{eq:var_h}
 =&\frac{s}{\sigma_{f_c}}\big(f_c(u)  -   \mathbb{E}[f_c(U)] \big).
\end{align}
Here $s = \textrm{sign}[f_b(\mathbf{z})] \in \{1, -1\}$. As $f_b()$ is a continuous function, and $f_b(\mathbf{z})\neq 0, \forall \mathbf{z} \in \mathcal{Z} $, $s = \textrm{sign}[f_b(\mathbf{z})]$ is a constant value $\forall \mathbf{z} \in \mathcal{Z}$, either 1 or -1, and $s \indep \mathbf{Z}$.  

So with $N \rightarrow \infty$, 
\begin{align*}
\widehat{U} = \frac{X -  \mu_{\phi}(\mathbf{Z})}{\sigma_{\phi}(\mathbf{Z})} \rightarrow \frac{s}{\sigma_{f_c}} \big(f_c(U)  -   \mathbb{E}[f_c(U)] \big).
\end{align*}
It shows that with $N \rightarrow \infty$,  $\mathbb{E}[\widehat{U}]  \rightarrow 0$, $\textrm{Var}[\widehat{U}]  \rightarrow 1$, and $\widehat{U} \indep \mathbf{Z}$, and $s= -1$ or $1$, $\forall z \in \mathcal{Z}$.
\end{proof}










\section{Exogenous Distribution Learning}\label{sec:edl_app}

\subsection{Exogenous Distribution and Counterfactual Identifiability of DGM}

Under the monotonicity assumption on $f()$, the EDS framework can be extended to BGMs, building upon the analysis in~\citep{Luetal20,nasr2023counterfactual,nasr2023counterfactual_b,chao2023interventional}. 
\emph{Counterfactual} queries utilize functional models of generative processes to reason about alternative outcomes for individual data points, effectively answering questions like: “What if I had done A instead of B?” Such queries are formally described as a three-step process: abduction, action, and prediction~\citep{pearl2009causality}. A model that can be learned from data and execute these three steps is said to be \emph{counterfactually identifiable}.

It is straightforward to show that a node SCM with a decomposable $f()$ is counterfactually identifiable. Thus, Theorem~\ref{thm:exo1} introduces a novel class of node SCMs that achieve counterfactual identifiability beyond BGMs~\citep{nasr2023counterfactual}.

\begin{remark}
We use the distribution of $\widehat{U}  = h(\mathbf{Z}, X)$, i.e., $p(\widehat{U})$, to represent $p(U)$ within the surrogate model. Suppose a node SCM  mechanism  $f()$  follows the  DGM assumption  and the conditions in Theorem~\ref{thm:exo1}, then we have $\widehat{U} \indep \mathbf{Z}$ and the node SCM is counterfactually identifiable.
\end{remark}

Here, the parent set $\mathbf{Z}$ may include action variables, and the learned $\widehat{U}$ remains independent of the actions or interventions. Therefore, we can leverage the action variables to optimize the target variable through causal intervention operations.

This work lies within the line of research on counterfactual identification, such as ANM~\citep{hoyer2008nonlinear},
BGM~\citep{nasr2023counterfactual}, and LSNM~\citep{immer2023identifiability}. The proposed DGM is a new family of models that are counterfactually identifiable and can be easily
implemented using GPs. Gaussian mixture models are employed to learn the recovered exogenous variable distribution, enabling a more accurate surrogate of the true data-generating mechanism, as demonstrated in the paper and our responses. The applicability of the proposed framework extends beyond CBO to broader causal inference tasks, including interventions and counterfactual inference.

\subsection{Analysis on BGMs}

We first present a lemma on the BGM equivalence class of a node SCM with a monotonic mechanism. 
 
\begin{lemma}~\label{lm:bgm}
Let  $(\mathbf{Z}, U, X, f)$ be a node SCM. $\forall \mathbf{z} \in \mathcal{Z}$, $f(\mathbf{z}, \cdot)$ is differentiable and strictly monotonic regarding  $u \in \mathcal{U}$.
We define a differentiable and invertible encoder function $h(): \mathcal{Z} \times  \mathcal{X} \to  \widehat{\mathcal{U}}$, i.e., $\widehat{U}:= h(\mathbf{Z}, X)$, and $\widehat{U} \indep \mathbf{Z}$.  The decoder is $g(): \mathcal{Z} \times  \widehat{\mathcal{U}} \to  \mathcal{X}$, i.e., $X= g(\mathbf{Z}, \widehat{U})$. 
Then $\widehat{U}= h(\mathbf{Z}, X)$ is a function of $U$, i.e.,  $\widehat{U} = \mathbf{s}(U)$, and $\mathbf{s}()$ is an invertible function.
\end{lemma}

\begin{proof}
According to the definition of node SCM, we have $\mathbf{Z} \indep U$. According to the assumption, $\forall \mathbf{z} \in \mathcal{Z}$, $f(\mathbf{z}, u)$ is differentiable and strictly monotonic regarding  $u$. Hence $X = f(\mathbf{Z}, U)$ is a BGM, and we use $\mathbb{F}$ to represent BGM class that satisfies the independence ($\mathbf{Z} \indep U$) and the function monotone conditions. We can see that $h^{-1} \in \mathbb{F}$, $h^{-1}(\mathbf{z}, \cdot)= g(\mathbf{z}, \cdot)$, and $h^{-1}(\mathbf{z}, \cdot)$ and $f(\mathbf{z}, \cdot)$ are counterfactually 
 equivalent BGMs  that generate the same distribution $p(\mathbf{Z},X)$. Based  Lemma B.2, Proposition 6.2, and Definition 6.1 in~\citep{nasr2023counterfactual}, there exists an invertible function $\mathbf{s}()$ that satisfies  $\forall \mathbf{z} \in \mathcal{Z}, x \in  \mathcal{X}, h(\mathbf{z}, x) = \mathbf{s}(f^{-1}(\mathbf{z}, x))$, i.e., $\widehat{u} = h(\mathbf{z}, x) = \mathbf{s}(f^{-1}(\mathbf{z}, x))=\mathbf{s}(u)$, which is $\widehat{U} = \mathbf{s}(U)$. 
\end{proof}

We can easily prove that an EDS model of a monotonic  node SCM belongs to its BGM equivalence class under the independence assumption  $\widehat{U} \indep \mathbf{Z}$. 

\begin{theorem}~\label{thm:exo3}
Let  $(\mathbf{Z}, U, X, f)$ be a node SCM. $\forall \mathbf{z} \in \mathcal{Z}$, $f(\mathbf{z}, \cdot)$ is differentiable and strictly monotonic regarding  $u \in \mathcal{U}$. Let $(\widehat{U}, \phi, h, g)$  be an EDS surrogate of $U$. We further assume that $\widehat{U} \indep \mathbf{Z}$. 
Then $\widehat{U}= h(\mathbf{Z}, X)$ is a function of $U$, i.e.,  $\widehat{U} = \mathbf{s}(U)$, and $\mathbf{s}()$ is an invertible function.
\end{theorem}

\begin{proof}
It is to prove that the encoder of an EDS, i.e., $\widehat{U} = h(\mathbf{Z}, X)= \frac{X -  \mu_{\phi}(\mathbf{Z})}{\sigma_{\phi}(\mathbf{Z})}$, is invertible regarding $\widehat{U}$ and $X$ given a value of $\mathbf{Z}$. With the assumption $\widehat{U} \indep \mathbf{Z}$, by using the results of Lemma~\ref{lm:bgm}, we have  $\widehat{U}= h(\mathbf{Z}, X)$ is a function of $U$,  i.e.,   $\widehat{U} = \mathbf{s}(U)$, and $\mathbf{s}()$ is an invertible function.
\end{proof}

Based on the proof of Theorem~\ref{thm:exo3}, a node SCM with a monotonic mechanism is counterfactually identifiable by using an EDS model with the $\widehat{U} \indep \mathbf{Z}$ constraint.

\section{Regret Analysis }\label{sec:regret_app}

\subsection{Remarks on Regret Bound}

The analysis in this paper focuses on the DGM mechanisms. To extend the analysis to BGMs, we need to consider the computation cost involving the independence penalization on variables $\widehat{U}$ and $\mathbf{Z}$.  For mechanisms beyond DGMs and BGMs, we conjecture that the surrogate approximation accuracy may decrease, but the convergence rate may not decrease a lot. The cumulative regret provides insight into the convergence behavior of the algorithm.

Our analysis follows the study in~\citep{sussex2022model}. 
 In the DAG $\mathcal{G}$ over $\{X_i\}_0^d$, let $N$ be the maximum distance from a root to $X_d$, i.e., $N = \max_i \text{dist}(X_i, X_d)$. Here $ \text{dist}(\cdot, \cdot)$ is a measure of the edges in the longest path from $X_i$ to the reward node $X_d$. Let $M$ denote the maximum number of parents of any variables in $\mathcal{G}, M = \max_i |\mathbf{pa}(i)|$.
Let $L_t$  be a function of  $L_{g}$,  $L_{\sigma_g}$. According to Theorem~\ref{thm:exo1}, with the EDS structure given in Figure~\ref{fig:node} in the main text, the exogenous variable and its distribution can be recovered.  For each observation of the dynamic surrogate model, we assume the sampling of $p(\widehat{U})$, $\tilde{\widehat{u}}= \mathbf{s}(\tilde{u})= \mathbf{s}(u)$. This maximum information gain is commonly used in many Bayesian Optimizations~\citep{srinivas2009gaussian}. Many common kernels, such as linear and squared exponential kernels, lead to sublinear information gain in $T$, and it results in an overall sublinear regret for EXCBO~\citep{sussex2022model}.

\subsection{Proof of Theorem~\ref{thm:regret}}
We give the assumptions used in the regret analysis.  Assumption~\ref{assum:Lipschitz} gives the Lipschitz conditions of $g_i$, $\sigma_{g,i}$, and $\mu_{g,i}$. It holds if the  RKHS  of each $g_i$ has a Lipschitz continuous kernel~\citep{curi2020efficient,sussex2022model}. Assumption~\ref{assum:calibrated} holds when we assume that the $i$th GP prior uses the same kernel as the RKHS of $g_i$ and that $\beta_{i,t}$ is sufficiently large to ensure the confidence bounds in 
\begin{align*}
&\bigg| g_i(\mathbf{z}_{i}, \mathbf{a}_i, \widehat{u}_i) - \mu_{g,i,t-1}(\mathbf{z}_{i}, \mathbf{a}_i, \widehat{u}_i)  \bigg|   \leq  \beta_{i,t}\sigma_{g,i, t-1} (\mathbf{z}_{i}, \mathbf{a}_i, \widehat{u}_i) \  , \ \  \  \forall  \mathbf{z}_i \in  \mathcal{Z}_i, \mathbf{a}_i \in  \mathcal{\mathbf{A}}_i, \widehat{u}_i \in  \widehat{\mathcal{U}}_i.
\end{align*}


\begin{assumption}~\label{assum:Lipschitz}
$\forall g_i \in \mathbf{G}$, $g_i$ is $L_{g}$-Lipschitz continuous; moreover,  $\forall i, t$, $\mu_{g,i,t}$ and  $\sigma_{g,i,t}$ are $L_{\mu_g}$ and  $L_{\sigma_g}$ Lipschitz continuous. 
\end{assumption}

\begin{assumption}~\label{assum:f}
$\forall f_i \in \mathbf{F}$,  $f_i$ is differentiable and has a decomposable structure with $X = f_i(\mathbf{Z}_i, U_i)= f_{i(a)}(\mathbf{Z}_i) + f_{i(b)}(\mathbf{Z}_i) f_{i(c)}(U_i)$, and $f_{i(b)}(\mathbf{z}_i) \neq 0, \forall \mathbf{z}_i \in \mathcal{Z}_i$.
\end{assumption}

\begin{assumption}~\label{assum:calibrated}
 $\forall i, t$, there exists sequence $\beta_{i,t} \in \mathbb{R}_{>0}$, with probability at least $(1-\alpha)$, for all $\mathbf{z}_{i}, \mathbf{a}_i, \widehat{u}_i \in \mathcal{Z}_i \times  \mathcal{\mathbf{A}}_i \times  \widehat{\mathcal{U}}_i$ we have $\big|  g_i(\mathbf{z}_{i}, \mathbf{a}_i, \widehat{u}_i) - \mu_{g,i,t-1}(\mathbf{z}_{i}, \mathbf{a}_i, \widehat{u}_i) \big| \leq \beta_{i,t}\sigma_{g,i, t-1} (\mathbf{z}_{i}, \mathbf{a}_i, \widehat{u}_i) $, and 
 $|h(\mathbf{z}_i, \mathbf{a}_i, x_i)  - \mu_{h, i,t-1}(\mathbf{z}_i, \mathbf{a}_i, x_i)| \leq \beta_{i,t}\sigma_{h,i,t-1}(\mathbf{z}_i, \mathbf{a}_i, x_i) $. 
\end{assumption}

Following~\cite{chowdhury2019online},  at time $t$, let $\tilde{\mathbf{G}}$ be the statistically plausible function set of $\mathbf{G}$, i.e.,  $\tilde{\mathbf{G}}=\{\tilde{g}_i\}_{i=0}^d$. 
The following lemma bounds the value of $\tilde{\widehat{u}}$ with the variance of the encoder.  

\begin{lemma}~\label{lm:u_bound}
\begin{align*}
 \| \widehat{u}_{i,t} -  \tilde{\widehat{u}}_{i,t} \| \leq 2\beta_t\|\sigma_{\widehat{u}_{i,t-1}}\| = 2\beta_t\|\sigma_{h,i,t-1}\|  .
\end{align*}
\end{lemma}

\begin{proof}
With Assumption~\ref{assum:calibrated} and $ \widehat{u}_{i,t} =h_{i,i-1}(\mathbf{z}_i, \mathbf{a}_i, x_i) $, let $\tilde{\widehat{u}}_{i,t} = \mu_{\widehat{u}_{i,t-1}}\mathbf{z}_i, \mathbf{a}_i, x_i+ \beta_t\sigma_{\widehat{u}_{i,t-1}}(\mathbf{z}_i, \mathbf{a}_i, x_i) \circ  \boldsymbol{\omega}_{\widehat{u}_{i,t-1}}(\mathbf{z}_i, \mathbf{a}_i, x_i)$, and here $|\boldsymbol{\omega}_{\widehat{u}_{i,t-1}}(\mathbf{z}_i, \mathbf{a}_i, x_i)|\leq 1$.
Then 
\begin{align*}
 \| \widehat{u}_{i,t} -  \tilde{\widehat{u}}_{i,t} \| =& \| \tilde{\widehat{u}}_{i,t} -\mu_{\widehat{u}_{i,t-1}}(\mathbf{z}_i, \mathbf{a}_i, x_i)-\beta_t \sigma_{\widehat{u}_{i,t-1}}(\mathbf{z}_i, \mathbf{a}_i, x_i) \circ  \boldsymbol{\omega}_{\widehat{u}_{i,t-1}}(\mathbf{z}_i, \mathbf{a}_i, x_i) \|\\
 \leq & \| \tilde{\widehat{u}}_{i,t} -\mu_{\widehat{u}_{i,t-1}}(\mathbf{z}_i, \mathbf{a}_i, x_i)\| + \beta_t \|\sigma_{\widehat{u}_{i,t-1}}(\mathbf{z}_i, \mathbf{a}_i, x_i) \circ  \boldsymbol{\omega}_{\widehat{u}_{i,t-1}}(\mathbf{z}_i, \mathbf{a}_i, x_i) \| \\
 \leq & 2\beta_t  \|\sigma_{\widehat{u}_{i,t-1}}(\mathbf{z}_i, \mathbf{a}_i, x_i) \|= 2\beta_t\|\sigma_{h,i,t-1}\|.
\end{align*}
\end{proof}
 With the decomposable Assumption~\ref{assum:f} on $f_i$,  $\sigma^2_{h,i,t-1} \propto f^2_{i(b)}(\mathbf{z}_i, \mathbf{a}_i)\big(f_{i(c)}(U)  -   \mathbb{E}[f_{i(c)}(U)] \big)^2$ according to the proof of Theorem~\ref{thm:exo1}. $f_{i(b)}()$ is learned with the variance of regression model $\phi()$,  i.e. $\sigma_{\phi, i, t}()$.

\begin{lemma}\label{eq:x_bound}
\begin{align*}
\|x_{d,t} - \tilde{x}_{d,t}  \| \leq 2\beta_t M^{N_i}( 2\beta_t L_{\sigma_{g}} + L_{g} )^{N_i} \sum_{j=0}^i \big( \sigma_{g, j,t-1}(\mathbf{z}_{j,t})  + \sigma_{\widehat{u}_{j,t-1}} \big).
\end{align*}
\end{lemma}

\begin{proof}
We use $g_i(\mathbf{z}_{i,t}, \widehat{u}_{i,t})$ to represent $g_i(\mathbf{z}_{i,t},\mathbf{a}_{i,t}, \widehat{u}_{i,t})$ because we assume the actions to be the same for the process generating $x_{i,t}$ and $\tilde{x}_{i,t}$. Similarly, 
$\mu_{g,i,t-1}(\tilde{\mathbf{z}}_{i,t}, \tilde{\widehat{u}}_{i,t})  = \mu_{g,i,t-1}(\tilde{\mathbf{z}}_{i,t}, \tilde{\mathbf{a}}_{i,t},\tilde{\widehat{u}}_{i,t})$, $\sigma_{g,i,t-1}(\tilde{\mathbf{z}}_{i, t}, \tilde{\widehat{u}}_{i,t}) = \sigma_{g,i,t-1}(\tilde{\mathbf{z}}_{i,t}, \tilde{\mathbf{a}}_{i,t}, \tilde{\widehat{u}}_{i,t})$. 

We use the reparameterization trick, and write $\tilde{x}_{i,t}$ as
\begin{align*}
\tilde{x}_{i,t} = \tilde{g}_i(\tilde{\mathbf{z}}_{i,t}, \tilde{\widehat{u}}_{i,t}) = \mu_{g,i,t-1}(\tilde{\mathbf{z}}_{i,t}, \tilde{\widehat{u}}_{i,t})  + \beta_t \sigma_{g,i,t-1}(\tilde{\mathbf{z}}_{i}, \tilde{\widehat{u}}_{i,t}) \circ \boldsymbol{\omega}_{g,i,t-1}(\tilde{\mathbf{z}}_{i}, \tilde{\widehat{u}}_{i,t}) .
\end{align*}
Here $|\boldsymbol{\omega}_{g,i,t-1}(\tilde{\mathbf{z}}_{i}, \tilde{\widehat{u}}_{i,t})| \leq 1$. 
Hence, we have 
\begin{align*}
\|x_{i,t} - \tilde{x}_{i,t} \|
&= \| g_i(\mathbf{z}_{i,t},\widehat{u}_{i,t}) - \mu_{g,i,t-1}(\tilde{\mathbf{z}}_{i,t}, \tilde{\widehat{u}}_{i,t})  - \beta_t \sigma_{g,i,t-1}(\tilde{\mathbf{z}}_{i}, \tilde{\widehat{u}}_{i,t}) \boldsymbol{\omega}_{g,i,t-1}(\tilde{\mathbf{z}}_{i}, \tilde{\widehat{u}}_{i,t})   \| \\
&= \| g_i(\tilde{\mathbf{z}}_{i,t}, \tilde{\widehat{u}}_{i,t})    - \mu_{g,i,t-1}(\tilde{\mathbf{z}}_{i,t}, \tilde{\widehat{u}}_{i,t})  - \beta_t \sigma_{g,i,t-1}(\tilde{\mathbf{z}}_{i}, \tilde{\widehat{u}}_{i,t}) \boldsymbol{\omega}_{g,i,t-1}(\tilde{\mathbf{z}}_{i}, \tilde{\widehat{u}}_{i,t}) \\
& ~~~~~~~ + g_i(\mathbf{z}_{i,t},\widehat{u}_{i,t})  - g_i(\tilde{\mathbf{z}}_{i,t}, \tilde{\widehat{u}}_{i,t}) \| \\
&\leq  \| g_i(\tilde{\mathbf{z}}_{i,t}, \tilde{\widehat{u}}_{i,t})    - \mu_{g,i,t-1}(\tilde{\mathbf{z}}_{i,t}, \tilde{\widehat{u}}_{i,t}) \| + \| \beta_t \sigma_{g,i,t-1}(\tilde{\mathbf{z}}_{i}, \tilde{\widehat{u}}_{i,t}) \boldsymbol{\omega}_{g,i,t-1}(\tilde{\mathbf{z}}_{i}, \tilde{\widehat{u}}_{i,t}) \| \\
&~~~~~~~~~ + \|g_i(\mathbf{z}_{i,t},\widehat{u}_{i,t})  - g_i(\tilde{\mathbf{z}}_{i,t}, \tilde{\widehat{u}}_{i,t}) \|\\
&\overset{\zeta_1}{\leq }  \beta_t \|\sigma_{g,i,t-1}(\tilde{\mathbf{z}}_{i}, \tilde{\widehat{u}}_{i,t})\| + \beta_t \|\sigma_{g,i,t-1}(\tilde{\mathbf{z}}_{i}, \tilde{\widehat{u}}_{i,t})\| + L_{g_i} \big\|[\mathbf{z}_{i,t}; \widehat{u}_{i,t}] -  [\tilde{\mathbf{z}}_{i,t}; \tilde{\widehat{u}}_{i,t}] \big\|
 \\
&= 2\beta_t \|\sigma_{g,i,t-1}(\mathbf{z}_{i}, \widehat{u}_{i,t}) + \sigma_{g,i,t-1}(\tilde{\mathbf{z}}_{i}, \tilde{\widehat{u}}_{i,t}) - \sigma_{g,i,t-1}(\mathbf{z}_{i}, \widehat{u}_{i,t}) \|  + L_{g_i} \big\|[\mathbf{z}_{i,t}; \widehat{u}_{i,t}] -  [\tilde{\mathbf{z}}_{i,t}; \tilde{\widehat{u}}_{i,t}] \big\|\\
&  \overset{\zeta_2}{\leq }   2\beta_t \bigg(  \| \sigma_{g,i,t-1}(\mathbf{z}_{i}, \widehat{u}_{i,t}) \|  + L_{\sigma_{g,i}} \big\|[\mathbf{z}_{i,t}; \widehat{u}_{i,t}] -  [\tilde{\mathbf{z}}_{i,t}; \tilde{\widehat{u}}_{i,t}] \big\| \bigg) + L_{g_i} \big\|[\mathbf{z}_{i,t}; \widehat{u}_{i,t}] -  [\tilde{\mathbf{z}}_{i,t}; \tilde{\widehat{u}}_{i,t}] \big\| \\
& =2\beta_t  \sigma_{g,i,t-1}(\mathbf{z}_{i}, \widehat{u}_{i,t})     + ( 2\beta_t L_{\sigma_{g,i}} + L_{g_i} )\big\|[\mathbf{z}_{i,t}; \widehat{u}_{i,t}] -  [\tilde{\mathbf{z}}_{i,t}; \tilde{\widehat{u}}_{i,t}] \big\|  \\
& \leq 2\beta_t \sigma_{g,i,t-1}(\mathbf{z}_{i}, \widehat{u}_{i,t})  + ( 2\beta_t L_{\sigma_{g,i}} + L_{g_i} )\|\mathbf{z}_{i,t}-  \tilde{\mathbf{z}}_{i,t}\|  + ( 2\beta_t L_{\sigma_{g,i}} + L_{g_i} )\| \widehat{u}_{i,t} -  \tilde{\widehat{u}}_{i,t} \|  \\
 & \overset{\zeta_3}{\leq }  2\beta_t \sigma_{g,i,t-1}(\mathbf{z}_{i}, \widehat{u}_{i,t})  + ( 2\beta_t L_{\sigma_{g,i}} + L_{g_i} )\|\mathbf{z}_{i,t}-  \tilde{\mathbf{z}}_{i,t}\|  + 2\beta_t( 2\beta_t L_{\sigma_{g,i}} + L_{g_i} )\sigma_{\widehat{u}_{i,t-1}} \\
&= 2\beta_t \sigma_{g,i,t-1}(\mathbf{z}_{i}, \widehat{u}_{i,t})   + 2\beta_t( 2\beta_t L_{\sigma_{g,i}} + L_{g_i} )\sigma_{\widehat{u}_{i,t-1}}  + ( 2\beta_t L_{\sigma_{g,i}} + L_{g_i} ) \sum_{j \in \mathbf{pa}(i)}\| \mathbf{z}_{j,t} - \tilde{\mathbf{z}}_{j,t} \|  \\
&\leq 2\beta_t \sigma_{g,i,t-1}(\mathbf{z}_{i}, \widehat{u}_{i,t})   + 2\beta_t( 2\beta_t L_{\sigma_{g}} + L_{g} )\sigma_{\widehat{u}_{i,t-1}}  + ( 2\beta_t L_{\sigma_{g}} + L_{g} )\sum_{j \in \mathbf{pa}(i)}\| x_{j,t} - \tilde{x}_{j,t} \| \\
&\overset{\zeta_4}{\leq }  2\beta_t \sigma_{g,i,t-1}(\mathbf{z}_{i}, \widehat{u}_{i,t}) + 2\beta_t ( 2\beta_t L_{\sigma_{g}} + L_{g} )\sigma_{\widehat{u}_{i,t-1}}   \\
& ~~~~ + ( 2\beta_t L_{\sigma_{g}} + L_{g} )\sum_{j \in \mathbf{pa}(i)} 2 \beta_t M^{N_j}( 2\beta_t L_{\sigma_{g}} + L_{g} )^{N_j} \sum_{h=0}^j \big( \sigma_{g, h,t-1}(\mathbf{z}_{h,t})  + \sigma_{\widehat{u}_{h,t-1}} \big) \\
&\leq 2\beta_t M^{N_i}( 2\beta_t L_{\sigma_{g}} + L_{g} )^{N_i} \sum_{j=0}^i \big( \sigma_{g, j,t-1}(\mathbf{z}_{j,t})  + \sigma_{\widehat{u}_{j,t-1}} \big)
\end{align*}
In steps $\zeta_1$  and $\zeta_2$,  we rely on the calibrated uncertainty and Lipschitz dynamics; in step  $\zeta_2$, we also apply the triangle inequality; step $\zeta_3$ is by Lemma~\ref{lm:u_bound};
 $\zeta_4$ applies the inductive hypothesis. 
\end{proof}

\noindent\textbf{Theorem~\ref{thm:regret}}{\it \ \ 
Consider the optimization problem in~\eqref{eq:cbo_exo}, with the SCM satisfying Assumptions~\ref{assum:Lipschitz}-~\ref{assum:calibrated}, where $\mathcal{G}$ is known but $\mathbf{F}$ is unknown. Then with probability at least $1-\alpha$, the cumulative regret of Algorithm~\ref{alg:ucb} is bounded by 
\begin{align*}
R_T &\leq  \mathcal{O}(L_{T} M^{N} d  \sqrt{ T \gamma_T}).
\end{align*}
}

\begin{proof}
The cumulative regret is 
\begin{align*}
&R_T = \sum_{t=1}^T \bigg[ \mathbb{E}[y |\mathbf{a}^*] -  \mathbb{E}[y | \mathbf{a}_{:,t}]  \bigg]. 
\end{align*}

At step $t$, the instantaneous regret is $r_t$. By applying Lemma~\ref{eq:x_bound}, $r_t$ is bounded by
\begin{align*}
r_t &= \mathbb{E}[y |\mathbf{F}, \mathbf{a}^*] -  \mathbb{E}[y | \mathbf{F}, \mathbf{a}_{:,t}] \\
&\leq \mathbb{E}[y_t |\tilde{\mathbf{F}},  \mathbf{a}_{:,t}] -  \mathbb{E}[y_t | \mathbf{F}, \mathbf{a}_{:,t}] \\
&=\mathbb{E}[\|x_{i,t} - \tilde{x}_{i,t} \| | \mathbf{a}_{:,t}]\\
& \leq 2\beta_t M^{N}( 2\beta_t L_{\sigma_{g}} + L_{g} )^{N}  \mathbb{E} \bigg[\sum_{i=0}^d\| \sigma_{g,i,t-1}(\mathbf{z}_{i,t}) \| + \|\sigma_{\widehat{u}_{i,t-1}}\|  \bigg] 
\end{align*}
Here $L_t = 2\beta_t ( 2\beta_t L_{\sigma_{g}} + L_{g} )^{N}$.
Thus,
\begin{align*}
r^2_t &\leq L_t^2 M^{2N}  \bigg( \mathbb{E} \bigg[\sum_{i=0}^d\| \sigma_{g,i,t-1}(\mathbf{z}_{i,t}) \| + \|\sigma_{\widehat{u}_{i,t-1}}\|   \bigg] \bigg)^2 \\
&\leq 2 d L_t^2 M^{2N}  \mathbb{E} \bigg[\sum_{i=0}^d\| \sigma_{g,i,t-1}(\mathbf{z}_{i,t}) \|_2^2 + \|\sigma_{\widehat{u}_{i,t-1}}\|_2^2  \bigg] 
\end{align*}

We define $R^2_T$ as
\begin{align*}
R^2_T&= ( \sum_{t=1}^T r_t )^2 \leq T \sum_{t=1}^T r_t^2 \\
&\leq 2 d TL_{T}^2 M^{2N}  \sum_{t=1}^T  \mathbb{E} \bigg[\sum_{i=0}^d\| \sigma_{g,i,t-1}(\mathbf{z}_{i,t}) \|_2^2  + \|\sigma_{\widehat{u}_{i,t-1}}\|_2^2 \bigg]  \\
&=  2 d TL_{T}^2 M^{2N}   \Gamma_T.
\end{align*}
Here,
\begin{align*}
\Gamma_T=& \max_{(\mathbf{z},\mathbf{a}, \widehat{u}) \in \mathcal{Z}\times \mathcal{\mathbf{A}} \times  \widehat{\mathcal{U}}} \sum_{t=1}^T \sum_{i=0}^d \bigg[\|\sigma_{i, t-1}(\mathbf{z}_{i,t},\mathbf{a}_{i,t})\|_2^2 + \|\sigma_{\widehat{u}_{i,t-1}}\|_2^2 \bigg] \\
\leq &\max_{\mathbf{A}, \widehat{U}}\sum_{t=1}^T \sum_{i=0}^d \bigg[ \|\sigma_{i, t-1}(\mathbf{z}_{i,t},\mathbf{a}_{i,t})\|_2^2  + \|\sigma_{\widehat{u}_{i,t-1}}\|_2^2 \bigg]
\end{align*}
\begin{align*}
\leq &\sum_{i=0}^d \max_{\mathbf{A}_i, \widehat{U}_i} \sum_{t=1}^T \bigg[ \|\sigma_{i, t-1}(\mathbf{z}_{i,t},\mathbf{a}_{i,t})\|_2^2  + \|\sigma_{\widehat{u}_{i,t-1}}\|_2^2 \bigg]\\
\leq &\sum_{i=0}^d \max_{\mathbf{A}_i, \widehat{U}_i} \sum_{t=1}^T \bigg[\sum_{l=1}^{d_i} 
 \|\sigma_{i, t-1}(\mathbf{z}_{i,t},\mathbf{a}_{i,t},l)\|_2^2  + \|\sigma_{\widehat{u}_{i,t-1}}\|_2^2\bigg]\\
\overset{\zeta_1}{\leq } & \sum_{i=0}^d \frac{2}{\ln (1 + \rho_i^{-2})} \gamma_{i,T} \\
 = & \mathcal{O}(d \gamma_T).
\end{align*}

Here $\zeta_1$ is due to the upper bound of the information gain~\citep{srinivas2009gaussian}, and $\gamma_T$ will often scale sublinearly in $T$~\citep{sussex2022model}.  Therefore, 
\begin{align*}
R^2_T &\leq 2TL_{T}^2 M^{2N} d   \mathcal{O}(d \gamma_T).
\end{align*}
And,
\begin{align*}
R_T &\leq  \mathcal{O}(L_{T} M^{N} d  \sqrt{ T \gamma_T}).
\end{align*}
This completes the proof of the theorem.

\end{proof}


\end{document}